\documentclass{article} 
\PassOptionsToPackage{numbers, compress}{natbib}
\usepackage[final]{neurips_2024}

\usepackage{amsmath,amsfonts,bm}

\def\eqref#1{equation~\ref{#1}}

\def\1{\bm{1}}

\DeclareMathAlphabet{\mathsfit}{\encodingdefault}{\sfdefault}{m}{sl}
\SetMathAlphabet{\mathsfit}{bold}{\encodingdefault}{\sfdefault}{bx}{n}

\usepackage{color}
\usepackage[dvipsnames]{xcolor}
\usepackage{colortbl}
\usepackage{graphicx}
\usepackage{booktabs}
\usepackage{xspace}
\usepackage[utf8]{inputenc} 
\usepackage[T1]{fontenc}    
\usepackage[colorlinks,
            linkcolor=VioletRed,      
            anchorcolor=VioletRed, 
            citecolor=VioletRed,       
            ]{hyperref}

\usepackage{multirow}
\usepackage{wrapfig}
\usepackage{subfigure}
\usepackage{cleveref}
\newcommand{\abbr}[0]{MME-RealWorld\xspace} 
\definecolor{Gray}{gray}{0.85}
\newcommand{\Gray}[0]{\rowcolor{gray!20}}
\newcommand{\Lgray}[0]{\rowcolor{gray!10}}

\title{\Large  MME-RealWorld: Could Your Multimodal LLM \\ Challenge High-Resolution Real-World Scenarios \\ that are Difficult for Humans?}

\author{
\vspace{-0.4cm}
    \\ 
    Yi-Fan Zhang$^{1,5, \spadesuit}$, Huanyu Zhang$^{1,5}$, Haochen Tian$^{1,5}$, Chaoyou Fu$^{2,\dagger}$ \\
    Shuangqing Zhang$^{2}$, Junfei Wu$^{1,5}$, Feng Li$^{3}$, Kun Wang$^{4,5}$, Qingsong Wen$^{6,\dagger}$ \\
    Zhang Zhang$^{1,5,\dagger}$, Liang Wang$^{1,5}$, Rong Jin$^{7}$, Tieniu Tan$^{1,2,5}$
    \\ \\
    $^{1}$CASIA, $^{2}$NJU, $^{3}$HKUST, 
    $^{4}$NTU, $^{5}$UCAS,
    $^{6}$Squirrel AI Learning,
    $^{7}$Meta AI
    \\
    \footnotesize{
    $^{\spadesuit}$~Project Leader \;
    $^{\dagger}$~Corresponding Author \;}
    \\ \\
    {\centering}
    \url{https://mme-realworld.github.io/}
}

\begin{document}

\maketitle

\begin{abstract}
Comprehensive evaluation of Multimodal Large Language Models (MLLMs) has recently garnered widespread attention in the research community. However, we observe that existing benchmarks present several common barriers that make it difficult to measure the significant challenges that models face in the real world, including: 1) small data scale leads to a large performance variance; 2) reliance on model-based annotations results in restricted data quality; 3) insufficient task difficulty, especially caused by the limited image resolution. To tackle these issues, we introduce \abbr. Specifically, we collect more than $300$ K images from public datasets and the Internet, filtering $13,366$ high-quality images for annotation. This involves the efforts of professional $25$ annotators and $7$ experts in MLLMs, contributing to $29,429$ question-answer pairs that cover $43$ subtasks across $5$ real-world scenarios, extremely challenging even for humans. As far as we know, \textbf{\abbr is the largest manually annotated benchmark to date, featuring the highest resolution and a targeted focus on real-world applications}. We further conduct a thorough evaluation involving $29$ prominent MLLMs, such as GPT-4o, Gemini 1.5 Pro, and Claude 3.5 Sonnet. Our results show that even the most advanced models struggle with our benchmarks, where none of them reach 60\% accuracy. The challenges of perceiving high-resolution images and understanding complex real-world scenarios remain urgent issues to be addressed. The data and evaluation code are released in our Project Page.
\end{abstract}

\begin{figure}
    \centering
    \includegraphics[width=\linewidth]{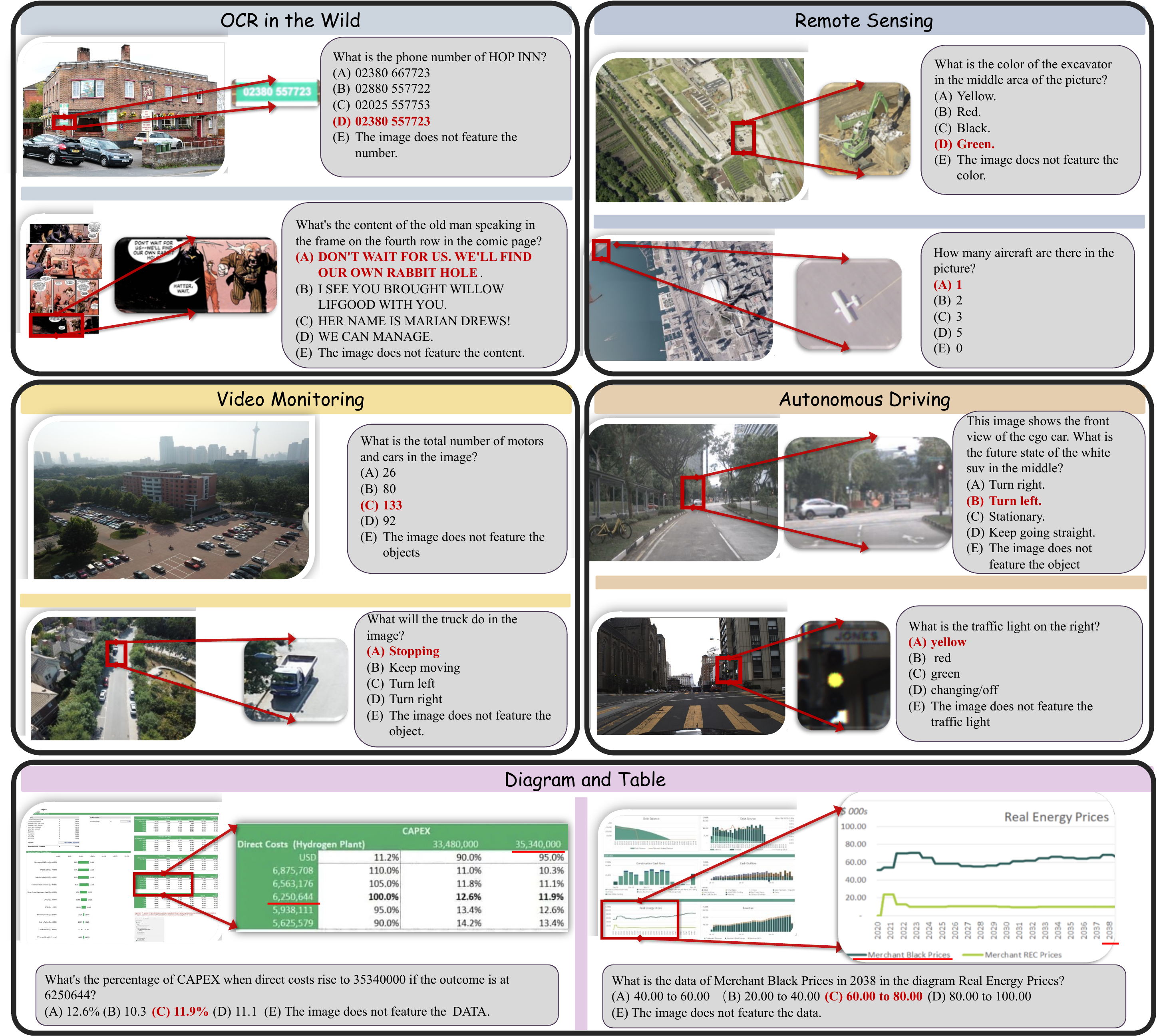}
\caption{\textbf{Diagram of \abbr.}
Our benchmark contains $5$ real-world domains, covering 43 perception and reasoning subtasks. Each QA pair offers $5$ options. We highlight and magnify the image parts relevant to the question in a red box for better visibility.}
\label{fig:teaser_tasks}
\end{figure}
\section{Introduction}

In recent years, we have witnessed a significant flourish of Multimodal Large Language Models (MLLMs)~\citep{dai2024instructblip, liu2023visual, zhang2024beyond}. A primary objective behind designing MLLMs has been to develop general intelligent agents capable of comprehensively perceiving human queries and environmental stituations through the integration of various multimodal sensory data. Consequently, a plethora of comprehensive evaluation benchmarks have emerged to rigorously assess model capabilities. 
However, some common concerns also arise:
\begin{itemize}
    \item \textbf{Data Scale.} 
    Many existing benchmarks contain fewer than $10$K Question-Answer (QA) pairs, such as MME~\citep{fu2023mme}, MMbench~\citep{liu2023mmbench}, MMStar~\citep{chen2024we}, MM-Vet~\citep{yu2024mm}, TorchStone~\citep{bai2023touchstone}, and BLINK~\citep{fu2024blink}. The limited number of QA can lead to large evaluation fluctuations.
    \item \textbf{Annotation Quality.} 
    While some benchmarks, such as MMT-Bench~\citep{mmtbench} and SEED-Bench~\citep{li2024seed}, are relatively larger in scale, their annotations are generated by LLMs or MLLMs. This annotation process is inherently limited by the performance of the used models. In our benchmark, for example, the best-performing model, InternVL-2, merely achieves $50\%$ accuracy. Consequently, relying on models would inevitably introduce significant noise, compromising the quality of the annotations.
    \item \textbf{Task Difficulty.} To date, the top performance of some benchmarks has reached the accuracy of $80\%$-$90\%$~\citep{mathew2021docvqa, masry2022chartqa, singh2019towards, liu2023mmbench, li2023seed}, and the performance margin between advanced MLLMs is narrow. This makes it challenging to verify the benefits or improvements of advanced models and to distinguish which one is significantly better.
\end{itemize}

\begin{figure*}[t]
\centering
\subfigure[\textbf{Real-World Tasks}]{
\begin{minipage}[t]{0.6\linewidth}
\centering
 \includegraphics[width=\linewidth]{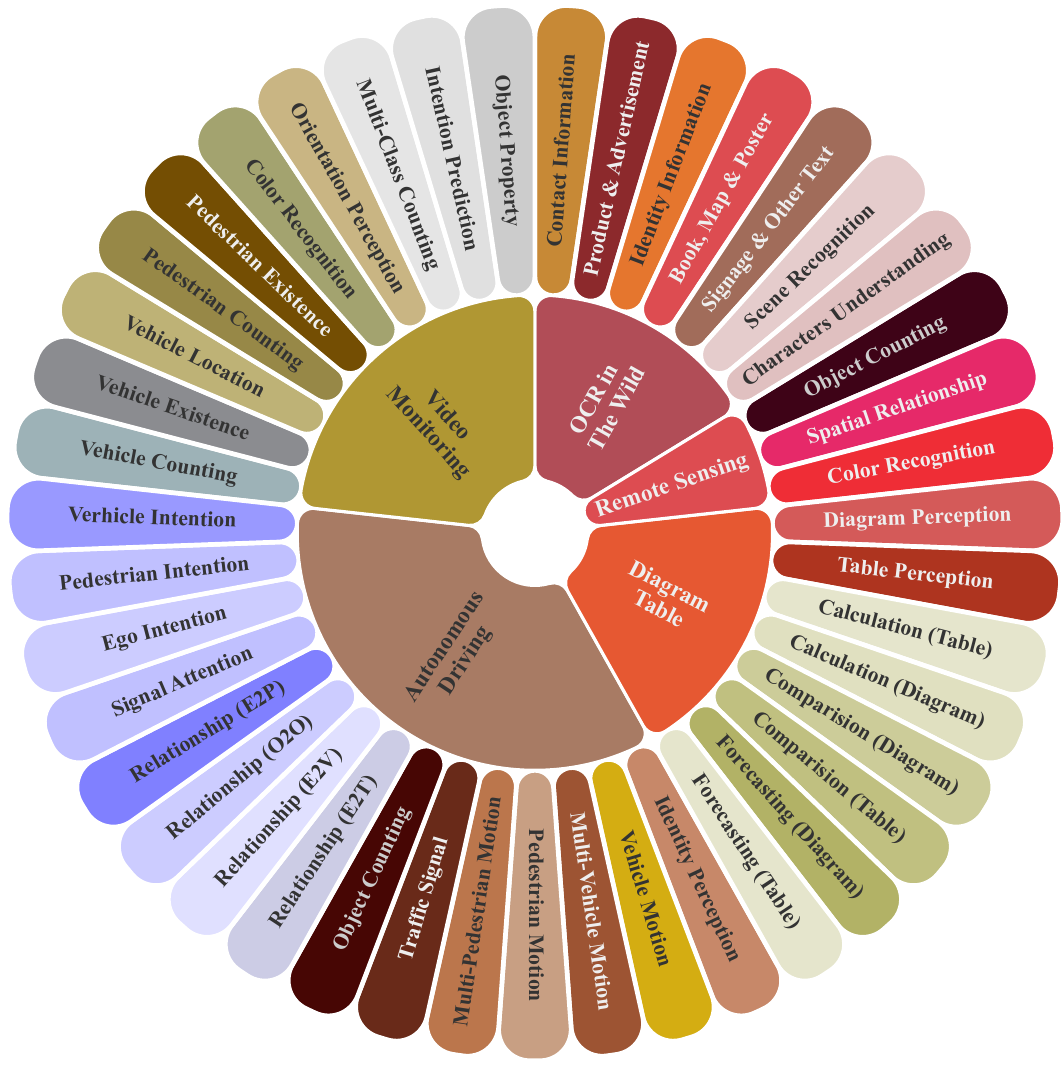}
\end{minipage}%
\label{label:teaser_task}
}%
\subfigure[\textbf{Leaderboard}]{
\begin{minipage}[t]{0.37\linewidth}
 \centering
 \includegraphics[width=\linewidth]{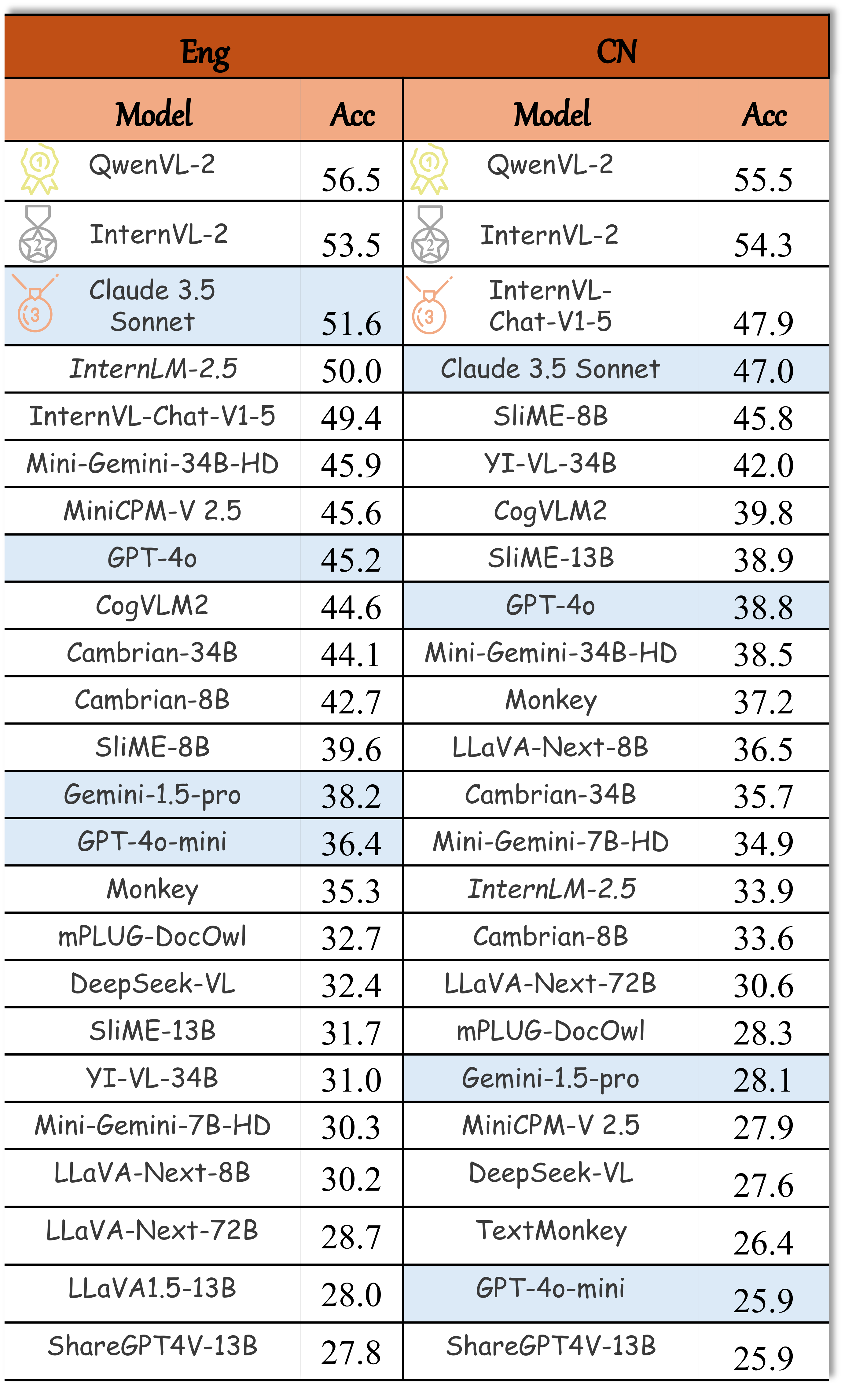}
\end{minipage}%
\label{label:teaser_acc}

}%
\centering
\vspace{-0.2cm}
\caption{\textbf{Task Categories} (left). Our benchmark spans $5$ key domains and $43$ subtasks highly related to real-world scenarios, including $13,366$ high-resolution images and $29,429$ annotations. \textbf{Model Performance} (right). Average accuracies of advanced MLLMs are shown across both the English and Chinese splits of the dataset.}
\label{fig:teaser}
\end{figure*}

In light of these concerns, we propose a new benchmark named \abbr. We first pay attention to a series of well-motivated families of datasets, considering images from sources such as autonomous driving, remote sensing, video surveillance, newspapers, street views, and financial charts. 
These scenarios are difficult even for humans, where we hope that MLLMs can really help.
Considering these topics, we collect a total of $13,366$ high-resolution images from more than $300$K public and internet sources. 
These images have an average resolution of $2,000$$\times$$1,500$, containing rich image details. 
$25$ professional annotators and $7$ experts in MLLMs are participated to annotate and check the data quality, and meanwhile ensuring that all questions are challenging for MLLMs. Note that most questions are even hard for humans, requiring multiple annotators to answer and double-check the results.
As shown in Fig.~\ref{label:teaser_task}, MME-RealWorld finally contains $29,429$ annotations for $43$ sub-class tasks, where each one has at least $100$ questions. 
$29$ advanced MLLMs are evaluated on our benchmark, along with detailed analysis. We conclude the main advantages of \abbr compared to existing counterparts as follows:

\begin{itemize}
    \item \textbf{Data Scale.} With the efforts of a total of $32$ volunteers, we have manually annotated $29,429$ QA pairs focused on real-world scenarios, making this the largest fully human-annotated benchmark known to date.
    \item \textbf{Data Quality.} 1) Resolution: Many image details, such as a scoreboard in a sports event, carry critical information. These details can only be properly interpreted with high-resolution images, which are essential for providing meaningful assistance to humans. To the best of our knowledge, \abbr features the highest average image resolution among existing competitors. 2) Annotation: All annotations are manually completed, with a professional team cross-checking the results to ensure data quality.
    \item \textbf{Task Difficulty and Real-World Utility.} The performance of different MLLMs is shown in Fig.~\ref{label:teaser_acc}, in which we can see that even the most advanced models have not surpassed $60\%$ accuracy. Additionally, as illustrated in Fig.~\ref{fig:teaser_tasks}, many real-world tasks are significantly more difficult than those in traditional benchmarks. For example, in video monitoring, a model needs to count the presence of $133$ vehicles, or in remote sensing, it must identify and count small objects on a map with an average resolution exceeding $5000$$\times$$5000$.
    \item \textbf{\abbr-CN.} Existing Chinese benchmark~\citep{liu2023mmbench} is usually translated from its English version. This has two limitations: 
    1) Question-image mismatch. The image may relate to an English scenario, which is not intuitively connected to a Chinese question.
    2) Translation mismatch~\citep{tang2024mtvqa}. The machine translation is not always precise and perfect enough. 
    We collect additional images that focus on Chinese scenarios, asking Chinese volunteers for annotation. This results in $5,917$ QA pairs.
\end{itemize}

\section{\abbr}
In this section, we outline the data collection process, question annotation procedure, and provide a statistical overview of each domain and subtask in \abbr and its Chinese version. We visualize different tasks from the $5$ image domains in Fig.~\ref{fig:teaser_tasks}.
Detailed information on data sources, evaluation tasks, and visualized results can be found in Sec.~\ref{sec:app_data_collection}.

\subsection{Instruction and Criterion}\label{sec:prompt}

For each question, we manually construct four options, with one being the correct answer and the other three being the texts appearing in the image or options similar to the correct one. This greatly enhances the difficulty, forcing the model to deeply understand the details of the image. We also provide an additional choice E, which allows the model to reject for answering because there is no right answer. We try to use the model’s default prompt for multiple-choice questions, but if the model does not have the default prompt, we use a common prompt as shown in Tab.~\ref{tab:prompt}.

\begin{table}[]
\centering
\caption{\textbf{Prompt setting of \abbr}.}\label{tab:prompt}
\begin{tabular}{l}
\toprule 
\begin{tabular}[c]{@{}l@{}}\texttt{[Image]} \texttt{[Question]} The choices are listed below:\\ 
(A) \texttt{[Choice A]}\\ 
(B) \texttt{[Choice B]}\\ 
(C) \texttt{[Choice C]}\\ 
(D) \texttt{[Choice D]}\\ 
(E) \texttt{[Choice E]}\\ 
Select the best answer to the above multiple-choice question based on the image. Respond with \\ only the letter (A, B, C, D, or E) of the correct option.\\ The best answer is:
\end{tabular} \\ \bottomrule
\end{tabular}%
\end{table}

\textbf{Evaluation Metric.} We first apply a rule-based filter to the answers generated by MLLM, aligning them with the given answer options and checking for correctness against the ground truth. Let the dataset be denoted as \(\mathcal{D} = \{\mathcal{D}_d = \{\mathcal{T}_t\}_{t=1}^{T_d}\}_{d=1}^{D}\), where each domain \(\mathcal{D}_d\) consists of \(T_d\) subtasks. For each subtask, we calculate the accuracy across all annotations. For each domain, we compute two metrics: 1) \textbf{Average Accuracy (Avg).} the weighted average accuracy across all subtasks, given by \({\sum_{t=1}^{T_d} \text{Avg}(\mathcal{T}_t) \times |\mathcal{T}_t|}/{|\mathcal{D}_d|}\), where $|\cdot|$ is the instance number contained in one set, and 2) \textbf{Class-based Average Accuracy (Avg-C).} the unweighted average accuracy across subtasks, given by \({\sum_{t=1}^{T_d} \text{Avg}(\mathcal{T}_t)}/{T_d}\). Similarly, for the entire dataset, we report the overall Average Accuracy across all samples, and the class-based average accuracy across domains.

\subsection{Data Collection and Annotation}

\textbf{Optical Character Recognition in the Wild (OCR).} 
It is specifically designed to evaluate the model's ability to perceive and understand textual information in the real-world. We manually selecte $3,293$ images with complex scenes and recognizable text information from $150,259$ images in existing high-resolution datasets as our image sources. These images span various categories such as street scenes, shops, posters, books, and competitions.
The volunteers are worked for annotation, each with at least a foundational understanding of multimodal models, to independently generate questions and answers. These annotations are subsequently reviewed and further refined by another volunteers. Based on the image annotations, we categorize these $3,297$ images into $5$ perception tasks, totaling $5,740$ QA pairs: contact information and addresses, identity information, products and advertisements, signage and other text, as well as natural text recognition in elevation maps and books. Additionally, there are two reasoning tasks with $500$ QA pairs: 1) scene understanding of the entire image, which requires the model to locate and comprehend important text such as competition results, and 2) character understanding, focusing on comics or posters where the model needs to analyze relationships and personalities based on dialogue or presentation.

\textbf{Remote Sensing (RS).} 
The images have a wide range of applications in real-world scenarios. Some images possess extremely high quality, with individual image sizes reaching up to $139$MB and containing very rich details, which makes it difficult even for humans to perceive specific objects. We manually select $1,298$ high-resolution images from over $70,000$ public remote sensing images, ensuring that each image is of high quality, with sufficient resolution and rich detail. One professional researcher is involved in annotating the data, and another researcher checks and improves the annotations, resulting in $3,738$ QA pairs. There are $3$ perception tasks: object counting, color recognition, and spatial relationship understanding.

\textbf{Diagram and Table (DT).} Although there are already some datasets related to table and chart understanding, they mostly feature simple scenes. We focus on highly complex chart data, such as financial reports, which contain extensive numerical information and mathematical content, presenting new challenges for MLLMs. We filter $2,570$ images from the internet, with annotations performed by two volunteer and reviewed by another one. We categorize these annotations into $4$ tasks based on the question format: 1) Diagram and Table Perception ($5,433$ QA pairs): involve locating specific values of elements within the diagrams and tables; 2) Diagram Reasoning ($250$ QA pairs): include tasks such as identifying the maximum and minimum values in a chart, performing simple calculations, and predicting trends; and 3) table Reasoning ($250$ QA pairs): focus on simple calculations related to specific elements, understanding mathematical concepts like maximum and minimum values, and locating corresponding elements.

\textbf{Autonomous Driving (AD).}
It demands extensive general knowledge and embodied understanding capability. We emphasize challenging driving scenarios that involve distant perceptions and intricate interactions among dynamic traffic agents. Specifically, we manually select a subset of $2,715$ images from over $40,000$ front-view images captured by onboard cameras in open-source datasets. These images cover a diverse range of weather conditions, geographic locations, and traffic scenarios. Besides, a volunteer carefully annotates each image, and the other one conducts a thorough review, resulting in $3,660$ QA pairs for perception tasks and $1,334$ QA pairs for reasoning tasks. The perception tasks include objects identification, object attribute identification, and object counting for traffic elements such as vehicle, pedestrian, and signals. 
The latter is categorized into $3$ main tasks: 1) Intention Prediction: focus on predicting driving intention of a designated traffic agent in the short-term future. 2) Interaction Relation Understanding: involve reasoning about ego vehicle's reaction to other traffic elements, and the interactions between these elements. 3) Driver Attention Understanding: require reasoning about the traffic signal that the driver should pay attention to.

\textbf{Monitoring (MO).}
The images are from various application scenarios for public safety, e.g., streets, shopping malls, and expressway intersections. We focus on complex high-resolution monitoring images that include many real-world challenges, like scale variations and out-of-view, as possible which could test whether the model handles them robustly in practice.
Specifically, $1,601$ high-resolution images are manually selected from over $10,000$ public dataset images, which are captured from a broad range of cameras, viewpoints, scene complexities, and environmental factors across day and night. In terms of annotations, two volunteers manually annotate each image carefully, and multi-stage careful inspections and modifications are performed by another one. When these refined image annotations are completed, $1,601$ images are categorized into $3$ main perception tasks, totaling $2,196$ QA pairs, including object counting and location, and attribute recognition. Furthermore, $3$ reasoning tasks are well-designed with $498$ QA pairs: 1) calculate the sum of different objects, which requires the model to perceive various objects and calculate their total number accurately; 2) intention reasoning, focusing on reasoning the next route and turn of the specific object; 3) attribute reasoning, focusing on reasoning the specific materials and functions of the given objects.

\subsection{\abbr-CN} 
The traditional general VQA approach~\citep{liu2023mmbench} uses a translation engine to extend QA pairs from English to Chinese. However, it may face visual-textual misalignment problems~\citep{tang2024mtvqa}, failing to address complexities related to nuanced meaning, contextual distortion, language bias, and question-type diversity. Additionally, asking questions in Chinese about images containing only English texts is not intuitive for benchmarking Chinese VQA capabilities. By contrast, we follow the steps below to construct a high-quality Chinese benchmark:

\begin{itemize}
    \item \textbf{Selection.} For video monitoring, autonomous driving, and remote sensing, many images do not contain English information. Therefore, we select a subset of the aforementioned question pairs, double-checking to ensure they do not contain any English information.
    \item \textbf{Translation.} Translate the questions and answers by four professional researchers, all of whom are familiar with both English and Chinese.
    \item \textbf{Collection.} For diagrams and tables, since the original images often contain English information (e.g., legends/captions), we collect additional $300$ tables and $301$ diagrams from the Internet, where the contents are in Chinese. This data is further annotated by one volunteer, resulting in $301$$\times$$4$ QA pairs, where the task type is the same as diagram and table in \abbr. Similarly, for OCR in the wild, we also collect additional $939$ images for all the subtasks.
\end{itemize}

In total, \abbr-CN has $1,889$ additional images and total $5,917$ QA pairs, which is a smaller version of \abbr, but it retains similar task types, image quality, and task difficulty. 
The examples can be seen in Fig.~\ref{fig:mme-hd-cn}.

\subsection{Quality Control and Analysis}
\begin{wraptable}{r}{0.5\linewidth}
\vspace{-0.23cm} 
\caption{\textbf{Comparison of benchmarks.} MME-RealWorld is the largest fully human-annotated dataset, featuring the highest average resolution and the most challenging tasks.}
\label{fig:teaser_image}
\resizebox{0.5\textwidth}{!}{%
\begin{tabular}{lccccc}
\bottomrule
\textbf{Benchmark} & \textbf{\# QA-Pair} & \textbf{\begin{tabular}[c]{@{}c@{}}Fully Human \\ Annotation\end{tabular}} & \textbf{CN} & \textbf{\begin{tabular}[c]{@{}c@{}}Average\\ Resolution\end{tabular}} & \textbf{\begin{tabular}[c]{@{}c@{}}LLaVA-1.5-7B \\ Performance\end{tabular}} \\ \hline
VizWiz & 8000 & × & × & 1224$\times$1224 & 50.0 \\
RealWorldQA & 765 & × & × & 1536$\times$863 & - \\
MMStar & 1500 & × & × & 512$\times$375 & 30.3 \\
ScienceQA & 21000 & × & × & 378$\times$249 & 71.6 \\
ChartQA & 32719 & × & × & 840$\times$535 & - \\
MM-Vet & 218 & × & × & 1200$\times$675 & 31.1 \\
Seed-Bench & 19242 & × & × & 1024$\times$931 & 66.1 \\
SEED-Bench-2-Plus & 2300 & × & × & 1128$\times$846 & 36.8 \\
MMT-Bench & 32325 & × & × & 2365$\times$377 & 49.5 \\
MathVista & 735 & × & × & 539$\times$446 & 26.1 \\
TouchStone & 908 & × & × & 897$\times$803 & - \\
VisIT-Bench & 1159 & × & × & 765$\times$1024 & - \\
BLINK & 3807 & × & × & 620$\times$1024 & 37.1 \\
CV-Bench & 2638 & × & × & 1024$\times$768 & - \\ 
TextVQA & 5734 & \checkmark & × & 985$\times$768 & 58.2 \\
MME & 2374 & \checkmark & × & 1161$\times$840 & 76.0 \\
MMBench & 3217 & \checkmark & \checkmark & 512$\times$270 & 64.3 \\ \Gray
\abbr & 29429 & \checkmark & \checkmark & 2000x1500 & 24.9 \\ \bottomrule
\end{tabular}%
}
\end{wraptable} 
During the annotation process, we impose the following requirements on annotators:
1. We ensure that all questions can be answered based on the image (except for specially constructed questions where the correct option is “E”), meaning that humans can always find the answers within the image. This approach prevents forcing annotators to provide answers based on low-quality images or images containing vague information.
2. The area of the object being questioned in each image must not exceed $1/10$ of the total image area. This ensures that the object is not overly prominent, preventing humans from easily identifying the answer at first glance.
3. Each annotation is cross-checked by at least two professional multimodal researchers to ensure accuracy and prevent annotation errors caused by human bias.

The comparison of benchmarks is shown in Tab.~\ref{fig:teaser_image}. The maximal resolution of \abbr is $42,177,408$ pixels, with dimensions of $5304$$\times$$7952$. The average resolution is $3,007,695$ pixels, equivalent to an image size of approximately $2000$$\times$$1500$. This resolution is significantly higher than that of existing benchmarks. For instance, the highest benchmark, MME, has an average resolution of $975,240$ pixels, corresponding to an image size of about $1161$$\times$$840$. The exceptional image quality and our strict, fully human annotation process make our tasks the most challenging among all benchmarks. This is evidenced by the baseline model LLaVA-1.5-7B achieving an accuracy of just 24.9\%, significantly lower than on other benchmarks. Although some benchmarks may approach our level of difficulty, this is primarily due to the inherent complexity of their tasks. For instance, MathVista focuses on pure mathematical problems, and MM-Vet involves multi-step reasoning—both of which are naturally challenging and result in lower baseline performance. However, the majority of our tasks are centered on real-world perception problems. This means that, current MLLMs still struggle to effectively address human-level perceptual challenges.

\section{Experiments}

We evaluate a total of 24 open-source MLLMs, including Qwen-VL-Chat~\citep{bai2023qwen}, LLaVA, LLaVA-Next~\citep{li2024llavanext-strong}, TextMonkey~\citep{liu2024textmonkey}, mPLUG-DocOwl 1.5~\citep{hu2024mplug}, ShareGPT4V~\citep{chen2023sharegpt4v}, MiniGPT-v2~\citep{chen2023minigpt}, Monkey~\citep{li2023monkey}, OtterHD~\citep{li2023otter}, Cambrian-1~\citep{tong2024cambrian}, Mini-Gemini-HD~\citep{li2024mini}, MiniCPM-V 2.5~\citep{hu2024minicpm}, DeepSeek-VL~\citep{lu2024deepseek}, YI-VL-34B\footnote{\url{https://huggingface.co/01-ai/Yi-VL-34B}}, SliME~\citep{zhang2024beyond}, CogVLM2\footnote{\url{https://github.com/THUDM/CogVLM2}}, InternLM-XComposer2.5~\citep{zhang2023internlm}, InternVL-Chat V1-5, InternVL-2~\citep{chen2023internvl}, and Qwen2-VL\footnote{\url{https://github.com/QwenLM/Qwen2-VL}}, as well as 4 close-source MLLMs, including, GPT-4o\footnote{\url{https://openai.com/index/hello-gpt-4o/}}, GPT-4o-mini, Gemini 1.5 pro~\citep{team2023gemini}, and Claude 3.5 Sonnet\footnote{\url{https://www.anthropic.com/news/claude-3-5-sonnet}}. 


\subsection{Results on \abbr}
\subsubsection{Perception}
\begin{table}[]
\caption{\textbf{Experimental results on the perception tasks.} Models are ranked according to their average performance. Rows corresponding to proprietary models are highlighted in gray for distinction. “OCR”, “RS”, “DT”, “MO”, and “AD” each indicate a specific task domain: Optical Character Recognition in the Wild, Remote Sensing, Diagram and Table, Monitoring, and  Autonomous Driving, respectively. “Avg” and “Avg-C” indicate the weighted average accuracy
and the unweighted average accuracy across
subtasks in each domain.}\label{tab:res_main_perception}
\centering
\resizebox{0.95\textwidth}{!}{%
\begin{tabular}{llccccccc}
\toprule \Gray
\multicolumn{1}{c}{\textbf{Method}} & \multicolumn{1}{c}{\textbf{LLM}} & \multicolumn{7}{c}{\textbf{Perception}}  \\\cmidrule{1-2} \cmidrule{3-9}\Gray
\multicolumn{2}{c}{\textbf{Task Split}} & \textbf{OCR} & \textbf{RS} & \textbf{DT} & \textbf{MO} & \textbf{AD} & \textbf{Avg} & \textbf{Avg-C} \\  \Gray
\multicolumn{2}{c}{\textbf{\# QA pairs}}  & 5740 & 3738 & 5433 & 2196 & 3660 & 20767 & 20767 \\ \hline 
Qwen2-VL & Qwen2-7B & 81.38 & 44.81 & 70.18 & 37.30 & 34.62 & 58.96 & 53.66 \\
{InternVL-2} & InternLM2.5-7B-Chat & 73.92 & 39.35 & 62.80 & 53.19 & 35.46 & 55.82 & 52.94 \\ \Lgray
{Claude 3.5 Sonnet} & - & 72.47 & 25.74 & 67.44 & 32.19 & 40.77 & 52.90 & 47.72 \\
{InternLM-XComposer2.5} & InternLM2-7B & 69.25 & 36.12 & 63.92 & 39.48 & 33.63 & 52.47 & 48.48 \\
{InternVL-Chat-V1.5} & InternLM2-Chat-20B & 71.51 & 33.55 & 55.83 & 51.16 & 31.42 & 51.36 & 48.69 \\
{Mini-Gemini-34B-HD} & Nous-Hermes-2-Yi-34B & 69.55 & 40.40 & 44.36 & 39.61 & 32.70 & 48.05 & 45.32 \\
{MiniCPM-V 2.5} & Llama3-8B & 66.79 & 27.69 & 52.81 & 38.70 & 34.15 & 47.37 & 44.03 \\
{Cambrian-1-34B} & Nous-Hermes-2-Yi-34B & 66.45 & 38.63 & 40.44 & 45.98 & 33.61 & 46.68 & 45.02 \\ \Lgray
{GPT-4o} & - & 77.69 & 28.92 & 46.68 & 33.93 & 22.43 & 46.43 & 41.93 \\
CogVLM2-llama3-Chat & Llama3-8B & 69.97 & 28.76 & 47.51 & 33.74 & 30.22 & 45.84 & 42.04 \\
Cambrian-1-8B & Llama3-8B-Instruct & 58.68 & 40.05 & 32.73 & 47.68 & 38.52 & 43.82 & 43.53 \\
SliME-8B & Llama3-8B & 53.45 & 42.27 & 29.34 & 40.62 & 33.66 & 40.29 & 39.87 \\\Lgray
Gemini-1.5-pro & - & 67.62 & 13.99 & 39.90 & 31.11 & 26.64 & 39.63 & 35.85 \\ \Lgray
GPT-4o-mini & - & 62.51 & 6.69 & 44.23 & 26.50 & 24.18 & 37.12 & 32.82 \\
Monkey & Qwen-7B & 54.63 & 24.99 & 32.51 & 28.01 & 29.67 & 36.30 & 33.96 \\
mPLUG-DocOwl 1.5 & Llama-7B & 51.15 & 23.71 & 29.34 & 24.97 & 28.28 & 33.71 & 31.49 \\
DeepSeek-VL & DeepSeek-LLM-7b-base & 49.55 & 25.49 & 23.38 & 26.97 & 33.39 & 33.14 & 31.76 \\
SliME-13B & Vicuna-13B & 50.58 & 25.82 & 20.93 & 24.73 & 27.16 & 31.50 & 29.84 \\
Mini-Gemini-7B-HD & Vicuna-7B-v1.5 & 42.02 & 31.30 & 22.31 & 34.15 & 24.81 & 31.07 & 30.92 \\
YI-VL-34B & Yi-34B-Chat & 44.95 & 31.62 & 15.99 & 34.85 & 28.31 & 30.97 & 31.14 \\
LLaVA-Next & Llama3-8B & 47.94 & 25.42 & 26.63 & 19.46 & 18.66 & 30.14 & 27.62 \\
LLaVA-Next & Qwen-72B & 37.07 & 29.13 & 27.68 & 29.37 & 17.98 & 29.01 & 28.25 \\
LLaVA1.5-13B & Vicuna-13B & 44.10 & 23.27 & 20.17 & 20.45 & 26.12 & 28.42 & 26.82 \\
ShareGPT4V-13B & Vicuna-13B & 44.55 & 23.06 & 20.17 & 19.26 & 26.12 & 28.38 & 26.63 \\
MiniGPT-v2 & Llama 2-7B-Chat & 39.02 & 23.33 & 20.41 & 19.26 & 25.96 & 26.94 & 25.60 \\
ShareGPT4V-7B & Vicuna-7B & 39.39 & 22.10 & 20.08 & 19.13 & 26.04 & 26.73 & 25.35 \\
LLaVA1.5-7B & Vicuna-7B & 38.69 & 22.12 & 20.08 & 19.13 & 26.04 & 26.54 & 25.21 \\
Qwen-VL-Chat & Qwen-7B & 32.37 & 15.14 & 15.59 & 22.13 & 15.08 & 20.75 & 20.06 \\
TextMonkey & Qwen-7B & 37.30 & 11.69 & 5.93 & 16.14 & 14.26 & 18.18 & 17.06 \\ \bottomrule
\end{tabular}%
}
\vspace{-0.4cm}
\end{table}
Tab.~\ref{tab:res_main_perception} presents the perception capabilities of different models across $5$ domains. Overall, InternVL-2 demonstrates the strongest perception abilities, outperforming other closed-source models. However, the performance varies across different tasks, with some key observations as follows:

\begin{itemize}
    \item  GPT-4o performs well in real-world OCR tasks, achieving 77\% accuracy, which is second only to Qwen2-VL. However, its performance significantly drops in more challenging tasks, lagging behind other top-ranked models. This trend is also observed in other closed-source models, such as Gemini-1.5-Pro and GPT-4o-mini, which perform well in OCR tasks but struggle significantly in other real-world tasks. There are three possible reasons: 1) Close-source models often have limitations on the maximum image size and resolution when uploading local images. For example, Claude 3.5 Sonnet has a maximum resolution limit of $8$K and a maximum image quality of $5$MB, while GPT-4o and Gemini-pro allow up to $20$MB. This restricts the input of some high-quality images, as we have to compress the images for upload. 2) Close-source models tend to be more conservative. We observe that the proportion of responses, where closed-source models output ``E'' indicating that the object in question is not present in the image, is high. This suggests that these models may adopt a conservative response strategy to avoid hallucinations or to provide safer answers. 3) Closed-source models sometimes refuse to answer certain questions. Due to different input/output filtering strategies, some samples are considered to involve privacy or harmful content and are therefore not answered.
    \item Models allowing higher resolution input, such as Mini-Gemini-HD and SliME, demonstrate a significant advantage over models directly using vision encoders like CLIP, such as ShareGPT4V and LLaVA1.5. At the same model size, these models consistently improve across different subtasks. This highlights the critical importance of high-resolution image processing for addressing complex real-world tasks.
    \item There are also notable trends across different domains. Remote sensing tasks involve processing extremely large images, demanding a deeper comprehension of image details. Models that focus on high-resolution input, such as Cambrian-1, Mini-Gemini-HD, and SliME, outperform other models in these tasks. Additionally, models trained on large amounts of chart data exhibit improved perception capabilities for complex charts. For instance, SliME and LLaVA1.5 have limited and relatively simple chart data in their training sets, resulting in inferior performance in this category compared to more recent models.
\end{itemize}

\subsubsection{Reasoning}
Experimental results on the reasoning tasks are shown in Tab.~\ref{tab:res_main_reasoning}. In terms of reasoning ability, Claude 3.5 Sonnet distinguishes itself as the top performer across most domains, particularly outpacing the second-place Qwen2-VL by $12.6\%$ in chart-related tasks. The closed-source model GPT-4o also performs well, trailing slightly behind the third-place InternVL-2 but even outperforming it in several domains. Most open-source models perform poorly, with traditional baseline methods such as LLaVA1.5 and Qwen-VL-Chat yielding results close to random guessing.
Furthermore, reasoning tasks are more challenging than perception tasks. Even the top-ranked model fails to achieve an average accuracy above $45\%$, with class-based accuracy not exceeding $50\%$. This indicates that current models still have a significant gap to bridge to reach human-level reasoning capabilities.

\begin{table}[ht]
\caption{\textbf{Experimental results on the reasoning tasks.} Models are ranked according to their average performance. Rows corresponding to proprietary models are highlighted in gray for distinction. “OCR”, “RS”, “DT”, “MO”, and “AD” each indicate a specific task domain: Optical Character Recognition in the Wild, Remote Sensing, Diagram and Table, Monitoring, and  Autonomous Driving, respectively. “Avg” and “Avg-C” indicate the weighted average accuracy
and the unweighted average accuracy across
subtasks in each domain.}\label{tab:res_main_reasoning}
\centering
\resizebox{0.9\textwidth}{!}{%
\begin{tabular}{llcccccc}
\toprule \Gray
\multicolumn{1}{c}{\textbf{Method}} & \multicolumn{1}{c}{\textbf{LLM}} & \multicolumn{6}{c}{\textbf{Reasoning}} \\ \cmidrule{3-8} \Gray
\multicolumn{2}{c}{\textbf{Task Split}} & \textbf{OCR} & \textbf{DT} & \textbf{MO} & \textbf{AD} & \textbf{Avg} & \textbf{Avg-C} \\ \Gray
\multicolumn{2}{c}{\textbf{\# QA pairs}} & 500 & 500 & 498 & 1334 & 2832 & 2832 \\ \hline \Lgray
Claude 3.5 Sonnet & - & 61.90 & 61.20 & 41.79 & 31.92 & 44.12 & 49.20 \\
Qwen2-VL & Qwen2-7B & 63.40 & 48.60 & 33.13 & 31.47 & 40.39 & 44.15 \\
InternVL-2 & InternLM2.5-7B-Chat & 57.40 & 39.00 & 43.57 & 29.84 & 38.74 & 42.45 \\ \Lgray
GPT-4o & - & 61.40 & 44.80 & 36.51 & 26.41 & 37.61 & 42.28 \\
CogVLM2-llama3-Chat & Llama3-8B & 54.00 & 32.80 & 41.16 & 31.18 & 37.25 & 39.79 \\
InternVL-Chat-V1-5 & InternLM2-Chat-20B & 56.80 & 35.40 & 37.35 & 28.94 & 36.48 & 39.62 \\
Cambrian-1-8B & Llama3-8B-Instruct & 53.20 & 27.40 & 42.37 & 30.73 & 36.16 & 38.43 \\
SliME-8B & Llama3-8B & 53.20 & 29.40 & 36.14 & 31.55 & 35.80 & 37.57 \\
MiniCPM-V 2.5 & Llama3-8B & 44.00 & 31.80 & 36.95 & 31.03 & 34.50 & 35.95 \\
SliME-13B & Vicuna-13B & 41.00 & 39.00 & 33.13 & 30.80 & 34.46 & 35.98 \\
InternLM-XComposer2.5 & InternLM2-7B & 53.40 & 41.00 & 17.67 & 29.99 & 33.90 & 35.52 \\ \Lgray
GPT-4o-mini & - & 47.00 & 39.80 & 25.81 & 26.79 & 32.48 & 34.85 \\
YI-VL-34B & Yi-34B-Chat & 42.40 & 26.00 & 31.33 & 31.55 & 32.45 & 32.82 \\
LLaVA-Next & Llama3-8B & 55.20 & 23.40 & 21.08 & 30.73 & 32.06 & 32.60 \\
Mini-Gemini-34B-HD & Nous-Hermes-2-Yi-34B & 59.20 & 39.20 & 20.48 & 22.84 & 31.73 & 35.43 \\ \Lgray
Gemini-1.5-pro & - & 52.70 & 33.20 & 28.33 & 19.20 & 29.19 & 33.36 \\
Monkey & Qwen-7B & 27.20 & 20.80 & 27.31 & 33.04 & 28.84 & 27.09 \\
DeepSeek-VL & DeepSeek-LLM-7b-base & 45.20 & 23.80 & 16.67 & 27.31 & 27.98 & 28.25 \\
LLaVA-Next & Qwen-72B & 17.20 & 34.20 & 27.31 & 29.69 & 27.86 & 27.10 \\
Cambrian-1-34B & Nous-Hermes-2-Yi-34B & 55.00 & 36.00 & 19.48 & 16.07 & 27.06 & 31.64 \\
mPLUG-DocOwl 1.5 & Llama-7B & 42.60 & 19.80 & 20.48 & 26.04 & 26.88 & 27.23 \\
Mini-Gemini-7B-HD & Vicuna-7B-v1.5 & 35.40 & 24.60 & 25.90 & 23.29 & 26.12 & 27.30 \\
LLaVA1.5-13B & Vicuna-13B & 30.20 & 20.80 & 27.51 & 24.78 & 25.51 & 25.82 \\
ShareGPT4V-13B & Vicuna-13B & 26.00 & 20.80 & 27.31 & 24.55 & 24.63 & 24.67 \\
LLaVA1.5-7B & Vicuna-7B & 26.00 & 20.60 & 25.90 & 24.18 & 24.17 & 24.17 \\
ShareGPT4V-7B & Vicuna-7B & 24.15 & 20.60 & 26.10 & 24.18 & 23.88 & 23.76 \\
MiniGPT-v2 & Llama 2-7B-Chat & 30.00 & 20.40 & 16.87 & 23.66 & 23.01 & 22.73 \\
Qwen-VL-Chat & Qwen-7B & 28.60 & 13.60 & 16.47 & 24.63 & 21.95 & 20.83 \\
TextMonkey & Qwen-7B & 30.40 & 2.20 & 4.42 & 20.01 & 15.96 & 14.26 \\ \bottomrule
\end{tabular}%
}
\end{table}

\subsection{Results on \abbr-CN}

\begin{table}[t]
\caption{\textbf{Experimental results on the perception tasks of \abbr-CN.} Models are ranked according to their average performance. Rows corresponding to proprietary models are highlighted in gray for distinction. “OCR”, “RS”, “DT”, “MO”, and “AD” each indicate a specific task domain: Optical Character Recognition in the Wild, Remote Sensing, Diagram and Table, Monitoring, and  Autonomous Driving, respectively. “Avg” and “Avg-C” indicate the weighted average accuracy
and the unweighted average accuracy across
subtasks in each domain.}\label{tab:res_main_perception-cn}
\centering
\resizebox{0.95\textwidth}{!}{%
\begin{tabular}{llccccccc}
\toprule \Gray
\multicolumn{1}{c}{\textbf{Method}} & \multicolumn{1}{c}{\textbf{LLM}} & \multicolumn{7}{c}{\textbf{Perception}}  \\\cmidrule{1-2} \cmidrule{3-9}\Gray
\multicolumn{2}{c}{\textbf{Task Split}} & \textbf{OCR} & \textbf{RS} & \textbf{DT} & \textbf{MO} & \textbf{AD} & \textbf{Avg} & \textbf{Avg-C} \\  \Gray
\multicolumn{2}{c}{\textbf{\# QA pairs}} & 1908 & 300 & 602 & 500 & 700 & 4010 & 4010 \\ \hline
Qwen2-VL & Qwen-7B & 70.28 & 38.33 & 89.20 & 29.40 & 36.86 & 59.80 & 52.81 \\

InternVL-2 & InternLM2.5-7B-Chat & 69.92 & 41.33 & 71.63 & 39.3 & 34.14 & 57.97 & 51.26 \\
InternVL-Chat-V1-5 & InternLM2-Chat-20B & 60.59 & 32.00 & 60.12 & 32.40 & 32.14 & 49.90 & 43.45 \\ \Gray
Claude 3.5 Sonnet & - & 54.44 & 32.67 & 74.09 & 25.00 & 32.43 & 48.25 & 43.73 \\
SliME-8B & Llama3-8B & 53.93 & 41.33 & 58.25 & 29.20 & 31.29 & 46.60 & 42.80 \\ \Gray
GPT-4o & - & 55.90 & 23.67 & 54.86 & 25.20 & 21.14 & 43.44 & 36.15 \\
YI-VL-34B & Yi-34B-Chat & 51.41 & 34.33 & 49.52 & 25.20 & 27.71 & 42.45 & 37.63 \\
SliME-13B & Vicuna-13B & 50.63 & 17.33 & 48.49 & 17.80 & 33.23 & 40.69 & 33.50 \\
Cambrian-1-34B & Nous-Hermes-2-Yi-34B & 48.11 & 33.79 & 44.34 & 27.60 & 26.43 & 40.13 & 36.05 \\
CogVLM2-llama3-Chat & Llama3-8B & 46.12 & 22.00 & 39.48 & 24.80 & 34.14 & 38.57 & 33.31 \\
Mini-Gemini-34B-HD & Nous-Hermes-2-Yi-34B & 41.82 & 38.28 & 40.60 & 27.80 & 34.29 & 38.31 & 36.56 \\
LLaVA-Next & Llama3-8B & 40.62 & 31.67 & 37.49 & 35.40 & 27.29 & 36.50 & 34.49 \\ \Gray
Gemini-1.5-pro & - & 48.32 & 12.33 & 39.78 & 25.20 & 17.57 & 36.10 & 28.64 \\
Monkey & Qwen-7B & 40.46 & 26.55 & 41.12 & 19.20 & 35.86 & 36.07 & 32.64 \\
InternLM-XComposer2.5 & InternLM2-7B & 39.26 & 38.33 & 38.88 & 19.40 & 33.57 & 35.66 & 33.89 \\
Mini-Gemini-7B-HD & Vicuna-7B-v1.5 & 39.66 & 17.24 & 39.29 & 16.80 & 28.29 & 33.09 & 28.26 \\
Cambrian-1-8B & Llama3-8B-Instruct & 32.71 & 35.86 & 30.28 & 27.60 & 35.57 & 32.44 & 32.40 \\
mPLUG-DocOwl 1.5 & LLama-7B & 33.33 & 18.62 & 31.83 & 25.60 & 28.43 & 30.19 & 27.56 \\
LLaVA-Next & Qwen-72B & 32.76 & 23.67 & 28.69 & 34.60 & 23.14 & 30.02 & 28.57 \\
MiniCPM-V 2.5 & Llama3-8B & 33.23 & 16.67 & 31.67 & 20.40 & 26.00 & 28.89 & 25.59 \\
DeepSeek-VL & DeepSeek-LLM-7b-base & 27.10 & 25.44 & 26.02 & 21.60 & 35.71 & 27.63 & 27.17 \\
TextMonkey & Qwen-7B & 31.24 & 11.38 & 30.76 & 19.60 & 26.71 & 27.44 & 23.94 \\ \Gray
GPT-4o-mini & - & 29.56 & 7.33 & 31.79 & 22.00 & 24.00 & 26.32 & 22.94 \\
Qwen-VL-Chat & Qwen-7B & 27.36 & 15.00 & 27.89 & 24.29 & 27.36 & 26.13 & 24.38 \\
ShareGPT4V-13B & Vicuna-13B & 27.94 & 17.59 & 27.57 & 16.80 & 28.14 & 25.75 & 23.61 \\
LLaVA1.5-13B & Vicuna-13B & 27.52 & 17.33 & 26.25 & 17.00 & 28.66 & 25.45 & 23.35 \\
MiniGPT-v2 & Llama 2-7B-Chat & 26.78 & 19.31 & 27.05 & 14.40 & 29.43 & 25.18 & 23.39 \\
ShareGPT4V-7B & Vicuna-7B & 26.73 & 17.24 & 25.75 & 16.60 & 28.14 & 24.86 & 22.89 \\
LLaVA1.5-7B & Vicuna-7B & 26.36 & 16.67 & 25.75 & 16.60 & 28.14 & 24.64 & 22.70 \\ \bottomrule
\end{tabular}%
}
\vspace{-0.2cm}
\end{table}

Results of perception tasks and reasoning tasks are presented in Tab.~\ref{tab:res_main_perception-cn} and Tab.~\ref{tab:res_main_reasoning_cn}, respectively. The models show different performances compared to the \abbr English version. 

1) Qwen2-VL and InternVL-2 significantly outperform existing models in both perception and reasoning tasks in the Chinese version. The performance of these two models even surpasses their performance on the English version of \abbr, indicating that they jave been specifically optimized for Chinese data. 

2) There is a substantial difference in how models handle Chinese and English data, with some models performing much worse in Chinese scenarios, particularly in reasoning tasks. For instance, GPT-4o and GPT-4o-mini show a performance drop of nearly $10\%$. However, some models seem to excel in Chinese-related tasks. Notably, models based on \texttt{Llama3-8B} generally achieve strong results in both Chinese perception and reasoning tasks, such as SliME and CogVLM2. This suggests that \texttt{Llama3-8B} may be an effective LLM backbone for Chinese tasks.

\begin{table}[]
\caption{\textbf{Experimental results on the reasoning tasks of \abbr-CN.} Models are ranked according to their average performance. Rows corresponding to proprietary models are highlighted in gray for distinction.  “OCR”,  “DT”, “MO”, and “AD” each indicate a specific task domain: Optical Character Recognition in the Wild, Diagram and Table, Monitoring and  Autonomous Driving, respectively. “Avg” and “Avg-C” indicate the weighted average accuracy
and the unweighted average accuracy across
subtasks in each domain.}\label{tab:res_main_reasoning_cn}
\centering
\resizebox{0.9\textwidth}{!}{%
\begin{tabular}{llcccccc}
\toprule \Gray
\multicolumn{1}{c}{\textbf{Method}} & \multicolumn{1}{c}{\textbf{LLM}} & \multicolumn{6}{c}{\textbf{Reasoning}} \\ \cmidrule{3-8} \Gray
\multicolumn{2}{c}{\textbf{Task Split}} & \textbf{OCR} & \textbf{DT} & \textbf{MO} & \textbf{AD} & \textbf{Avg} & \textbf{Avg-C} \\ \Gray
\multicolumn{2}{c}{\# QA pairs} & 207 & 602 & 298 & 800 & 1907 & 1907 \\
InternVL-2 & InternLM2.5-7B-Chat & 44.93 & 74.92 & 38.14 & 29.00 & 46.65 & 46.75 \\ 
Qwen2-VL & Qwen2-7B & 38.16 & 72.92 & 57.00 & 33.37 & 46.46 & 50.36\\
\Gray
Claude 3.5 Sonnet & - & 74.44 & 65.79 & 31.54 & 25.12 & 44.31 & 49.22 \\
SliME-8B & LLama3-8B & 44.44 & 70.93 & 30.54 & 29.13 & 44.21 & 43.76 \\
InternVL-Chat-V1-5 & InternLM2-Chat-20B & 48.79 & 67.11 & 30.20 & 29.88 & 43.74 & 44.00 \\
CogVLM2-llama3-Chat & LLama3-8B & 33.81 & 65.24 & 37.25 & 29.00 & 42.25 & 41.33 \\
YI-VL-34B & Yi-34B-Chat & 37.68 & 61.46 & 33.22 & 29.75 & 41.16 & 40.53 \\
Monkey & Qwen-7B & 43.96 & 56.81 & 32.55 & 28.38 & 39.70 & 40.43 \\
Mini-Gemini-7B-HD & Vicuna-7B-v1.5 & 28.50 & 68.61 & 25.50 & 24.00 & 38.80 & 36.65 \\
Mini-Gemini-34B-HD & Nous-Hermes-2-Yi-34B & 29.95 & 60.47 & 25.50 & 29.63 & 38.75 & 36.39 \\
LLaVA-Next & LLama3-8B & 14.93 & 58.14 & 29.53 & 28.25 & 36.44 & 32.71 \\
Cambrian-1-8B & LLama3-8B-Instruct & 27.54 & 47.51 & 31.21 & 31.25 & 35.97 & 34.38 \\
SliME-13B & Vicuna-13B & 44.93 & 49.17 & 30.54 & 23.87 & 35.18 & 37.13 \\
LLaVA-Next & Qwen-72B & 24.64 & 40.87 & 27.52 & 28.12 & 31.67 & 30.29 \\
InternLM-XComposer2.5 & InternLM2-7B & 18.36 & 40.70 & 16.78 & 30.00 & 30.05 & 26.46 \\ \Gray
GPT-4o & - & 33.81 & 39.87 & 20.81 & 22.75 & 29.05 & 29.31 \\
DeepSeek-VL & DeepSeek-LLM-7b-base & 36.23 & 23.59 & 25.50 & 29.25 & 27.63 & 28.64 \\
Cambrian-1-34B & Hermes2-Yi-34B & 21.74 & 31.40 & 23.49 & 25.12 & 26.48 & 25.44 \\
LLaVA1.5-13B & Vicuna-13B & 36.23 & 27.08 & 25.84 & 23.25 & 26.27 & 28.10 \\
ShareGPT4V-13B & Vicuna-13B & 35.75 & 27.91 & 24.83 & 22.88 & 26.17 & 27.84 \\
MiniCPM-V 2.5 & LLama3-8B & 36.23 & 29.90 & 16.44 & 23.87 & 25.95 & 26.61 \\
LLaVA1.5-7B & Vicuna-7B & 33.33 & 25.91 & 25.17 & 23.25 & 25.48 & 26.92 \\
ShareGPT4V-7B & Vicuna-7B & 33.33 & 25.91 & 24.83 & 23.25 & 25.43 & 26.83 \\
Qwen-VL-Chat & Qwen-7B & 30.92 & 36.41 & 13.42 & 19.88 & 25.29 & 25.16 \\ \Gray
GPT-4o-mini & - & 27.88 & 27.08 & 14.77 & 26.87 & 25.16 & 24.15 \\
MiniGPT-v2 & Llama 2-7B-Chat & 34.30 & 28.57 & 19.80 & 22.13 & 25.12 & 26.20 \\
mPLUG-DocOwl 1.5 & LLama-7B & 37.68 & 24.42 & 19.80 & 22.38 & 24.28 & 26.07 \\
TextMonkey & Qwen-7B & 27.53 & 31.07 & 12.08 & 22.50 & 24.12 & 23.29 \\ \Gray
Gemini-1.5-pro & - & 5.30 & 5.32 & 14.77 & 15.67 & 11.14 & 10.26 \\ \bottomrule
\end{tabular}%
}
\end{table}

\subsection{Fine-grained Analysis and Findings}

\begin{figure}
    \centering
    \includegraphics[width=\linewidth]{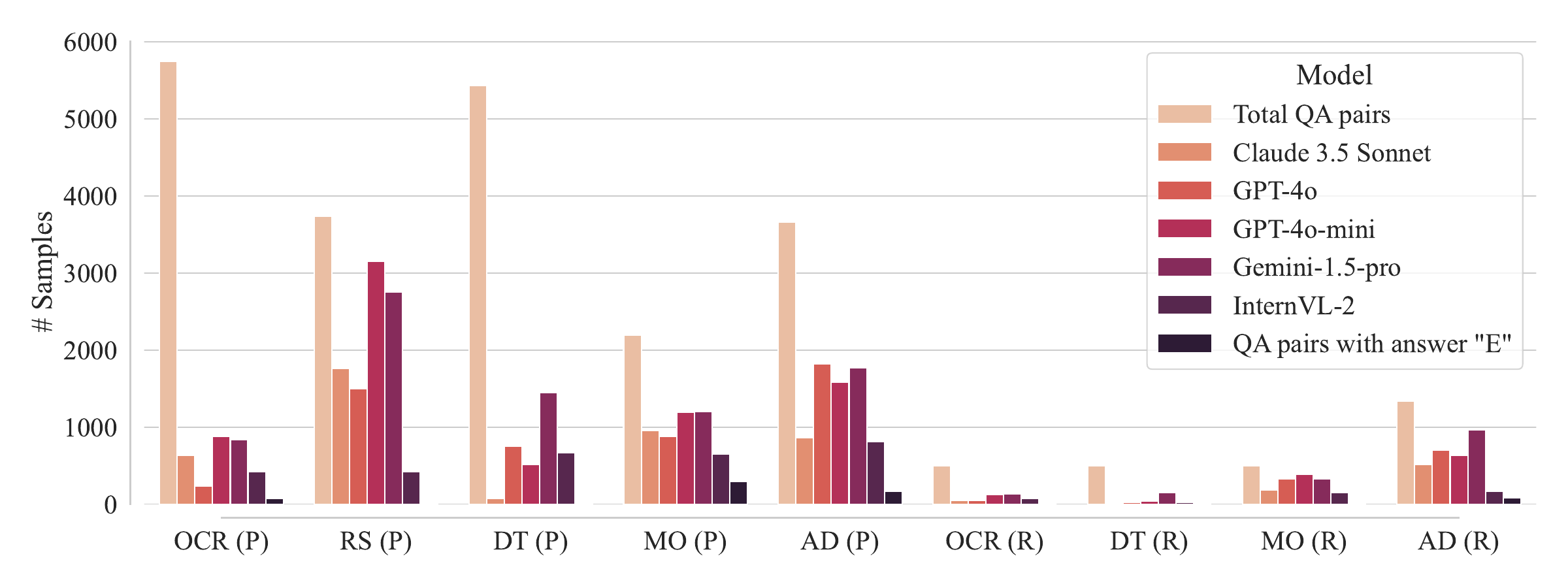}
\caption{\textbf{Frequency of outputting answer ``E'' for different models} across various domains. The notation in parentheses indicates the task type: P for perception and R for reasoning. The total QA pairs and those with answer ``E'' are also presented for comparison.}
    \label{fig:close_e}
\end{figure}

\textbf{Existing Models Still Lacking in Image Detail Perception.} Fig.~\ref{fig:close_e} displays the frequency with which various models choose “E” as their answer. We compare $4$ close-source models with the top-level open-source model, InternVL-2. During our annotation process, the frequency of “E” answers does not exceed $5\%$ of the overall data, meaning it represents only a small portion of the total QA pairs. However, nearly all models show a much higher frequency of “E” outputs than the actual number of “E” instances present in our benchmark. This indicates that most models' visual perception modules fail to identify the objects in the images corresponding to our questions. 

\textbf{Limitations of MLLMs in Understanding Dynamic Information.} 
In combination with the results from autonomous driving and monitoring tasks, we observe that MLLMs exhibit significant deficiencies in understanding, predicting, and reasoning about the dynamic information of objects, such as predicting the steering of a car. Although the input to these models is a single frame image rather than a video, there remains a considerable gap between their performance and that of humans. Therefore, it seems that these MLLMs are still far from having the capability to be world models.

\textbf{Computation Efficiency.} 
There is a significant disparity in computation efficiency among different models when processing high-resolution images. For example, using models similar to LLMs (e.g., Vicuna-13B), the computational requirements for handling images exceeding $1024$$\times$$1024$ resolution are as follows: LLaVA1.5 requires $16.37$ TFLOPS, SliME requires $40.82$ TFLOPS, while LLaVA-Next and Mini-Gemini-HD require $78.37$ and $87.59$ TFLOPS, respectively. LLaVA-Next and SliME employ dynamic chunking and encoding of images, while Mini-Gemini-HD uses a higher-resolution vision encoder and significantly increases the number of vision tokens, resulting in a computation cost approximately $5$ times that of LLaVA1.5. 
Additionally, existing methods have inherent limitations in handling high-resolution images. For example, Mini-Gemini-HD resizes images larger than $672$$\times$$672$ to this size, causing a loss of more details. Moreover, we observe interesting phenomena in closed-source models regarding image resolution. For instance, GPT-4o-mini uses over $10,000$ tokens for some large images, which is about $10$ times more than other closed-source models, although its performance does not significantly surpass other models. Overall, we currently lack methods that can efficiently handle higher resolution images with lower computational overhead.

\begin{figure*}[t]
\subfigure[{Claude 3.5 Sonnet}]{
\begin{minipage}[t]{0.33\linewidth}
\centering
 \includegraphics[width=\linewidth]{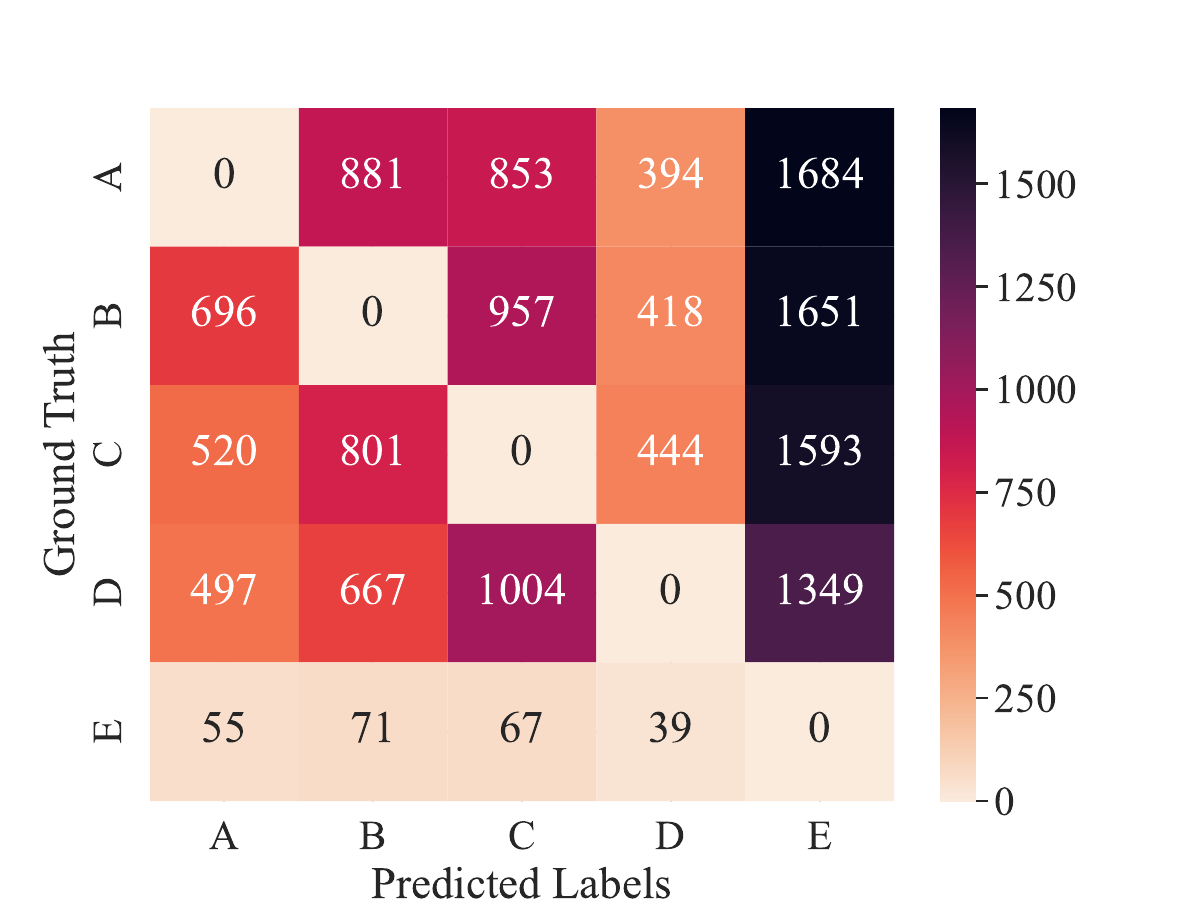}
    \label{fig:claude}
\end{minipage}%
}%
\subfigure[{GPT-4o}]{
\begin{minipage}[t]{0.33\linewidth}
 \includegraphics[width=\linewidth]{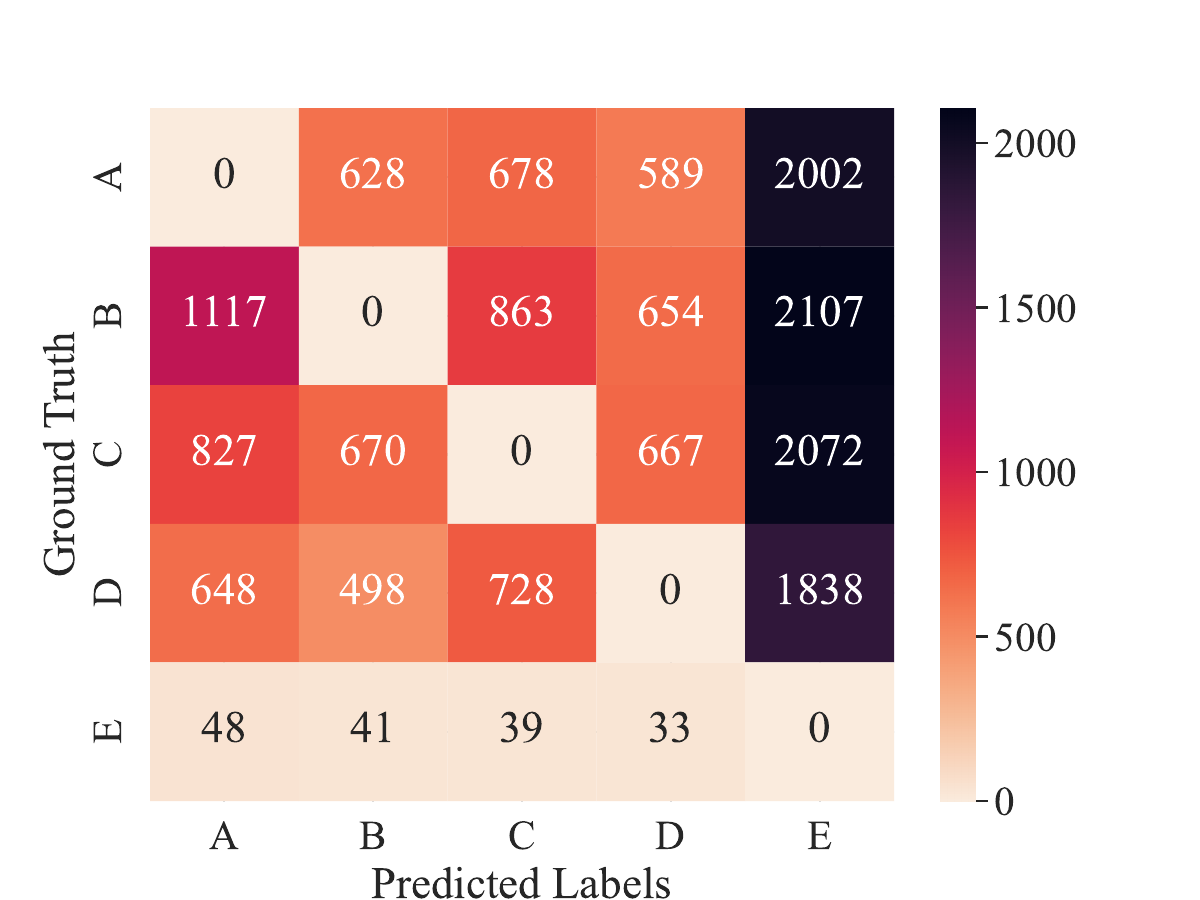}
    \label{fig:gpt4o}
\end{minipage}%
}%
\subfigure[{Cambrian-1-34B}]{
\begin{minipage}[t]{0.33\linewidth}
\centering
 \includegraphics[width=\linewidth]{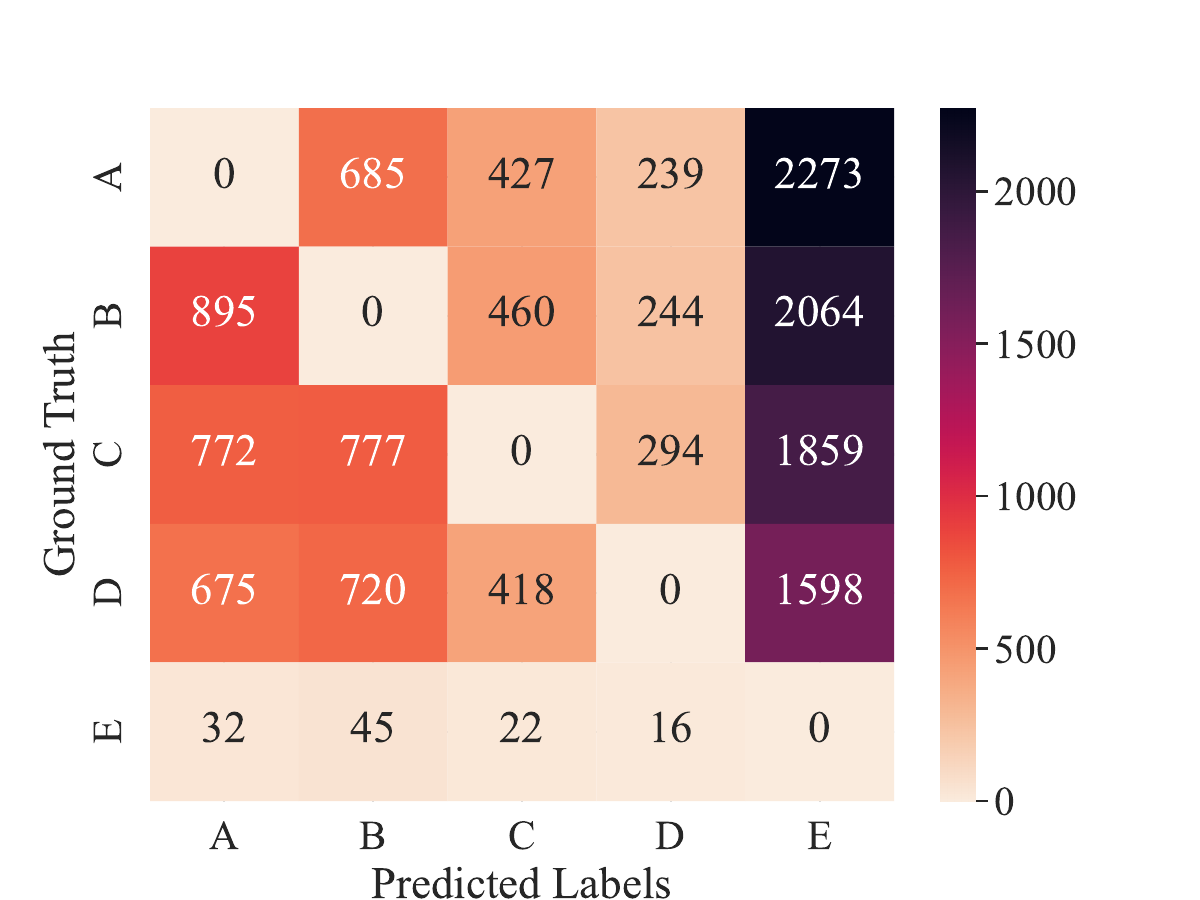}
    \label{fig:cambrian}
\end{minipage}%
}%

\subfigure[{Monkey}]{
\begin{minipage}[t]{0.33\linewidth}
 \includegraphics[width=\linewidth]{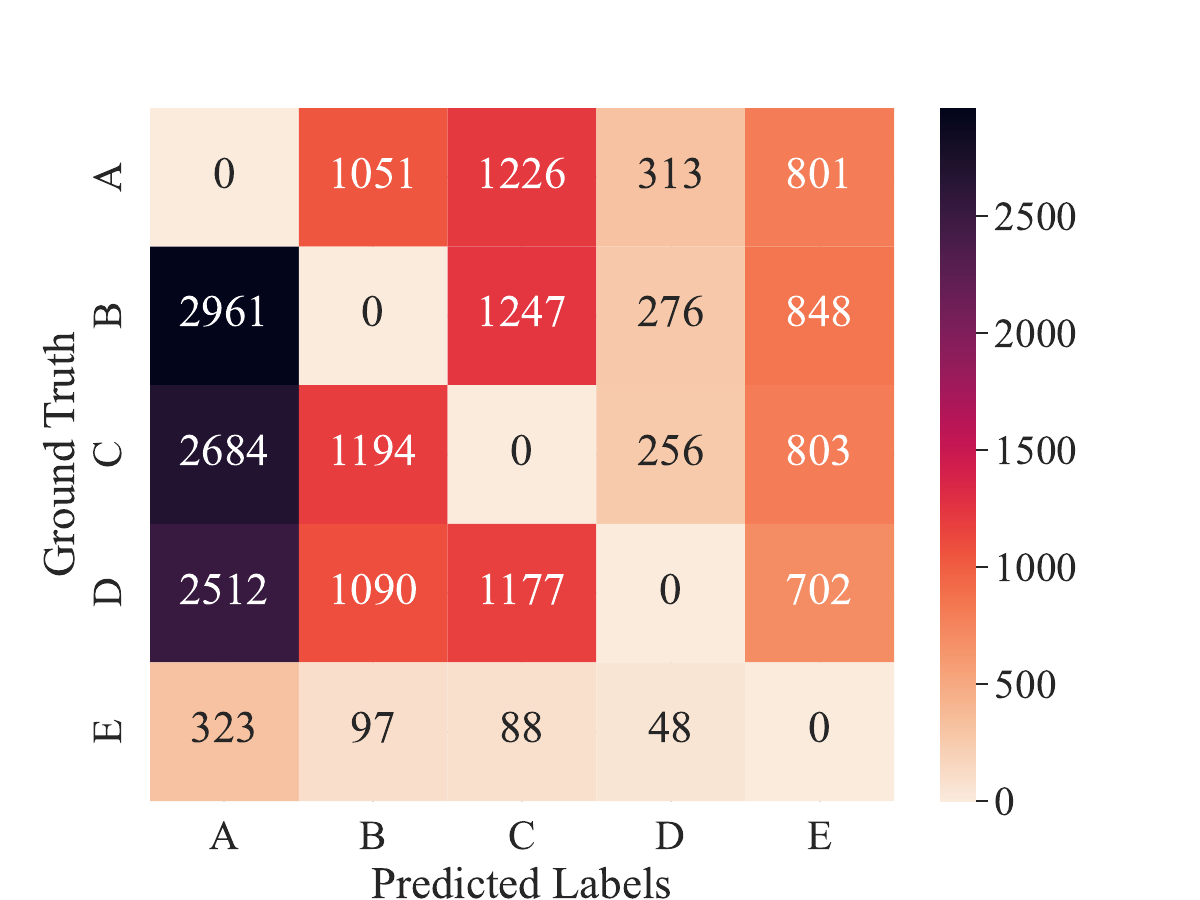}
    \label{fig:monkey}
\end{minipage}%
}%
\subfigure[{mPLUG-DocOwl 1.5}]{
\begin{minipage}[t]{0.33\linewidth}
 \includegraphics[width=\linewidth]{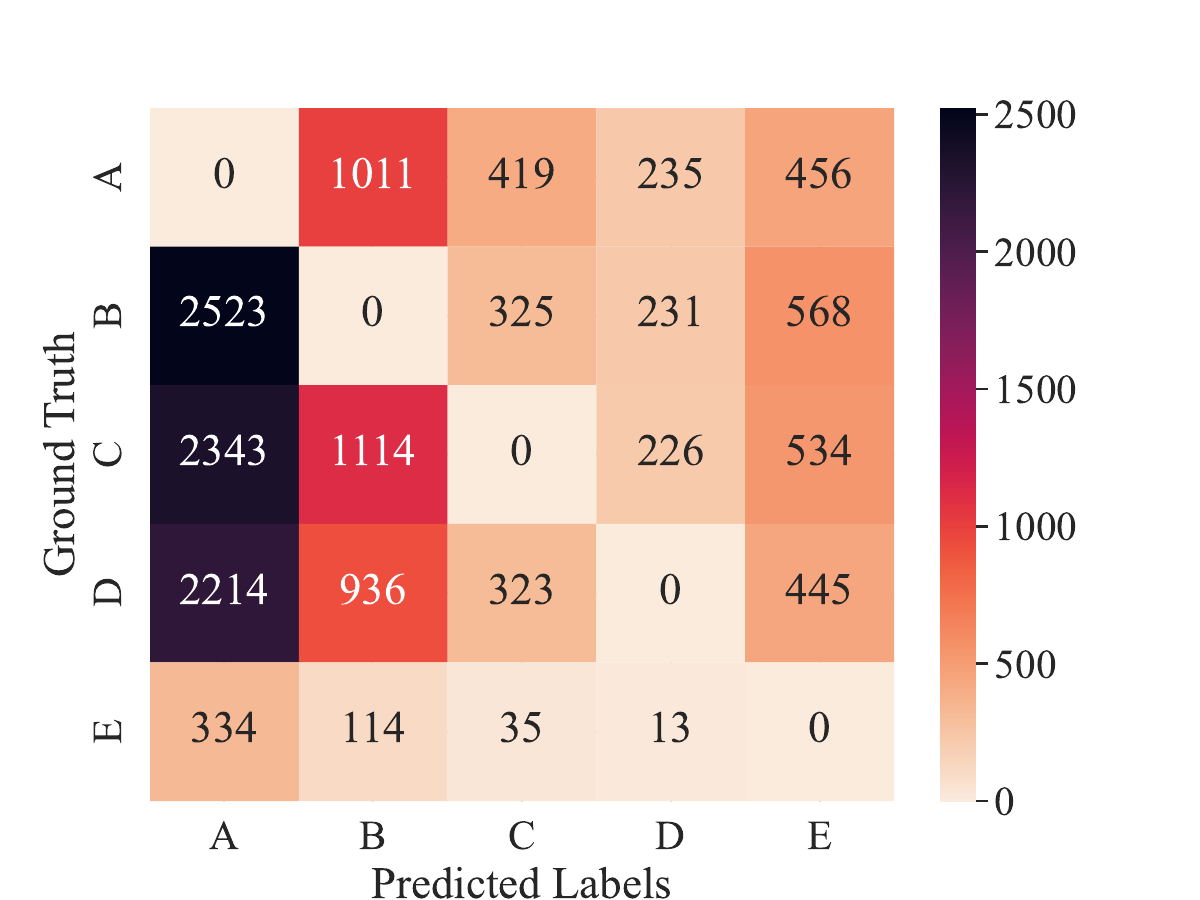}
    \label{fig:docowl}
\end{minipage}%
}%
\subfigure[{InternVL-2}]{
\begin{minipage}[t]{0.33\linewidth}
 \includegraphics[width=\linewidth]{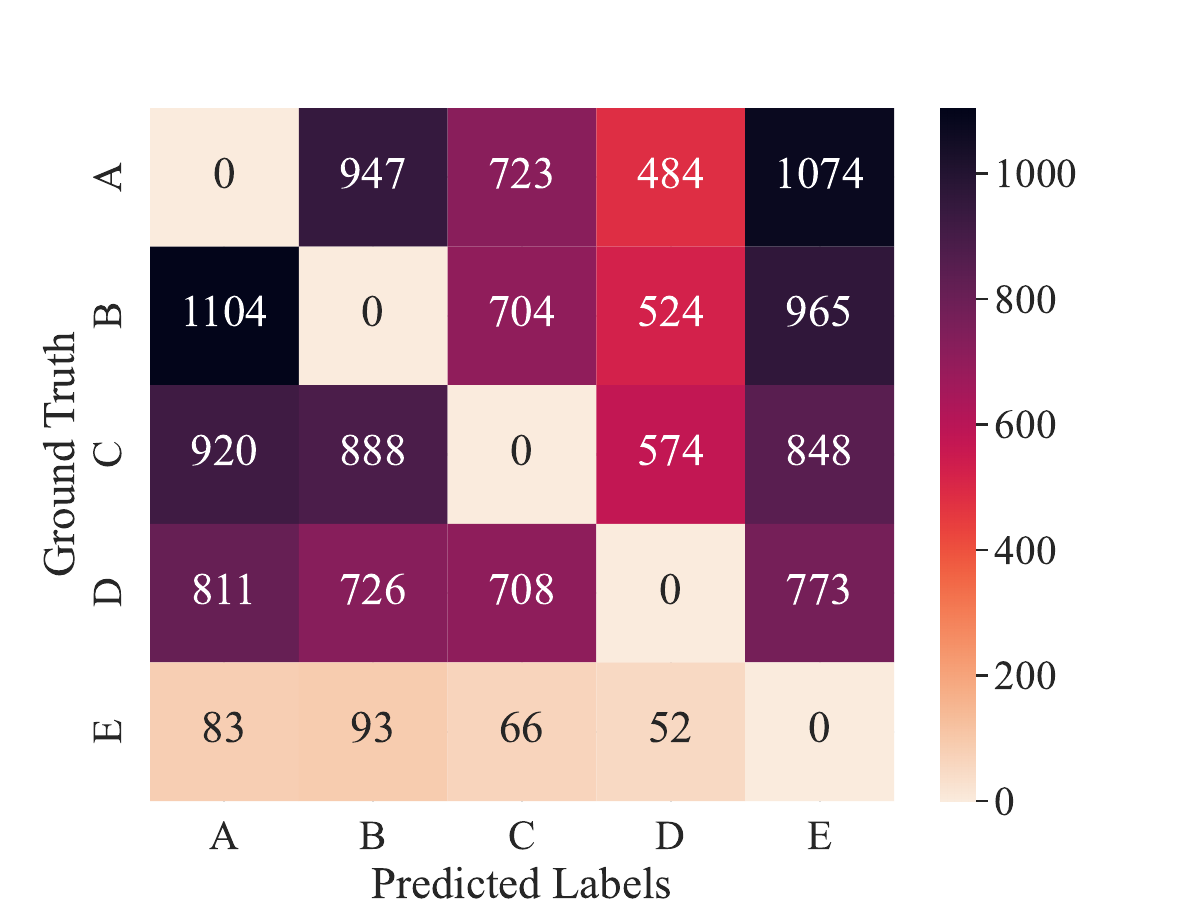}
    \label{fig:internvl}
\end{minipage}%
}%

\centering
\vspace{-0.2cm}
\caption{\textbf{Distribution of incorrec choices.} The matrix reveals distinct response behaviors among different MLLMs. Larger models tend to select the safer option ``E'', while smaller models exhibit a bias toward the first option ``A''. InternVL-2, however, shows a unique uniform error distribution.}
\label{fig:con_mat}
\end{figure*}
\textbf{Analyzing Incorrect Choices.}
We investigate the distribution of incorrect choices across a range of models, as shown in Fig.~\ref{fig:con_mat}. 
We can see that MLLMs show different response strategies when dealing with questions imbued with uncertainty. 
Larger models generally adopt a more conservative approach, often opting for the safer response ``E'', as illustrated from Fig.~\labelcref{fig:claude,fig:gpt4o,fig:cambrian}.
In contrast, smaller MLLMs often lean towards the first option—usually option ``A''—in similar situations, as shown in Fig.~\labelcref{fig:monkey,fig:docowl}. 
Notably, InternVL-2 presents a unique distribution of incorrect choices that is remarkably uniform, which may account for its exceptional performance in our evaluation.

\textbf{Instruction Following Abilities.} As described in Sec.~\ref{sec:prompt}, our prompts specify that the model should directly select and output a single answer. In this regard, closed-source models generally perform better, with outputs being more concise and directly aligned with the instructions. However, we have observed that some open-source models do not strictly adhere to our queries and generate a significant amount of additional analysis. Sometimes, they even produce outputs that are excessively verbose, continuing until the token count reaches the predefined maximum limit. 
This indicates that the open-source models have a lot of room for optimization in the ability of instruction following.

For detailed results and analysis of all domains and subtasks, please refer to Appendix Sec.~\ref{app:sec_res}.

\section{Conclusion}
In this paper, we have introduced \abbr, a comprehensive benchmark designed to address key limitations in existing evaluations of MLLMs, such as data scale, annotation quality, and task difficulty. As the largest purely human-annotated dataset with the highest resolution to date, \abbr benefits from the participation of $32$ annotators, ensuring high data quality and minimal individual bias. 
Most QA pairs focus on real-world scenarios, such as autonomous driving and video surveillance, which have significant applicability.
Furthermore, we propose \abbr-CN, a benchmark specifically focused on Chinese scenarios, ensuring that all images and questions are relevant to Chinese contexts. 
Our evaluation of a wide range of models reveals significant performance gaps, highlighting the current models' shortcomings in complex image perception and underscoring, and the need for further advancements.

\bibliography{neurips_2024}
\bibliographystyle{plain}

\clearpage
\newpage
\appendix

\begin{center}
{\LARGE \textbf{\abbr\\ $\;$ \\ ————Appendix————}}
\end{center}

{
  \hypersetup{hidelinks}
  \tableofcontents
  \noindent\hrulefill
}

\newpage
\addtocontents{toc}{\protect\setcounter{tocdepth}{2}} 

\section{Data Collection and Task Split}\label{sec:app_data_collection}

\subsection{OCR in the Wild}

\textbf{Data Characteristics.} The data is from real-world street scenes and high-resolution images of product advertisements. Text is dense or difficult to detect and requires careful observation to be identified.

\subsubsection{Data Sources and Annotation Process}

 \textbf{Data Sources.} We manually select images with complex scenes and recognizable text information from existing high-resolution datasets for our test images. The open-source datasets used include DIV2K and Flickr2K~\citep{Agustsson_2017_CVPR_Workshops}, which offer paired high-resolution RGB images and their corresponding downscaled low-resolution RGB images by a factor of two. In our approach, we exclusively utilize high-resolution images, selecting and preserving images with complex scenes and contexts. Additionally, we include the LIU4K~\citep{Liu4K} dataset, which contains $2,000$ images with resolutions of at least $3$K, most ranging from 4K to $6$K. This dataset provides abundant materials for testing and evaluating performance on $4$K/$8$K display devices, featuring diverse and complex low-level signal distributions and backgrounds. We also incorporate two large-scale Ultra-High-Definition datasets, UHD4K and UHD8K~\citep{zhang2021benchmarking}, which collectively contain $23,000$ images. These datasets cater to various low-level image enhancement tasks, including image super-resolution (SR), image deraining (Derain), low-light image enhancement (LLIE), and image reflection removal (IRR). Finally, we use HQ-50K~\citep{yang2023hq}, a large-scale, high-quality image restoration dataset containing $50,000$ high-quality images. HQ-50K stands out for its large scale, high resolution, varying compression rates, rich texture details, and semantic diversity.

\textbf{Annotation.} 20 volunteers annotate the question and answer pairs. 3 experts are tasked with checking and correcting the annotations to ensure quality.

\subsubsection{Evaluation Dimensions and Benchmark Statistics} 

\begin{figure*}[t]
\subfigure[{Phone and Address Perception.}]{
\begin{minipage}[t]{0.45\linewidth}
\centering
 \includegraphics[width=\linewidth]{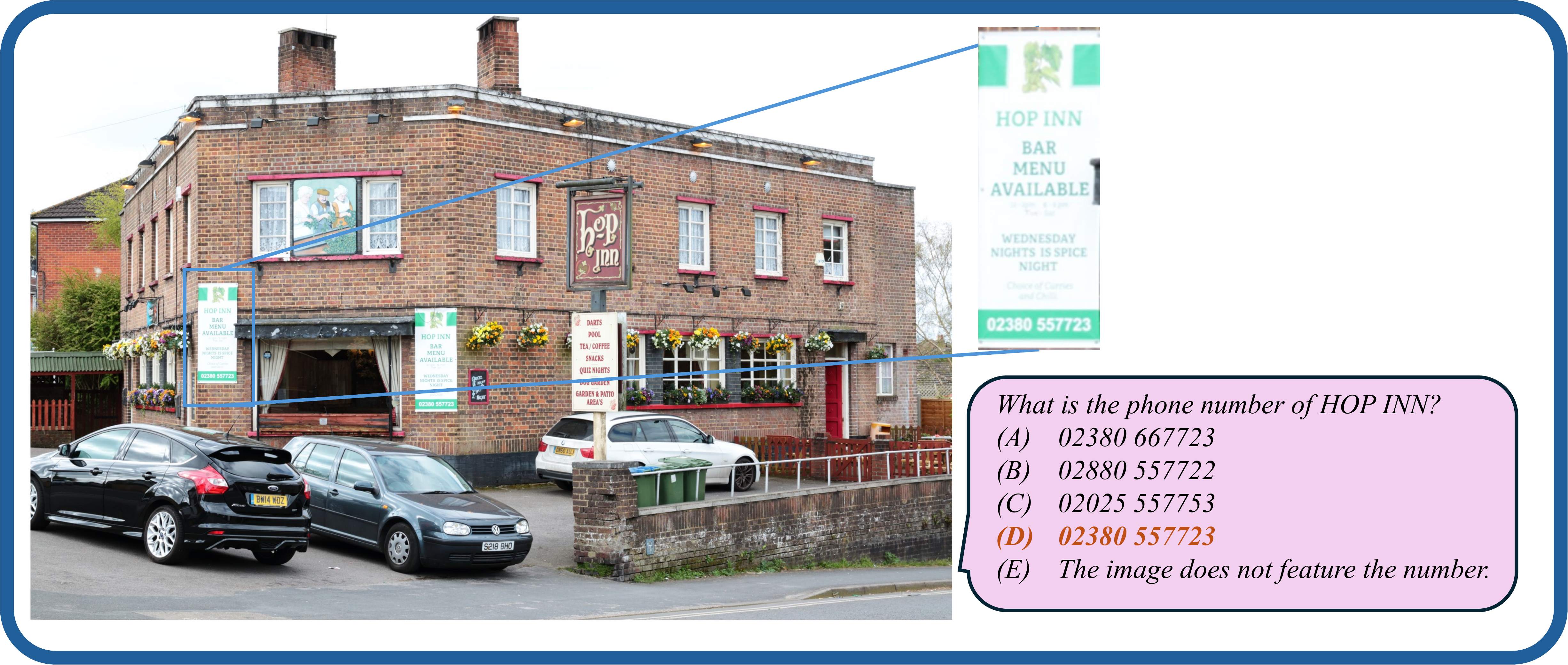}
    \label{fig:phone}
\end{minipage}%
}%
\subfigure[{Product and Advertisement.}]{
\begin{minipage}[t]{0.54\linewidth}
 \includegraphics[width=\linewidth]{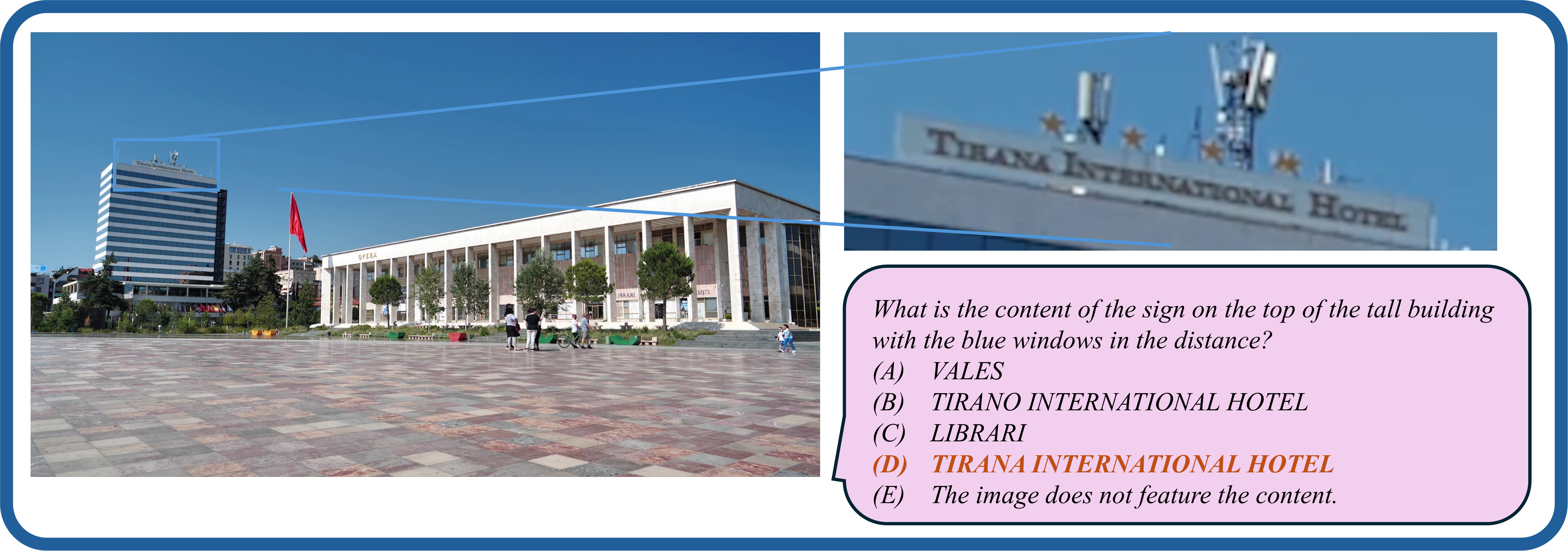}
    \label{fig:adver}
\end{minipage}%
}%

\subfigure[{Human/Car License.}]{
\begin{minipage}[t]{0.45\linewidth}
\centering
 \includegraphics[width=\linewidth]{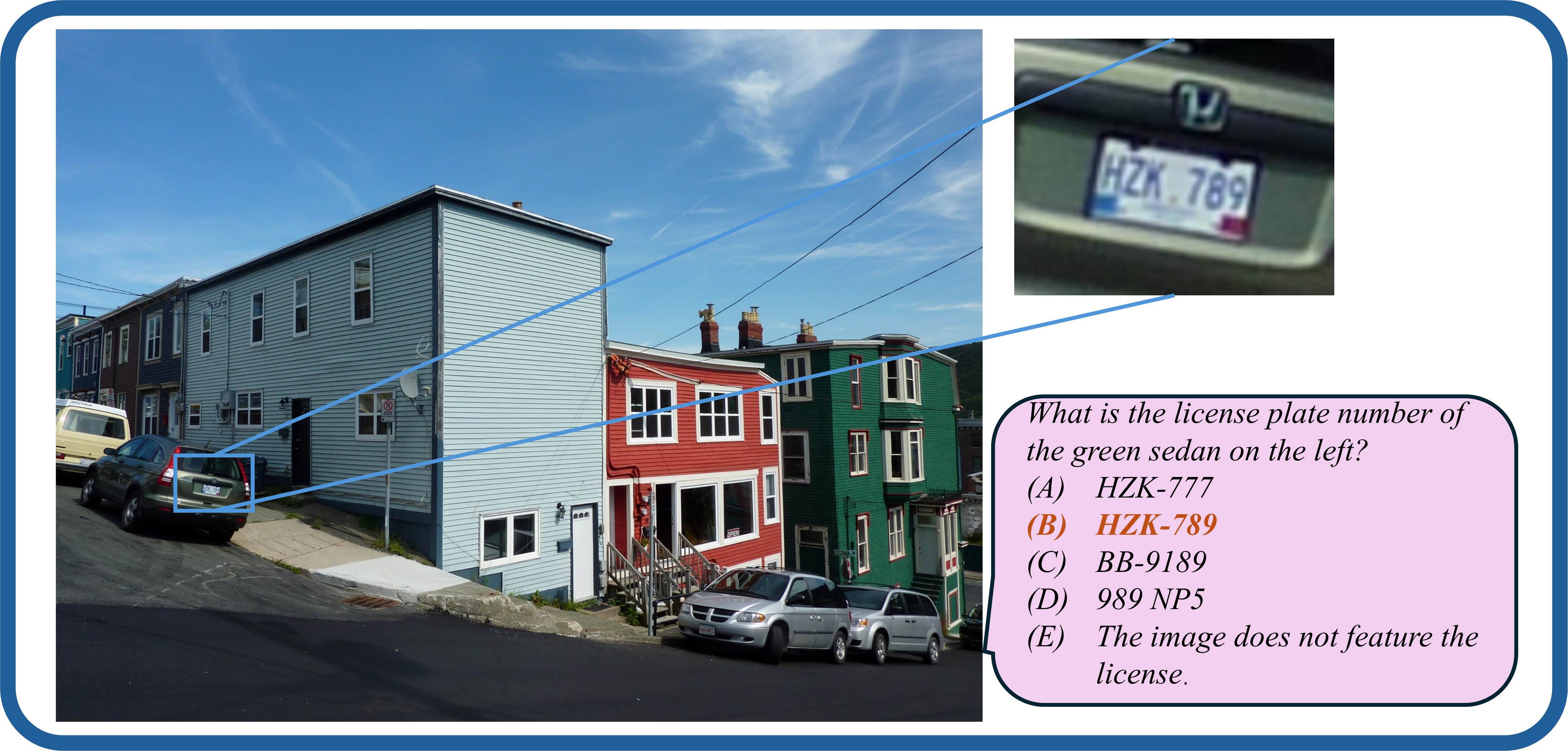}
    \label{fig:license}
\end{minipage}%
}%
\subfigure[{Other kind of small Text on Signals or Indicators.}]{
\begin{minipage}[t]{0.54\linewidth}
 \includegraphics[width=\linewidth]{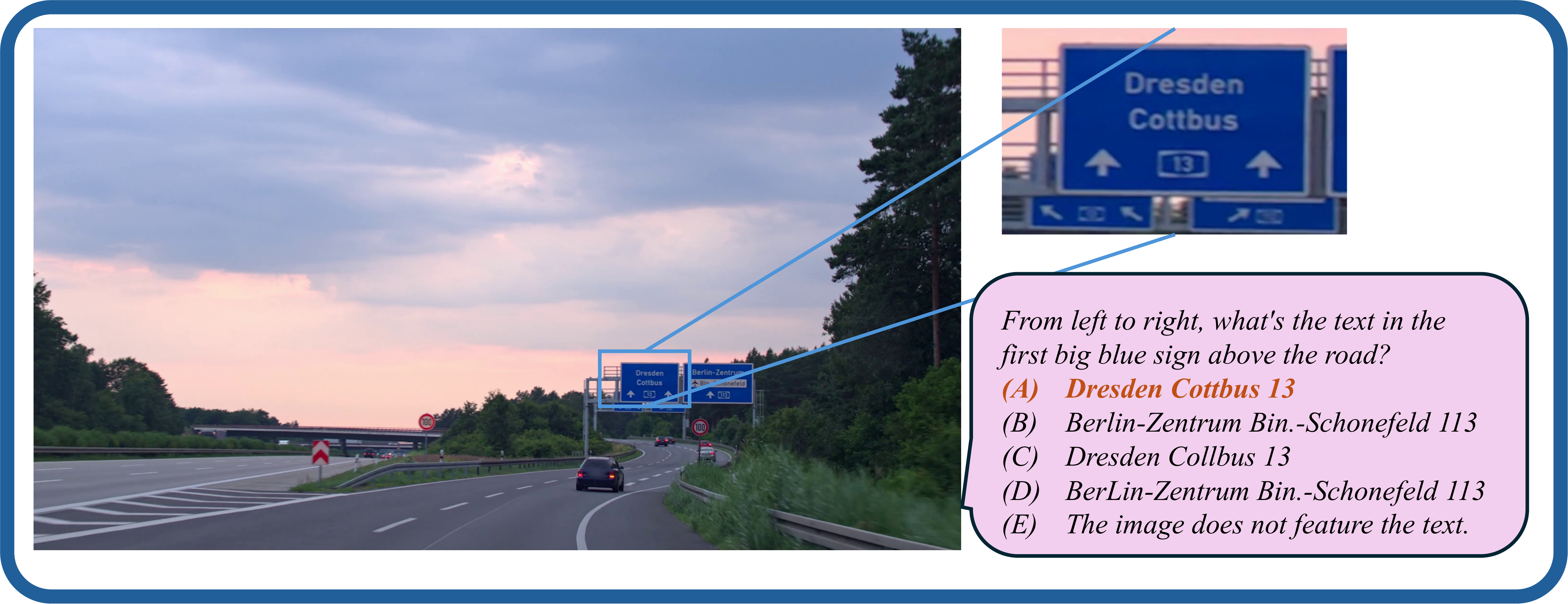}
    \label{fig:text}
\end{minipage}%
}%

\subfigure[{Book, Map and Poster.}]{
\begin{minipage}[t]{0.54\linewidth}
 \includegraphics[width=\linewidth]{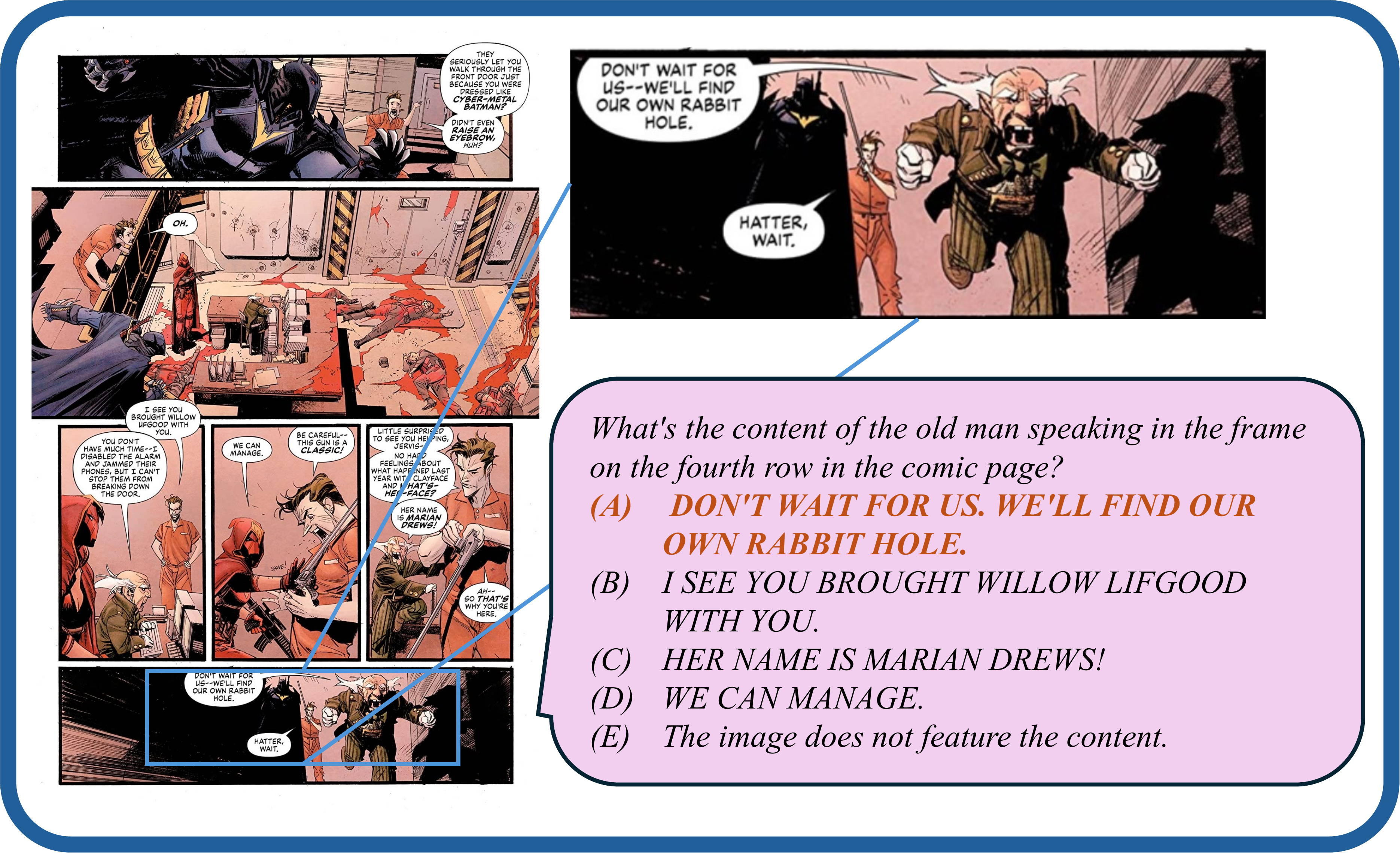}
    \label{fig:book}
\end{minipage}%
}%
\centering
\vspace{-0.2cm}
\caption{\textbf{Data Examples for Perception Tasks in OCR in the Wild}}
\label{fig:ocr_cc_perception}
\end{figure*}

The evaluation of models in real-world complex scenes involves their ability to recognize and understand text, enabling us to ascertain their capacity to comprehend and process textual information within visual content, thereby enhancing the overall practicality and reliability of intelligent systems. Specifically, Optical Character Recognition (OCR) in complex contexts comprises five perception tasks and two reasoning tasks. For perception tasks,

1. \textbf{Contact information and addresses (Fig.~\ref{fig:phone}).} Recognizing telephone numbers, names of countries/cities/streets, and buildings ($469$ images and $577$ QA pairs).

2. \textbf{Product and Advertisement Perception (Fig.~\ref{fig:adver}).} Identifying product names/prices or advertisements of shops or brands ($803$ images and $1,588$ QA pairs).

3. \textbf{Identity Information Perception (Fig.~\ref{fig:license}).} Recognizing license numbers or ID cards of cars/humans ($852$ QA pairs).

4. \textbf{Other kind of Small Text on Signals or Indicators Perception (Fig.~\ref{fig:text}).} Recognizing small text on indicators, signals, and similar objects ($626$ images and $1,198$ QA pairs).

5. \textbf{Book, Map and Poster Perception  (Fig.~\ref{fig:book}).} Recognizing dialogues/information on posters and specific locations involving a country/region on maps ($785$ images and $1,555$ QA pairs).

\begin{figure*}[t]
\subfigure[{Scene Recognition.}]{
\begin{minipage}[t]{0.54\linewidth}
\centering
 \includegraphics[width=\linewidth]{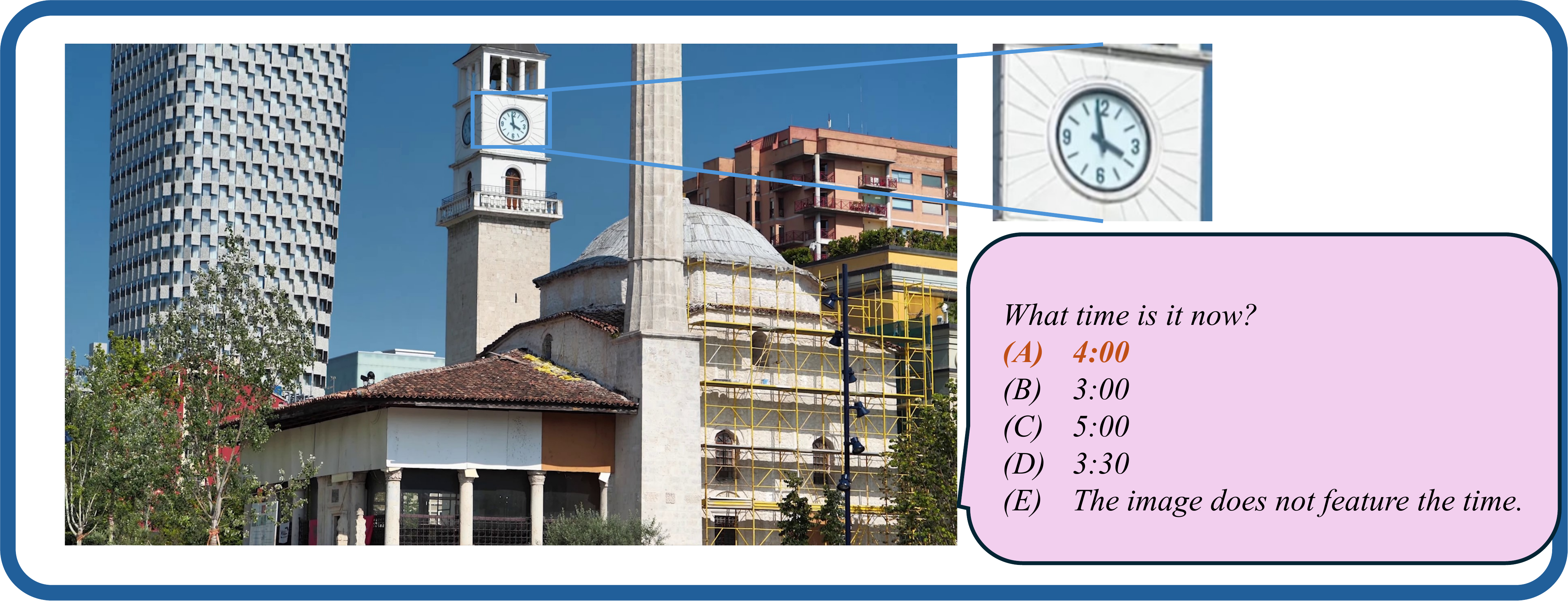}
    \label{fig:reasoning_scene}
\end{minipage}%
}%
\subfigure[{Characters Understanding.}]{
\begin{minipage}[t]{0.45\linewidth}
 \includegraphics[width=\linewidth]{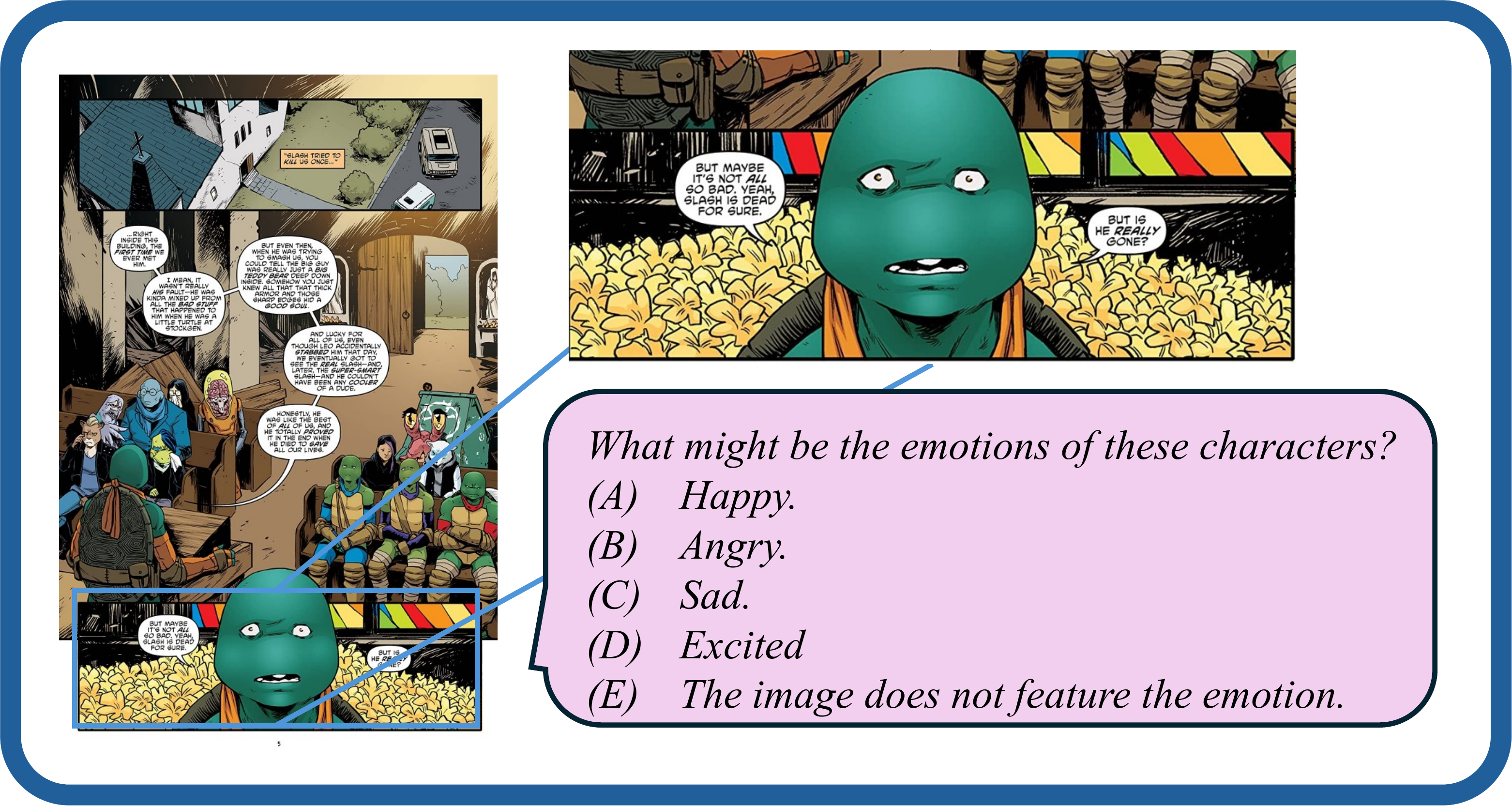}
    \label{fig:reasoning_charac}
\end{minipage}%
}%

\centering
\vspace{-0.2cm}
\caption{\textbf{Data Examples for Reasoning Tasks in OCR in the Wild}}
\label{fig:ocr_cc_reasoning}
\end{figure*}

Additionally, our two reasoning tasks include:

1. \textbf{Scene Recognition (Fig.~\ref{fig:reasoning_scene}).} Understanding the meaning of scenes in images, such as predicting the outcome of a game based on the scoreboard or what might happen in the future based on the scene, inferring the time by looking at a clock, or calculating object prices ($250$ images and $250$ QA pairs).

2. \textbf{Characters Understanding (Fig.~\ref{fig:reasoning_charac}).} Understanding the pertinent characteristics of characters in a poster or comic, including their relationships, emotions, intentions, or quantities ($250$ images and $250$ QA pairs).

Note that although we have $3,293$ unique images, some tasks use overlapping image sets, so the total number of images listed in all the tasks is not exactly $3,293$.

\subsection{Diagram and Table}
\textbf{Data Characteristics.} Diagrams and tables with rich content present significant challenges for rapid localization and analysis, even for human researchers. These tasks demand a high level of perceptual capability.

\subsubsection{Data Sources and Annotation Process}
Although there are existing datasets for evaluating diagrams and tables, such as ChartQA~\citep{masry2022chartqa} and some open-source scientific chart data like Arxiv QA~\citep{li2024multimodal}, we observe that these datasets often have relatively low image resolutions and limited content richness. Consequently, they are relatively easy for humans to interpret quickly, which does not align with the design goals of our benchmark. To address this, we source complex diagram data from the internet, such as detailed financial reports with large charts. Analyzing these large charts poses significant perceptual challenges, even for humans, and thus better aligns with the objectives of our benchmark.

\textbf{Annotation.} 20 volunteers are involved to generate question and answer pairs for the perception task. Additionally, one expert researcher is responsible for generating reasoning annotations. To ensure high-quality annotations, three experts are assigned to review and correct the annotations.

\subsubsection{Evaluation Dimensions and Benchmark Statistics} 
\begin{figure*}[t]
\centering
\subfigure[{Table Perception.}]{
\begin{minipage}[t]{0.5\linewidth}
\centering
 \includegraphics[width=\linewidth]{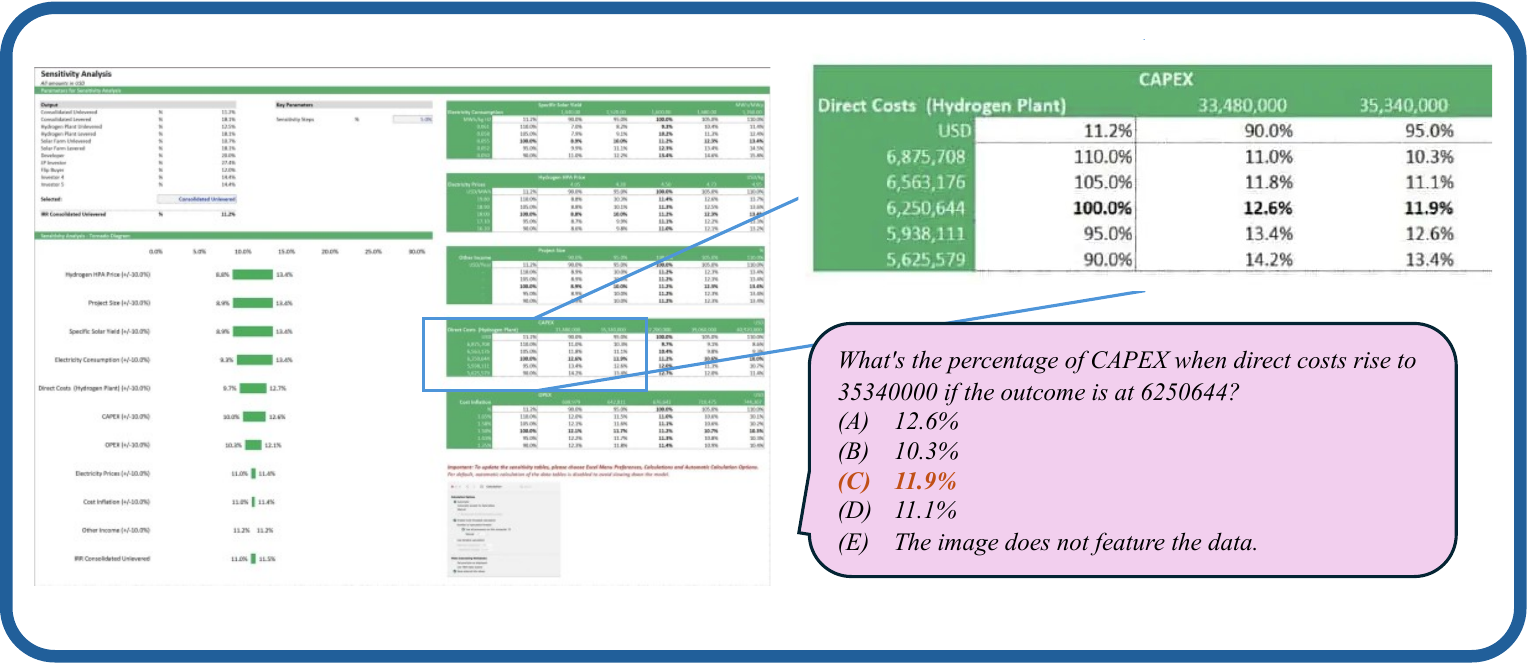}
 \label{fig:table_p}
\end{minipage}%
}%
\subfigure[{Diagram Perception.}]{
\begin{minipage}[t]{0.5\linewidth}
\centering
 \includegraphics[width=\linewidth]{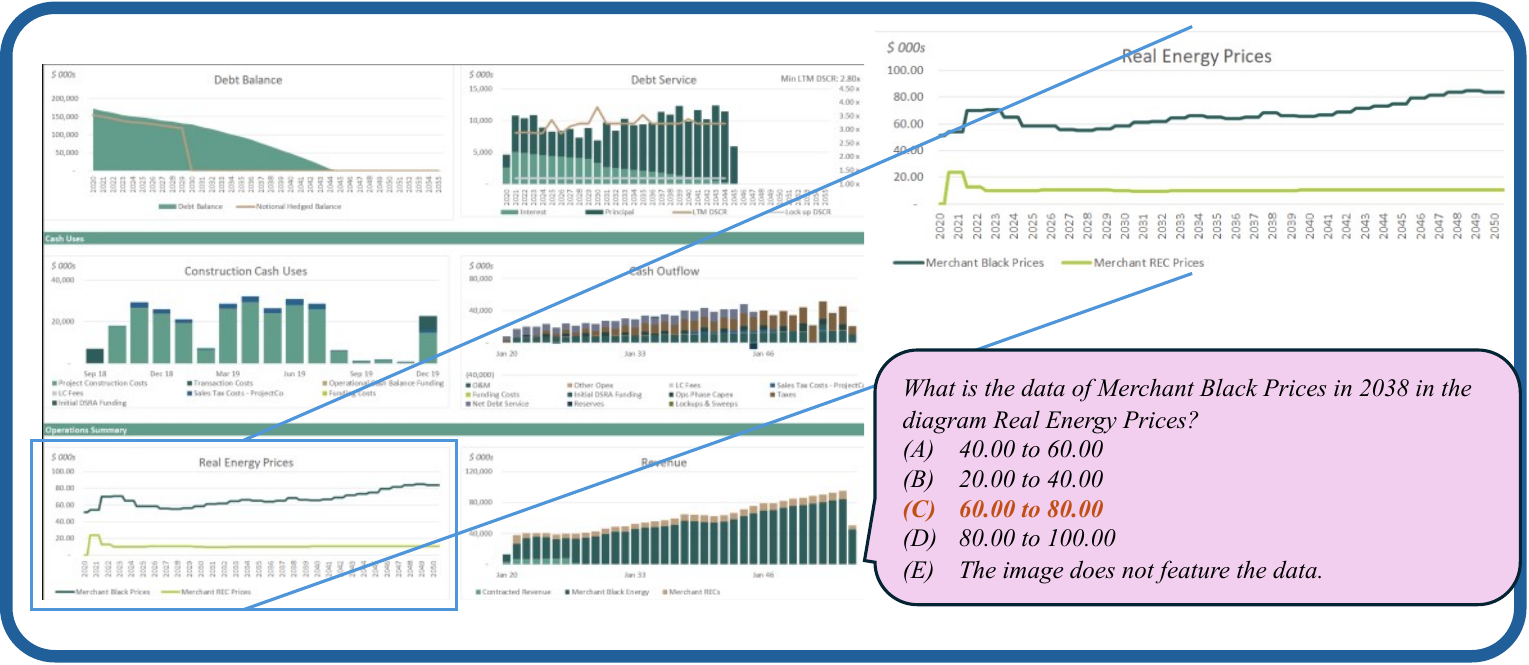}
    \label{fig:diagram_p}
\end{minipage}%
}%

\subfigure[{Table Reasoning.}]{
\begin{minipage}[t]{0.5\linewidth}
\centering
 \includegraphics[width=\linewidth]{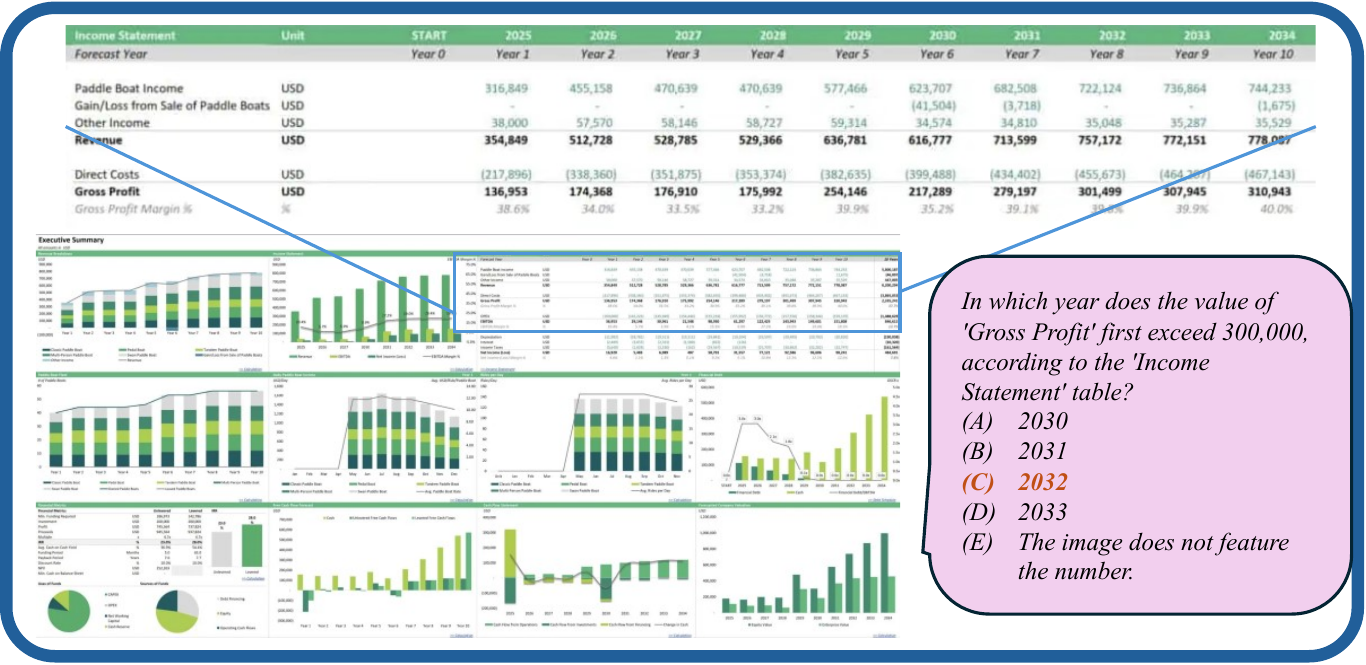}
    \label{fig:table_r}
\end{minipage}%
}%
\subfigure[Diagram Reasoing]{
\begin{minipage}[t]{0.5\linewidth}
\centering
 \includegraphics[width=\linewidth]{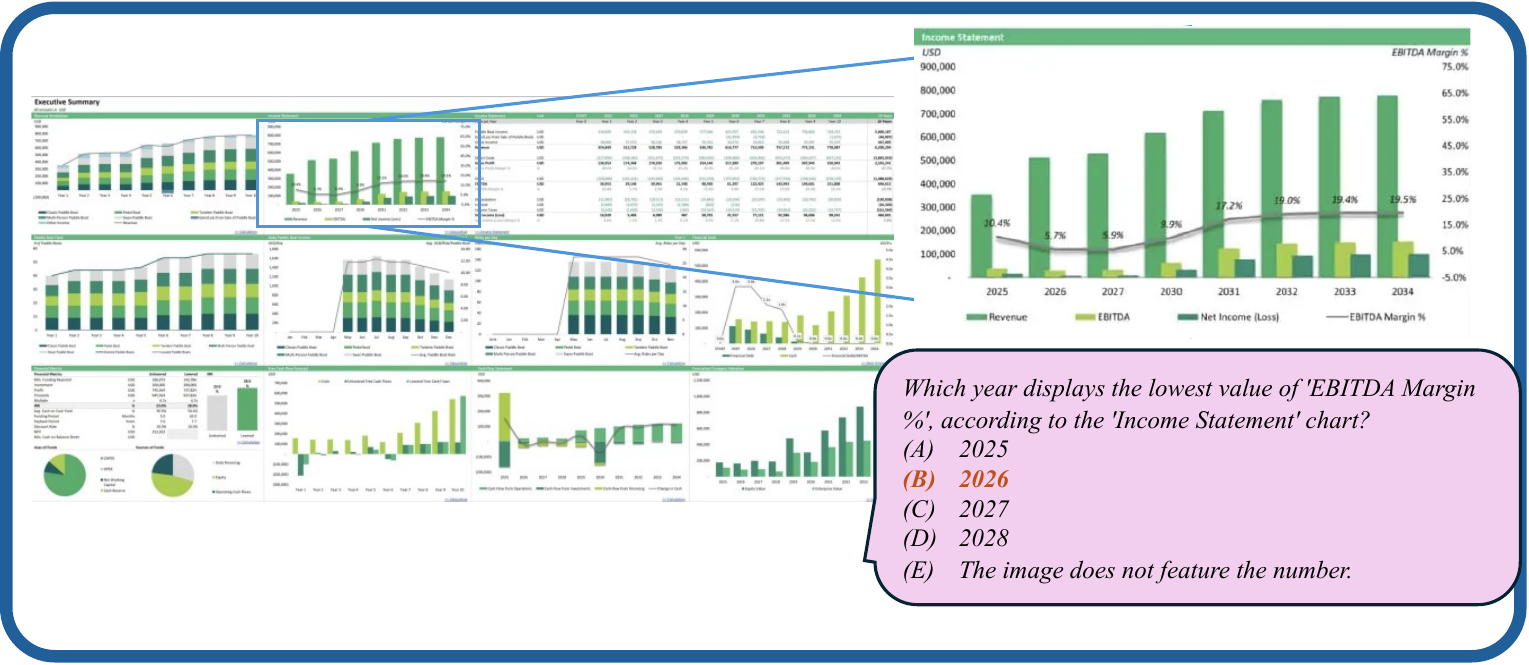}
    \label{fig:diagram_r}
\end{minipage}%
}%
\centering
\vspace{-0.2cm}
\caption{\textbf{Data Examples for Diagram and Table Tasks }}
\label{fig:tab_and_diagram}
\end{figure*}

The ability of multimodal models to perceive and understand diagram and table data has long been a focus of research. In our Diagram and Table domain, we have elevated the difficulty level to a point where even humans find it challenging to solve easily. We have collected $2,570$ images and $5,933$ annotations, categorizing the annotations into the following four types:

1. \textbf{Table Perception (Fig.~\ref{fig:table_p}).} Identifying specific elements within a table by using the given table name, horizontal axis coordinates, and related location information to determine the value of elements in specific positions ($4,018$ QA pairs).

2. \textbf{Diagram Perception (Fig.~\ref{fig:diagram_p}).} Identifying specific elements within a diagram by using the provided legend or title, along with specific location information, to determine the value of elements or the intervals they belong to ($1,415$ QA pairs).

Additionally, our two reasoning tasks include:

1. \textbf{Table Reasoning (Fig.~\ref{fig:table_r}).} This involves tasks that go beyond simple perception, such as comparing the values of two elements in specific positions within a table, filtering the table based on given conditions, or determining the maximum and minimum values ($174$ QA pairs).

2. \textbf{Diagram Reasoning (Fig.~\ref{fig:diagram_r}).} Similar to table reasoning, but reasoning with diagrams involves distinguishing specific colors in the legend and assessing the height of curves or bars ($326$ QA pairs).

\subsection{Remote sensing}
\textbf{Data Characteristics.}  From real remote sensing data, some images have extremely high quality, with individual image sizes reaching up to $139$MB and containing rich details.

\subsubsection{Data Sources and Annotation Process}

We select high-resolution images from public remote sensing datasets with rich information. For example, the FAIR1M dataset~\citep{sun2021fair1m} focuses on fine-grained object recognition and detection using high-resolution ($0.3-0.8$m) RGB images from Gaogen (GF) satellites extracted via Google Earth. It contains $15,000$ images annotated with rotated bounding boxes across $5$ main categories (ships, vehicles, airplanes, courts, and roads), further divided into $37$ sub-categories. The Potsdam dataset\footnote{https://paperswithcode.com/dataset/isprs-potsdam} dataset includes $38$ patches of true orthophotos (TOP) extracted from larger mosaics. VGoogle~\citep{hou2019v}, VBing~\citep{hou2019v}, and VArcGIS~\citep{hou2019v} datasets, derived from Google Earth, Bing World Imagery, and ArcGIS World Imagery respectively, each feature 38 classes with a total of approximately $59,000$ images per dataset. Each class contains at least $1,500$ images, with spatial resolutions ranging from $0.07$ to $38.22$ meters.

\textbf{Annotation.} For all the questions in this subsection, 20 volunteers manually create the questions and answers, while another expert reviews the quality of the questions to ensure they meet the required standards.

\subsubsection{Evaluation Dimensions and Benchmark Statistics} 

\begin{figure*}[t]
\centering
\subfigure[{Object Counting.}]{
\begin{minipage}[t]{0.75\linewidth}
\centering
 \includegraphics[width=\linewidth]{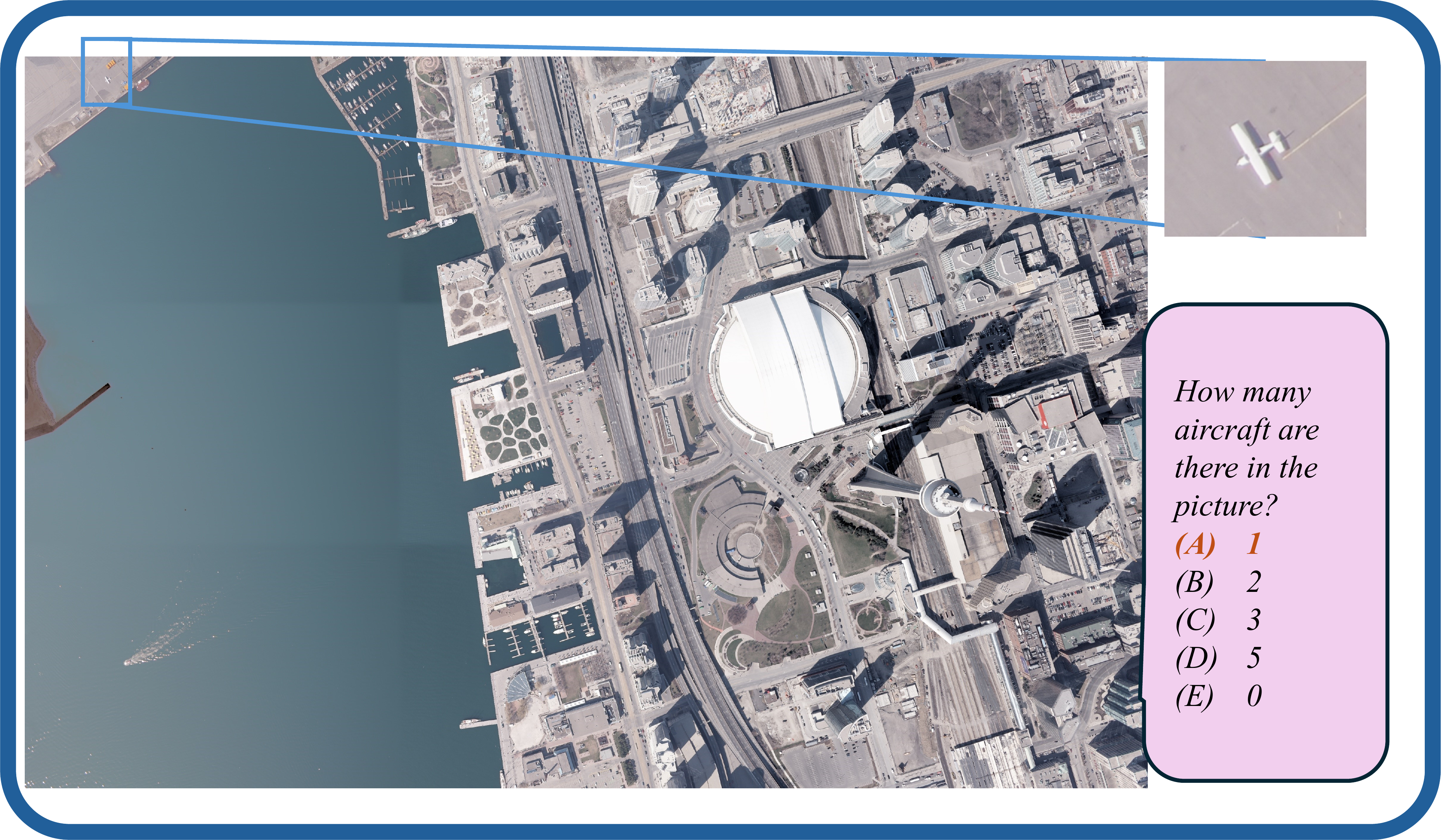}
    \label{fig:remote_cny}
\end{minipage}%
}%

\subfigure[{Color Recognition.}]{
\begin{minipage}[t]{0.75\linewidth}
\centering
 \includegraphics[width=\linewidth]{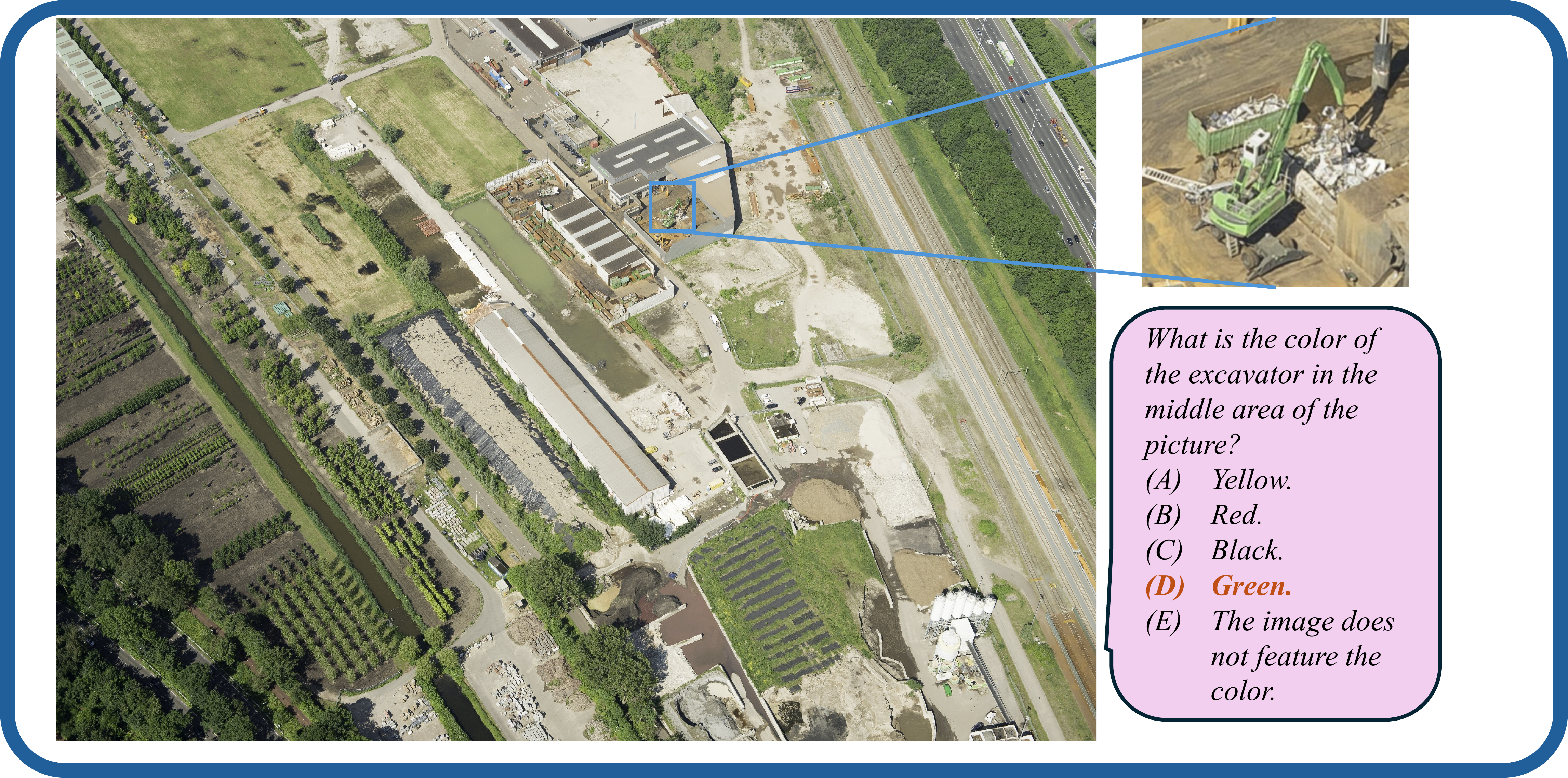}
    \label{fig:remote_color}
\end{minipage}%
}%

\subfigure[{Spatial Relationship Understanding.}]{
\begin{minipage}[t]{0.75\linewidth}
\centering
 \includegraphics[width=\linewidth]{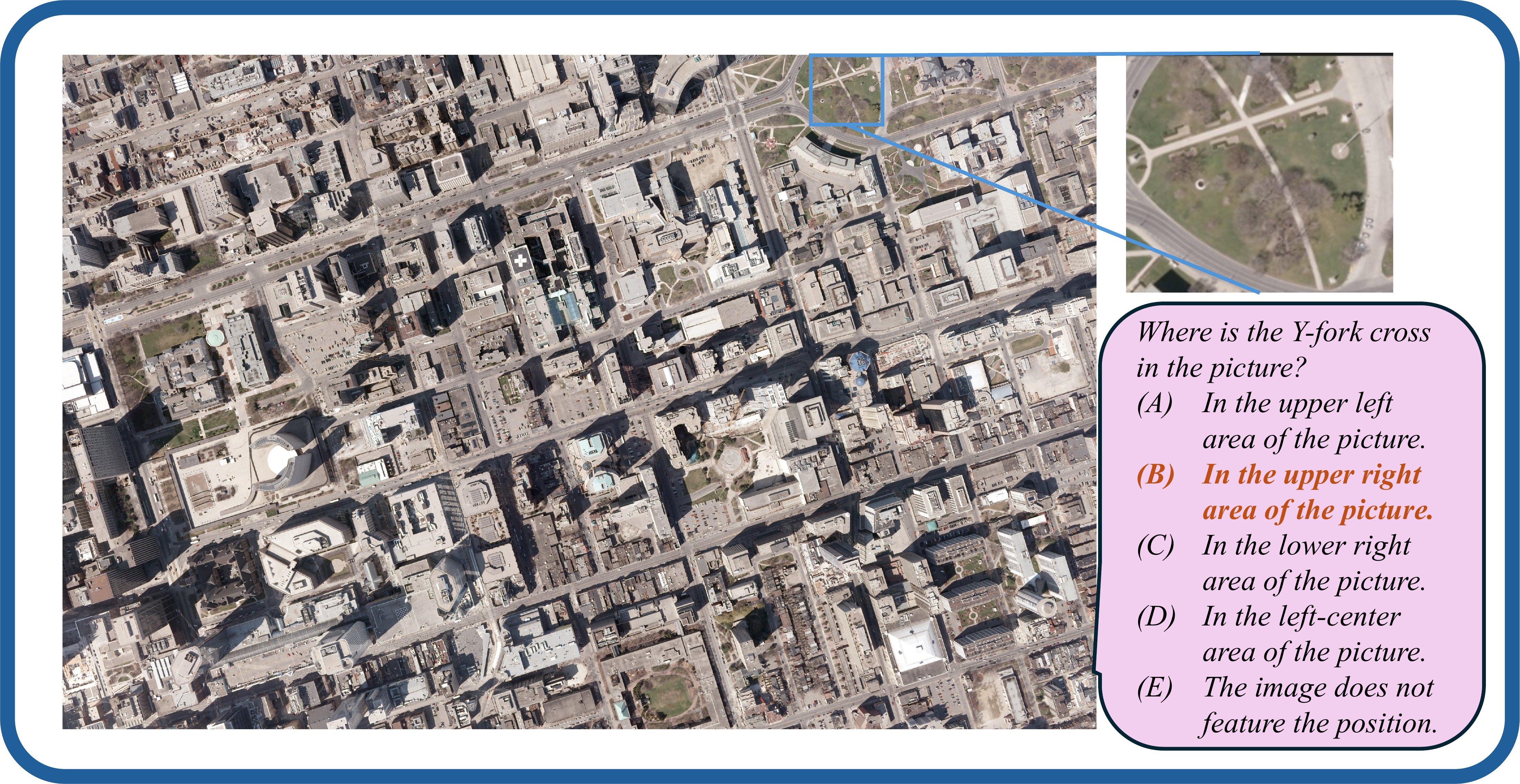}
    \label{fig:remote_position}
\end{minipage}%
}%
\centering
\vspace{-0.2cm}
\caption{\textbf{Data Examples for Perception Tasks in Remote Sensing}}
\label{fig:remote}
\end{figure*}

Remote sensing images have a wide range of applications in real-world scenarios. During the construction of our dataset, we observe that many tasks are challenging for humans. For example, counting the number of airplanes in Fig.~\ref{fig:remote_cny} requires careful observation and counting by human annotators. Automating this process with multimodal large models would be highly valuable for remote sensing applications. We select a total of $1,298$ high-quality images and design three specific tasks tailored for remote sensing images:

1. \textbf{Object Counting (Fig.~\ref{fig:remote_cny}).} Task involves counting specific objects such as airplanes, ships, or buildings within a given image ($1,255$ QA pairs).

2. \textbf{Color Recognition (Fig.~\ref{fig:remote_color}).} Task involves identifying and describing the colors of specific objects in the image ($1,226$ QA pairs).

3.\textbf{Spatial Relationship Understanding (Fig.~\ref{fig:remote_position}).} Understanding both the absolute spatial relationships and relative spatial relationships between objects in the images ($1,257$ QA pairs).

\subsection{Autonomous Driving}
\textbf{Data Characteristics.} The front-view driving datas are recorded using onboard cameras with various sensor configurations. The images encompass diverse weather conditions (e.g., sunny, night, rainy, etc.), geographic locations (e.g., US, SG, CN), and complex traffic scenarios (e.g., urban, highway, etc.).

\subsubsection{Data Sources and Annotation Process}

\textbf{Data Sources.} We select high-quality images from large open-source driving datasets, each with distinct advantages. The Rank2Tell dataset \citep{sachdeva2024rank2tell} ranks the importance level of surrounding objects for driving safety. Additionally, it provides dense annotations of semantic, spatial, and relational attributes with bounding boxes for approximately $2,600$ frames captured at intersections, and it stitches images from three cameras to deliver a wide field of view (FOV). To enhance the reliability of autonomous driving systems, the CODA dataset \citep{li2022coda} collects $1,500$ driving scenes, each containing object-level corner cases, and labels more than $30$ novel categories (e.g., garbage bag, concrete block, etc.). It focuses on evaluating performance of perception systems in detecting out-of-distribution (OOD) objects compared to common traffic elements. The nuScenes dataset \citep{caesar2020nuscenes}, one of the most popular real-world autonomous driving datasets, provides abundant 3D perception annotations with a semantic map and CAN bus expansion \citep{li2023open}. Based on nuScenes \citep{caesar2020nuscenes}, DriveLM-nuScenes \citep{sima2023drivelm} links approximately $4,800$ key frames with driving behaviors and motions by formulating 3P reasoning (perception, prediction, planning) as a series of rich question-answer pairs in a directed graph.

\textbf{Annotation.} For all the questions in this subsection, a professional researcher manually generates the questions and answers based on the source datasets’ labels, achieving their non-ambiguity, challenge and complexity. Another expert reviews the quality of the questions to ensure
they meet the required standards.

\subsubsection{Evaluation Dimensions and Benchmark Statistics} 

\begin{figure*}[t]
\centering
\subfigure[{Object Identification.}]{
\begin{minipage}[t]{0.5\linewidth}
\centering
 \includegraphics[width=\linewidth]{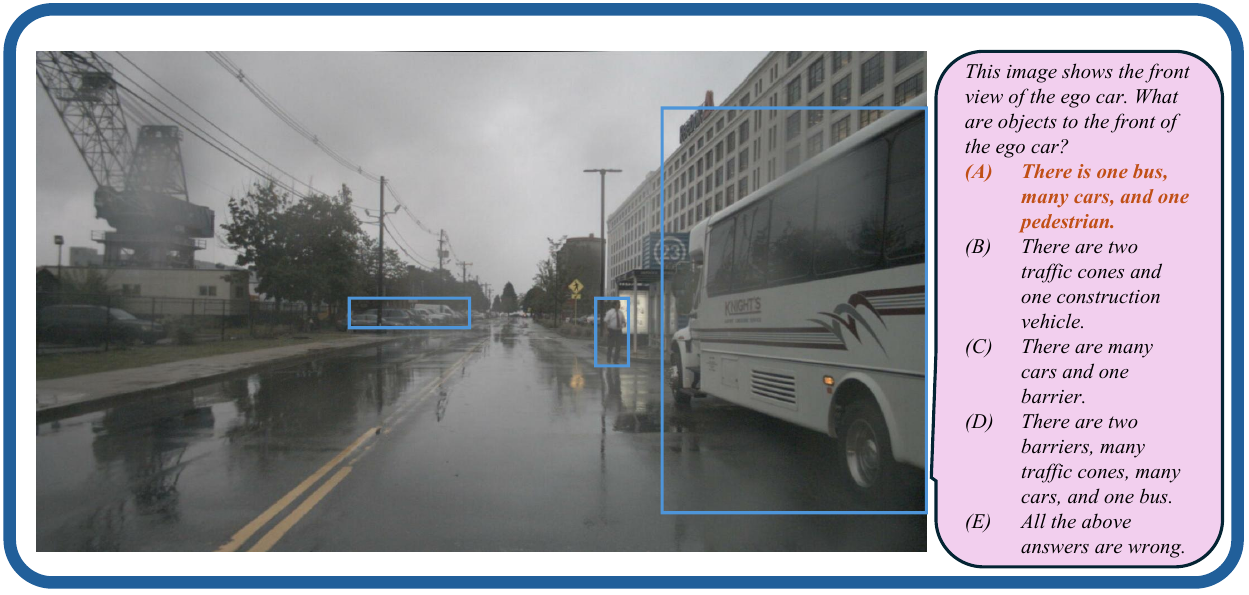}
    \label{fig:ad_object_identify}
\end{minipage}%
}%
\subfigure[{Object Counting.}]{
\begin{minipage}[t]{0.5\linewidth}
\centering
 \includegraphics[width=\linewidth]{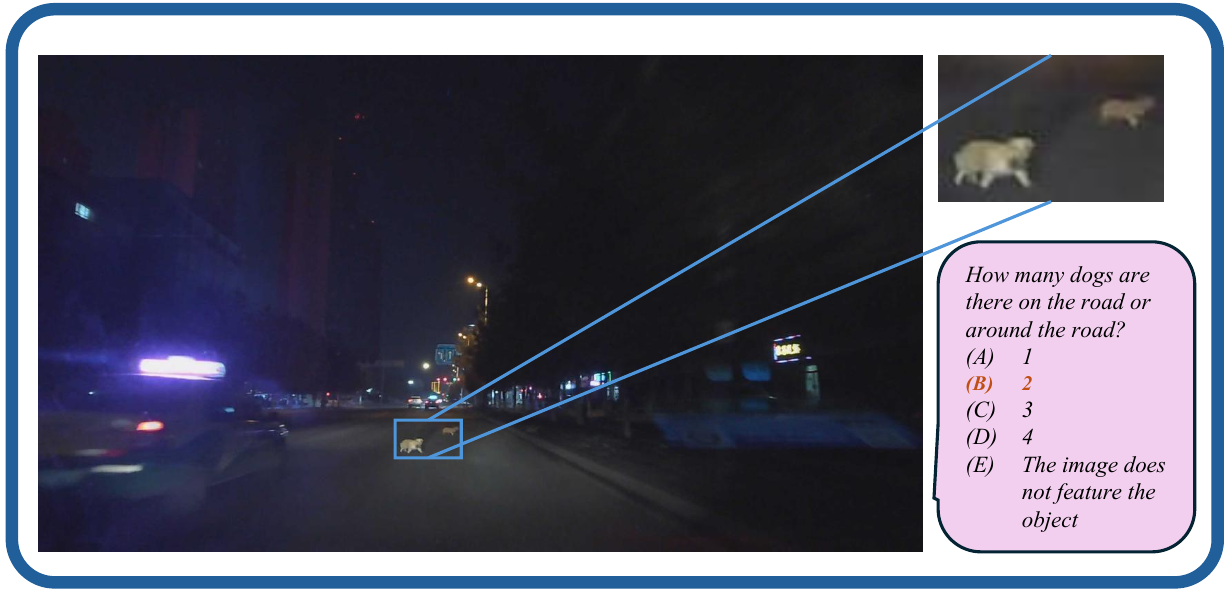}
    \label{fig:ad_object_counting}
\end{minipage}%
}%

\subfigure[{Motion Attribute Identification of Multiple Pedestrians.}]{
\begin{minipage}[t]{0.5\linewidth}
\centering
 \includegraphics[width=\linewidth]{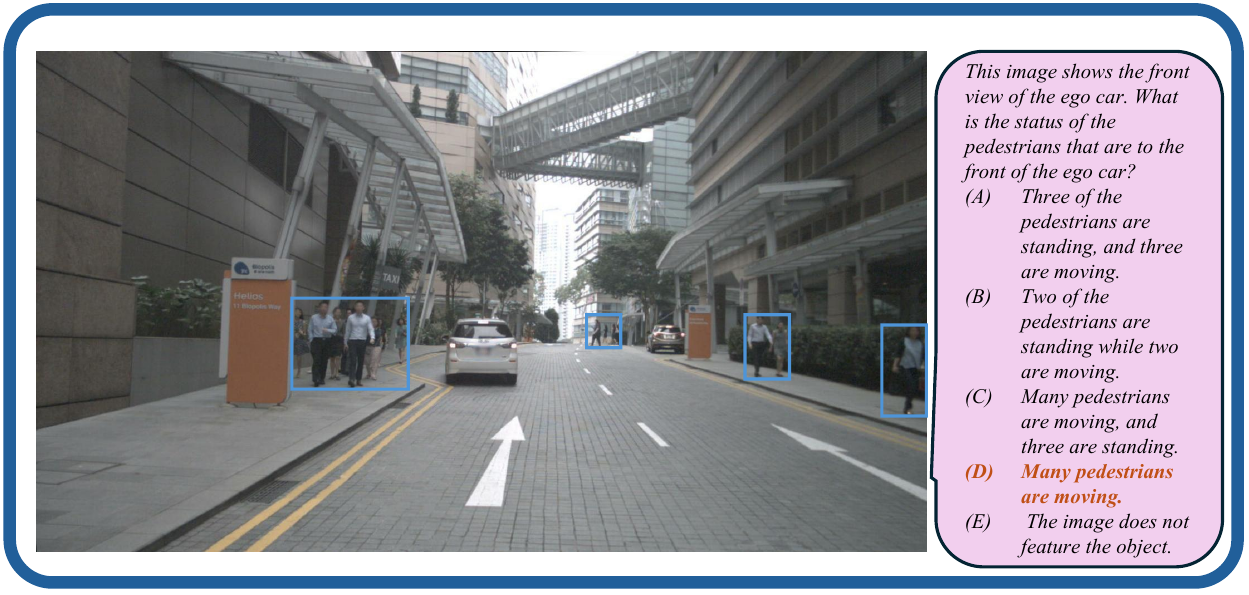}
    \label{fig:ad_attri_multipedes}
\end{minipage}%
}%
\subfigure[{Motion Attribute Identification of Multiple Vehicles.}]{
\begin{minipage}[t]{0.5\linewidth}
\centering
 \includegraphics[width=\linewidth]{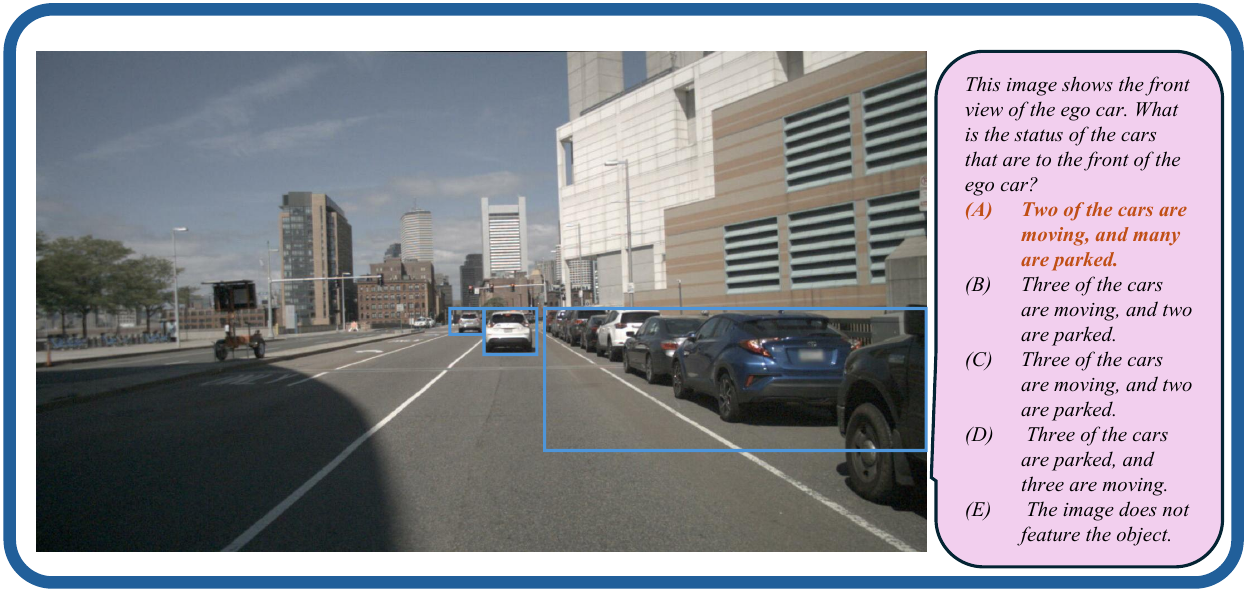}
    \label{fig:ad_attri_multivehicle}
\end{minipage}%
}%

\subfigure[{Visual Attribute Identification of a Specific Traffic Signal}]{
\begin{minipage}[t]{1\linewidth}
\centering
 \includegraphics[width=\linewidth]{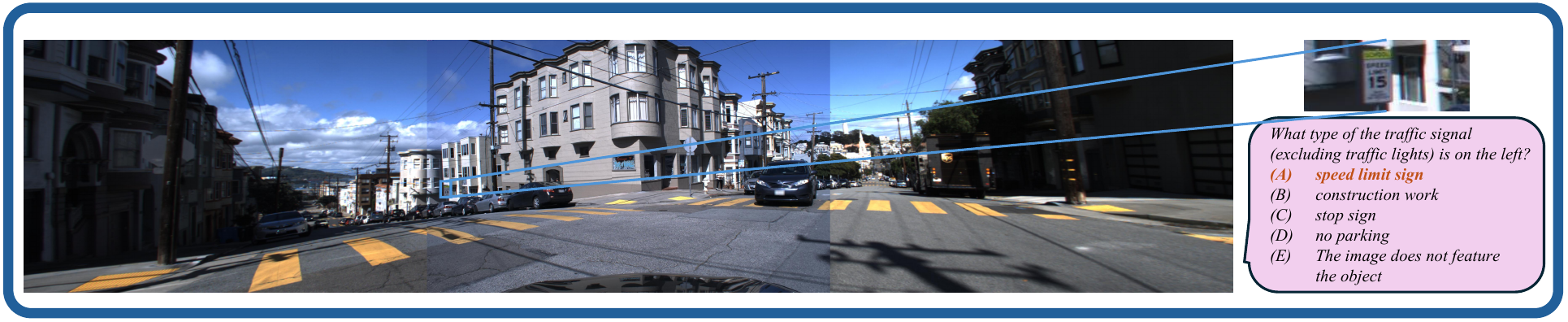}
    \label{fig:ad_attri_signal}
\end{minipage}%
}%

\subfigure[{Motion Attribute Identification of a Specific Pedestrian}]{
\begin{minipage}[t]{1\linewidth}
\centering
 \includegraphics[width=\linewidth]{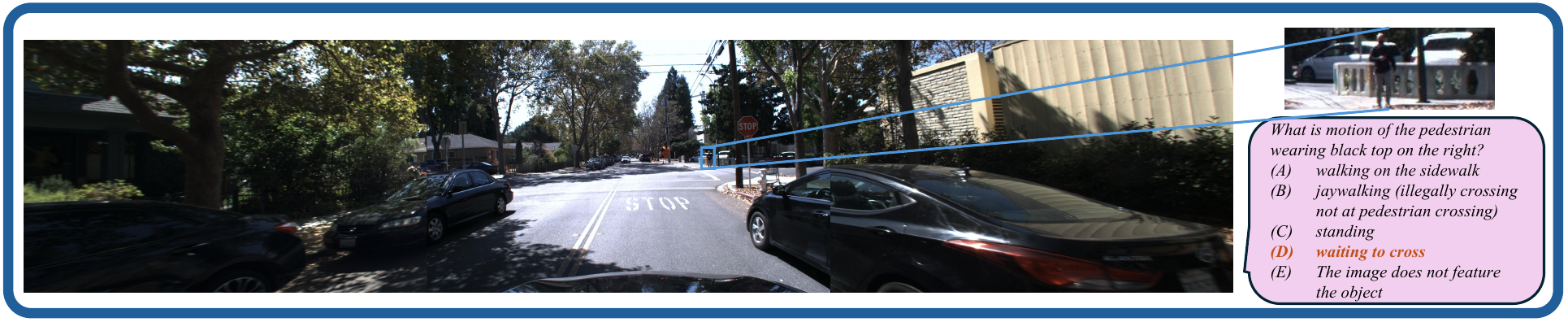}
    \label{fig:ad_attri_pedes}
\end{minipage}%
}%

\centering
\vspace{-0.2cm}
\caption{\textbf{Data Examples for Perception Tasks in Autonomous Driving }}
\label{fig:ad_perception}
\end{figure*}

Vision-centric autonomous driving is one of the most significant applications of artificial intelligence. However, unresolved issues remain, including both object-level and task-level corner cases, as well as safe-critical and human-like planning \citep{yang2023llm4drive}. MLLMs with general knowledge and the ability of driving scenarios embodied understanding \citep{gao2024survey, zhou2024embodied} are seen as a promising solution to achieve Level $4$ autonomous driving. Specifically, we have designed three main perception tasks and three main reasoning tasks, which are further subdivided into a total of fifteen sub-tasks. It is worth noting that, as traditional detection tasks in autonomous driving have largely been addressed by modern perception models, our focus is shifting towards perception challenges involving small or distant objects, specifically those that occupy less than $1/100$ of the total image area. Meanwhile, LLMs must possess extensive driving expertise and even a deep understanding of 3D spatial concepts in order to effectively address the complex reasoning challenges. For perception tasks:

1. \textbf{Object Identification (Fig.~\ref{fig:ad_object_identify}).} Describing the main traffic elements in front of the ego car including their categories and corresponding quantities ($1,101$ images and $1,101$ QA pairs).

2. \textbf{Object Attribute Identification.} Task involves identifying the attribute of a specific object according to its appearance and location (a total of $454$ images and $523$ QA pairs), and describing the attributes of all objects within a specific category (a total of $1,167$ images and $1,315$ QA pairs) in traffic scenarios. In terms of sub-tasks, the former includes the visual attribute of a traffic signal (Fig.~\ref{fig:ad_attri_signal}, $157$ images and $201$ QA pairs) and the motion attribute of a pedestrian (Fig.~\ref{fig:ad_attri_pedes}, $152$ images and $164$ QA pairs) or a vehicle ($145$ images and $158$ QA pairs), and the latter includes the motion attributes of multiple pedestrians (Fig.~\ref{fig:ad_attri_multipedes}, $493$ images and $492$ QA pairs) or vehicles (Fig.~\ref{fig:ad_attri_multivehicle}, $674$ images and $823$ QA pairs).

3. \textbf{Object Counting (Fig.~\ref{fig:ad_object_counting}).} Counting special traffic elements in the given image, such as cars, trucks, traffic signals, etc., especially some novel objects compared to traditional autonomous driving tasks such as garbage bags, dogs, concrete blocks, etc. ($647$ images and $720$ QA pairs).

\begin{figure*}[t]
\centering
\subfigure[{Intention Prediction of Ego Vehicle.}]{
\begin{minipage}[t]{0.5\linewidth}
\centering
 \includegraphics[width=\linewidth]{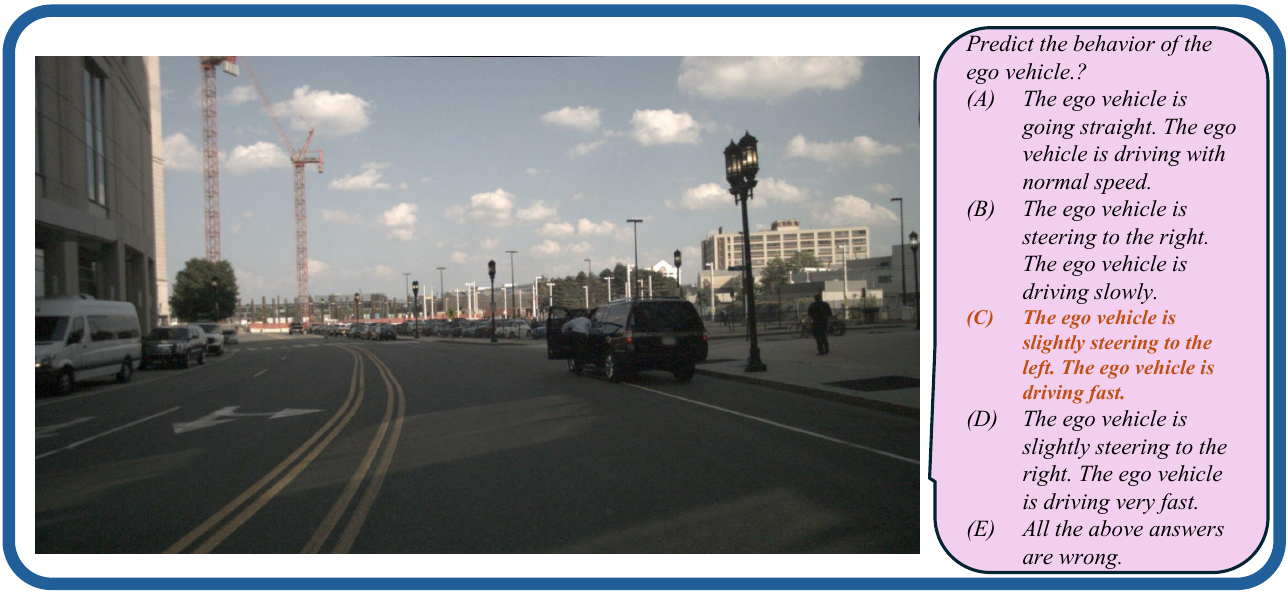}
 \label{fig:ad_intention_prediction_ego}
\end{minipage}%
}%
\subfigure[{Intention Prediction of a Specific Vehicle.}]{
\begin{minipage}[t]{0.5\linewidth}
\centering
 \includegraphics[width=\linewidth]{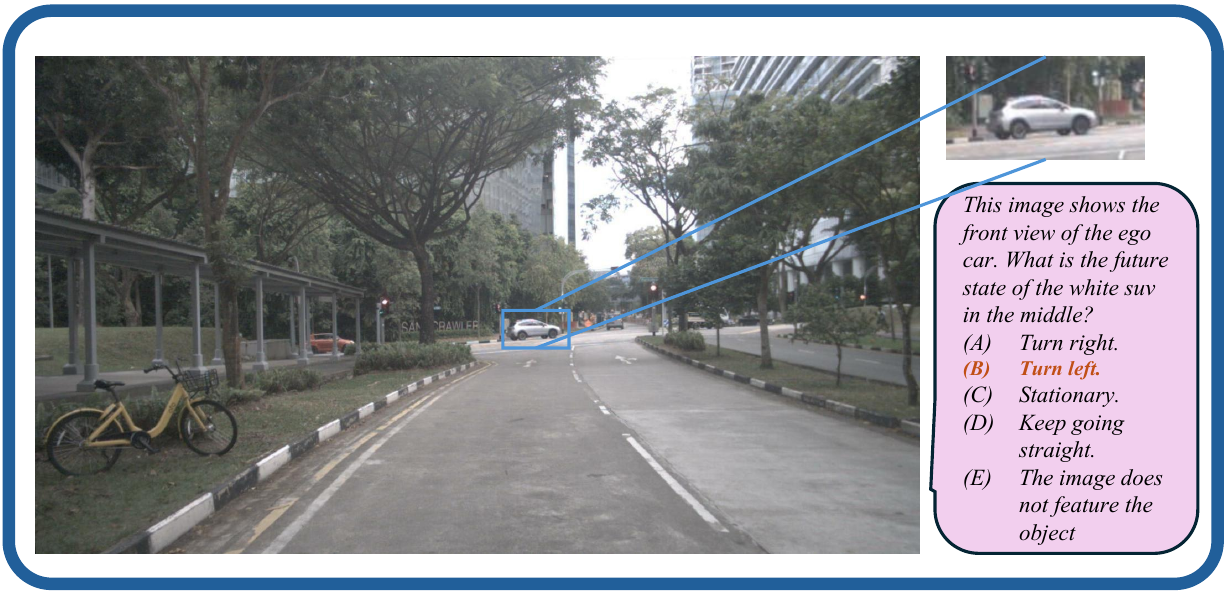}
    \label{fig:ad_intention_prediction_vehicle}
\end{minipage}%
}%

\subfigure[{Driver Attention Understanding}]{
\begin{minipage}[t]{0.5\linewidth}
\centering
 \includegraphics[width=\linewidth]{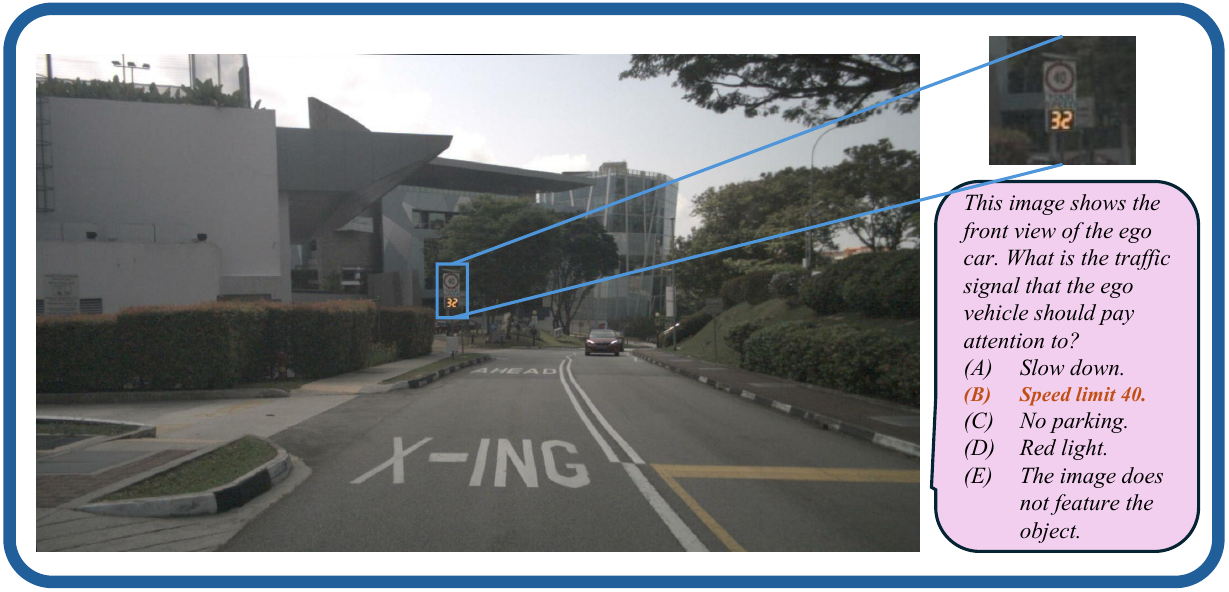}
    \label{fig:ad_driver_attention}
\end{minipage}%
}%
\subfigure[{Interaction Relation between Traffic Elements}]{
\begin{minipage}[t]{0.5\linewidth}
\centering
 \includegraphics[width=\linewidth]{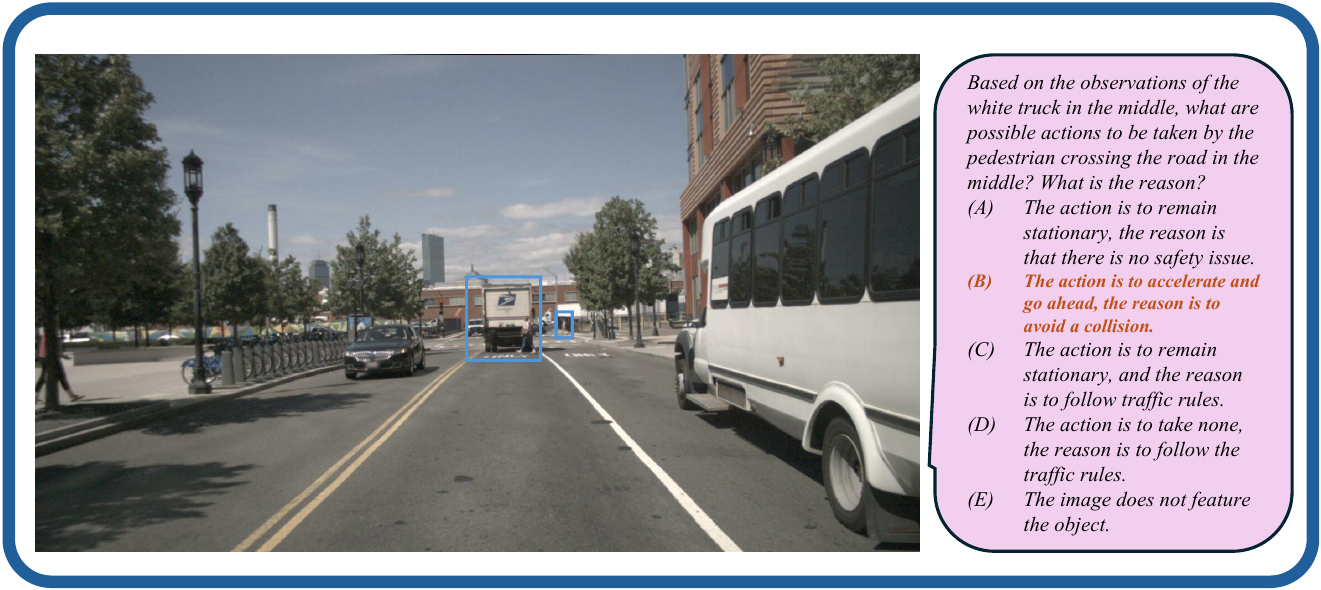}
    \label{fig:ad_relation_o2o}
\end{minipage}%
}%

\subfigure[{Interaction Relation between Ego Vehicle and Traffic Elements}]{
\begin{minipage}[t]{1\linewidth}
\centering
 \includegraphics[width=\linewidth]{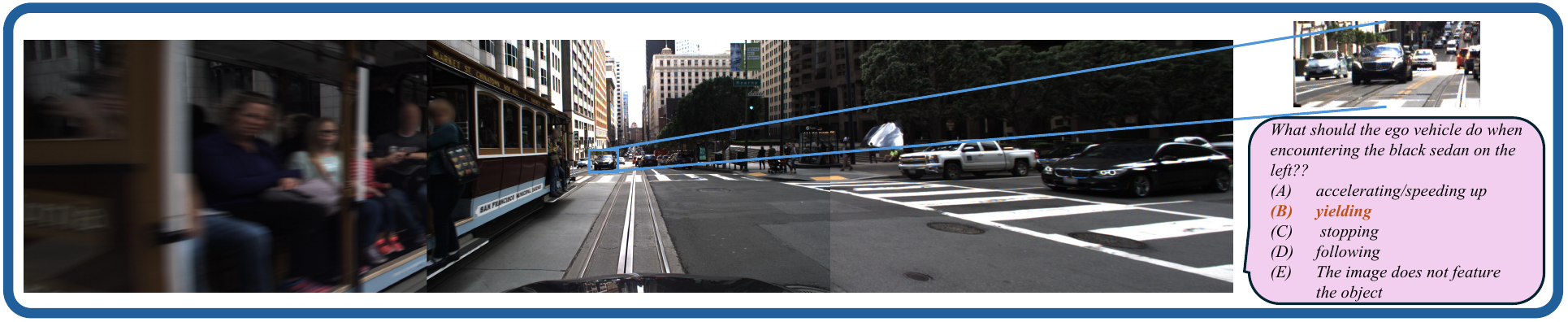}
    \label{fig:ad_relation_ego2vehicle}
\end{minipage}%
}%

\centering
\vspace{-0.2cm}
\caption{\textbf{Data Examples for Reasoning Tasks in Autonomous Driving }}
\label{fig:ad_reasoning}
\end{figure*}

Furthermore, reasoning tasks are as follows:

1. \textbf{Intention Prediction.} Task involves predicting the intention of a designated traffic agent in the given image (a total of $582$ images and $614$ QA pairs). In terms of sub-tasks, it contains fine-grained behavior prediction of the ego vehicle (Fig.~\ref{fig:ad_intention_prediction_ego}, $304$ images and $304$ QA pairs) and future intention of a pedestrian ($95$ images and $103$ QA pairs) or a vehicle (Fig.~\ref{fig:ad_intention_prediction_vehicle}, $183$ images and $207$ QA pairs).

2. \textbf{Interaction Relation Understanding.} Task involves reasoning the interaction relation between two specific traffic elements (a total of $444$ images and $513$ QA pairs). In terms of sub-tasks, it contains the ego vehicle’s reaction to a specific object (Fig.~\ref{fig:ad_relation_ego2vehicle}), which is further categorized into three categories: pedestrian ($102$ images and $106$ QA pairs), vehicle ($95$ images and $101$ QA pairs), and traffic signal ($81$ images and $105$ QA pairs). Additionally, another sub-task is predicting the interactions between the aforementioned objects, excluding the ego vehicle (Fig.~\ref{fig:ad_relation_o2o}, $166$ images and $201$ QA pairs).

3. \textbf{Driver Attention Understanding (Fig.~\ref{fig:ad_driver_attention}).} Reasoning the traffic signal that the driver should pay attention to in the given front view image, such as yellow light, speed limit sign, no parking sign, etc. ($217$ images and $217$ QA pairs).

\subsection{Monitoring}
\subsubsection{Data Sources and Annotation Process}

\textbf{Data Characteristics.} Monitoring images are captured from different cameras (e.g., drone-equipped cameras, fixed surveillance cameras, infrared cameras), viewpoints (arbitrary and fixed viewpoints), scene complexities (e,g., streets, shopping malls, intersections, campus, etc.), and environmental factors (day and night).

We select high-resolution images from public monitoring image datasets with many real-world challenges. For example, the VisDrone dataset \citep{zhu2021detection} brings several challenges, e.g., viewpoint variations, scale variations, and out-of-view, etc. Additionally, its dataset contains $263$ video clips with $179,264$ frames and $10,209$ static images, which are captured via various drone-equipped cameras across various categories (e.g., pedestrian, people, bicycle, car, van, truck, tricycle, awning-tricycle, bus, and motor), density (e.g., sparse and crowded scenes) and environments (e.g., urban and rural regions). Additionally, The second dataset \footnote{This dataset is publicly accessible through the AI Studio website at https://aistudio.baidu.com/datasetdetail/28831.}, is collected in diverse environments (e.g., street, mall, elevator, etc.) for crowd density prediction task, features $3,000$ images with only person category, captured by fixed surveillance cameras. This dataset is highly diverse from the camera viewpoints (low altitude, high altitude, fisheye, etc.), scale size, and scene complexities. The LLVIP dataset \citep{jia2021llvip}, which is a visible-infrared paired dataset for low-light vision, contains $30,976$ images taken in binocular cameras, and contains a large number of pedestrians. We only select manually infrared images from it to test the model's robustness on different modals.

\textbf{Annotation.} For all the questions in this subsection, two professional researchers manually create the questions and answers, and another expert reviews the quality of the questions to ensure they meet the required standards.

\subsubsection{Evaluation Dimensions and Benchmark Statistics} 

\begin{figure*}[t]
\centering
\subfigure[{Object Counting in Monitoring Images.}]{
\begin{minipage}[t]{0.5\linewidth}
\centering
 \includegraphics[width=\linewidth]{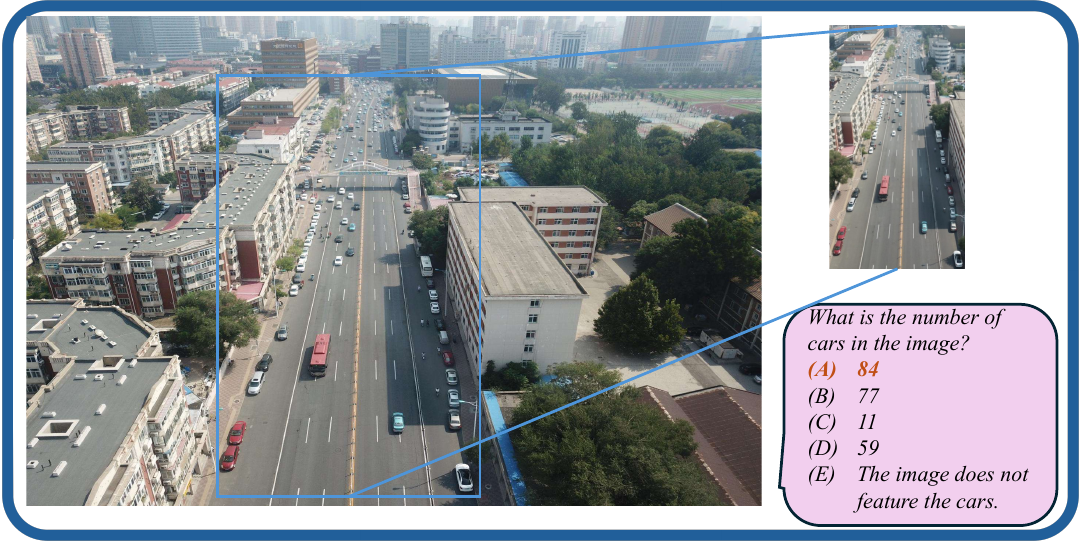}
    \label{fig:monitoring_cny}
\end{minipage}%
}%
\subfigure[{Object Existence in Monitoring Images.}]{
\begin{minipage}[t]{0.5\linewidth}
\centering
 \includegraphics[width=\linewidth]{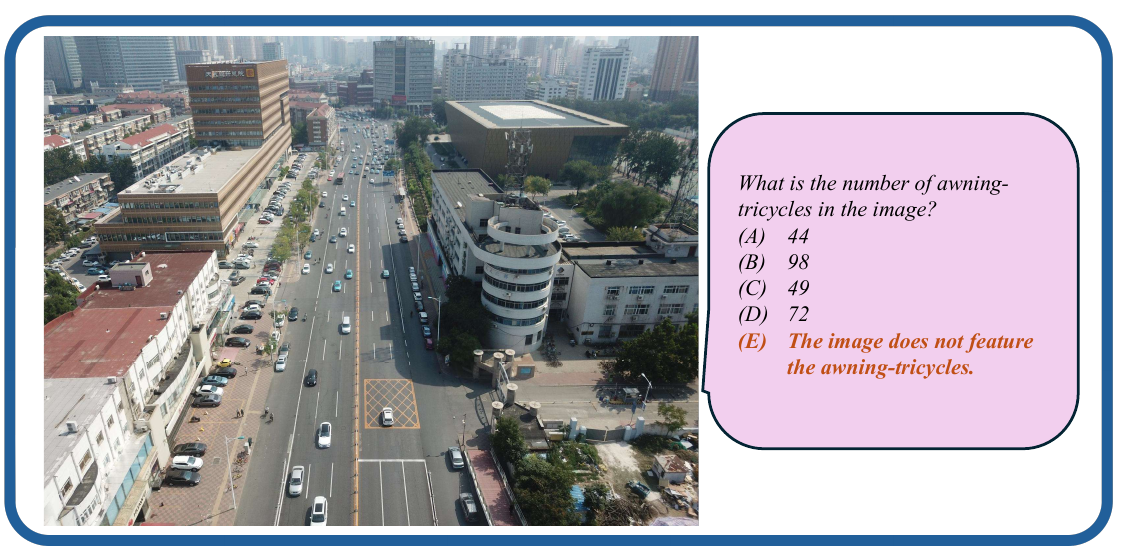}
    \label{fig:monitoring_existence}
\end{minipage}%
}%

\subfigure[{Object Location in Monitoring Images.}]{
\begin{minipage}[t]{0.5\linewidth}
\centering
 \includegraphics[width=\linewidth]{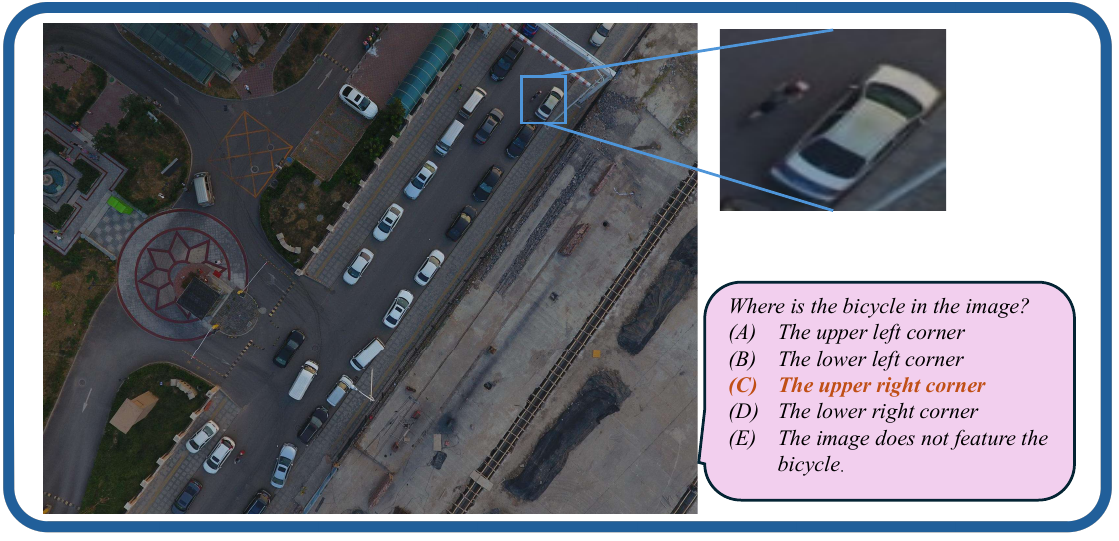}
    \label{fig:monitoring_location}
\end{minipage}%
}%
\subfigure[{Color Recognition in Monitoring Images.}]{
\begin{minipage}[t]{0.5\linewidth}
\centering
 \includegraphics[width=\linewidth]{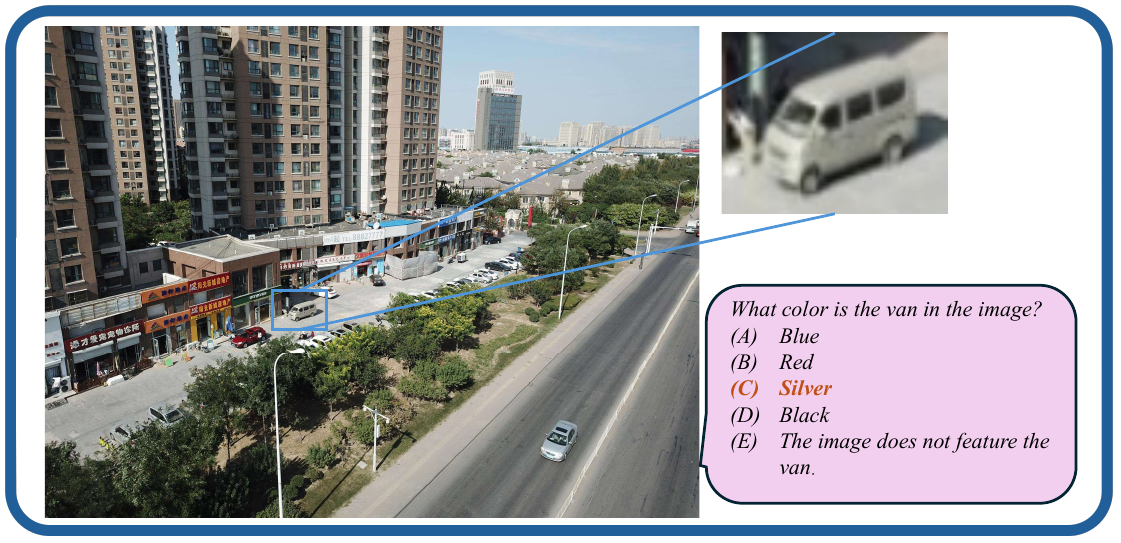}
    \label{fig:monitoring_color}
\end{minipage}%
}%

\subfigure[{Orientation Perception in Monitoring Images.}]{
\begin{minipage}[t]{0.75\linewidth}
\centering
 \includegraphics[width=\linewidth]{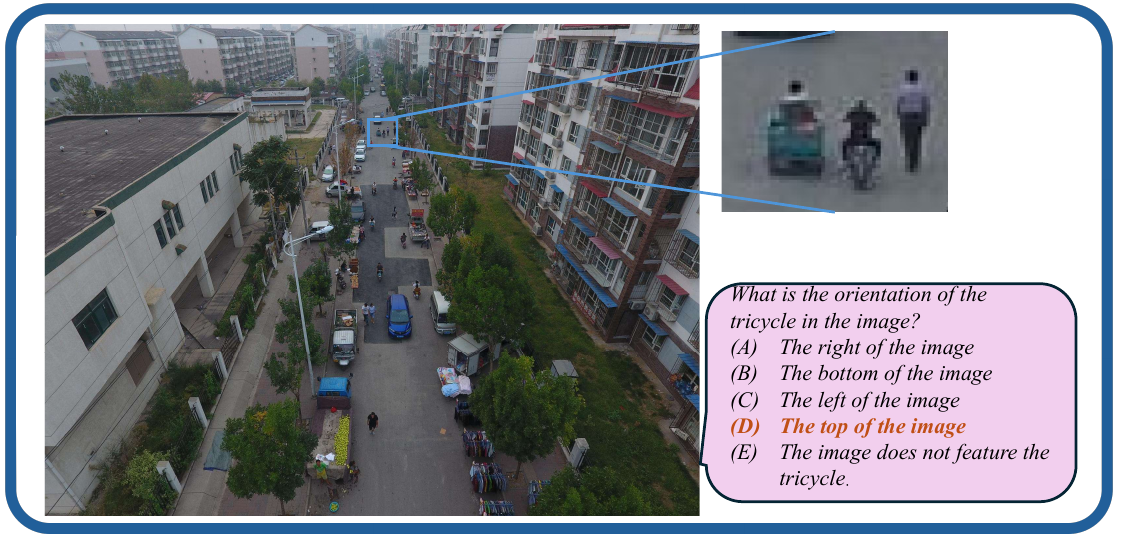}
    \label{fig:monitoring_orientation}
\end{minipage}%
}%

\centering
\vspace{-0.2cm}
\caption{\textbf{Data Examples for Perception Tasks in Monitoring Images}}
\label{fig:monitoring_perception}
\end{figure*}

Monitoring images are widely applied in real-world scenarios to increase public safety. Analyzing the monitoring images accurately with MLLMs would be highly valuable for public safety and crowd management. Specifically, we have designed three main tasks for monitoring images:

1. \textbf{Object Counting (Fig.~\ref{fig:monitoring_cny}).} Task involves counting specific objects such as pedestrians, cars, or trucks in the given monitoring images (a total of $1,600$ images and $1,600$ QA pairs). Noted that, when the count of a specific object is equal to zero, this object counting task can be transformed well into the object existence task (Fig.~\ref{fig:monitoring_existence}, for judging whether a specific object exists in the given images. Thus, the object existence task can be regarded as a special case of the counting task. Additionally, we categorize this task into two sub-tasks for vehicle counting ($608$ images and $608$ QA pairs) and person counting ($992$ images and $992$ QA pairs), respectively.

2. \textbf{Object Location (Fig.~\ref{fig:monitoring_location}).} Task involves judging the location of the specific vehicles, like cars, or trucks in the given monitoring images (a total of $136$ images and $136$ QA pairs).

3. \textbf{Attribute Recognition.} Task involves identifying and describing the attributes of specific objects, e.g., \textit{color recognition (Fig.~\ref{fig:monitoring_color})} and \textit{orientation perception (Fig.~\ref{fig:monitoring_orientation})}, in the monitoring images (a total of $460$ images and $460$ QA pairs). Additionally, we categorize this task into two sub-tasks for the vehicle ($352$ images and $352$ QA pairs) and person attribute recognition tasks ($108$ images and $108$ QA pairs), respectively.

In addition, there are three seasoning tasks described as follows.

\begin{figure*}[t]
\centering
\subfigure[{Reasoning Task of Calculating the Sum of Different Objects in Monitoring Images.}]{
\begin{minipage}[t]{0.75\linewidth}
\centering
 \includegraphics[width=\linewidth]{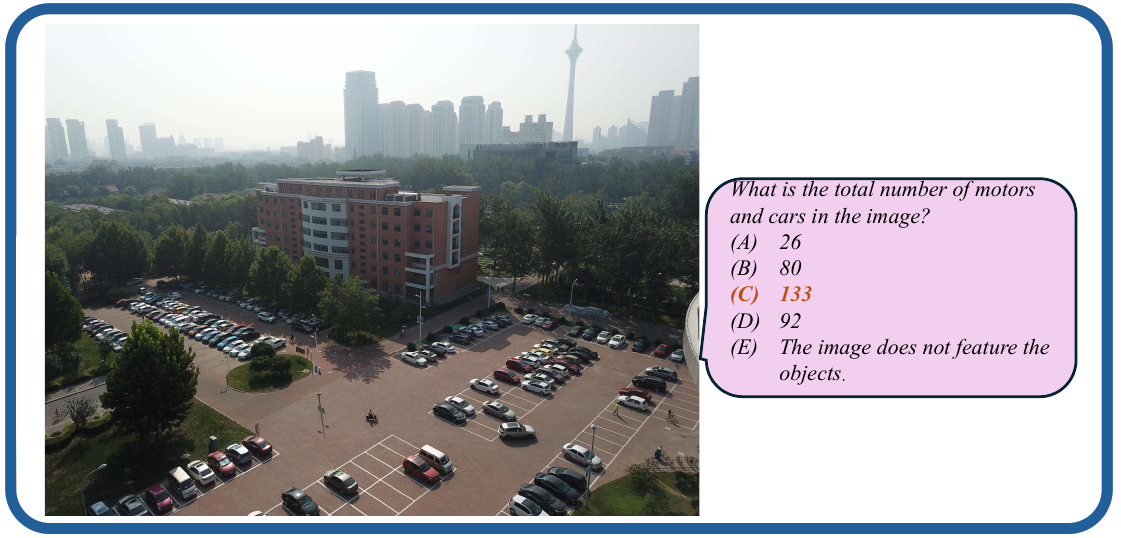}
    \label{fig:reasoning_cal}
\end{minipage}%
}%

\subfigure[{Reasoning task of Intention of the Special Object in Monitoring Images.}]{
\begin{minipage}[t]{0.75\linewidth}
\centering
 \includegraphics[width=\linewidth]{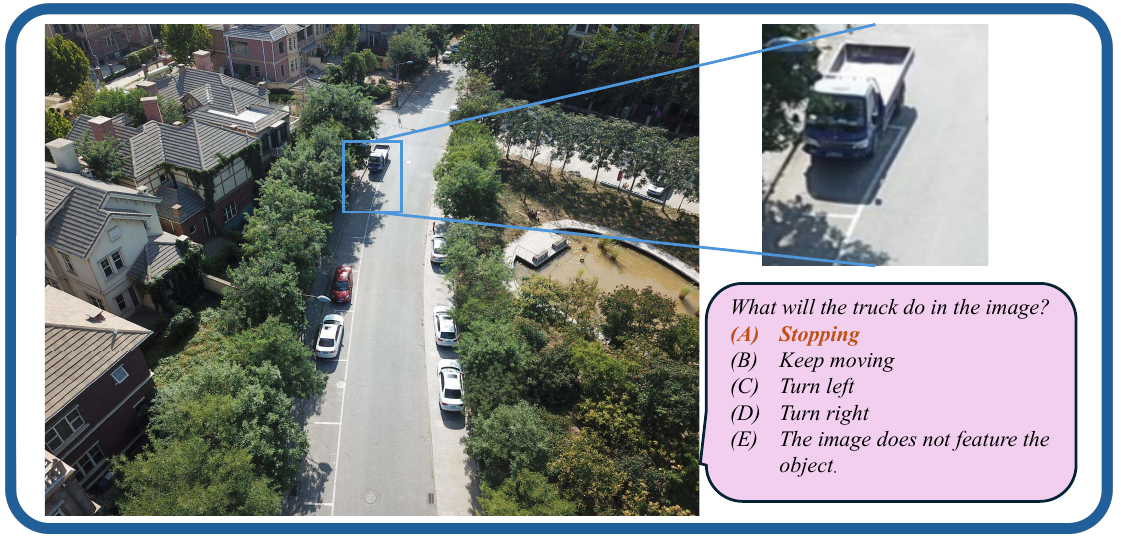}
    \label{fig:reasoning_intention}
\end{minipage}%
}%

\subfigure[{Reasoning task of Attribute of the Special Object in Monitoring Images.}]{
\begin{minipage}[t]{0.75\linewidth}
\centering
 \includegraphics[width=\linewidth]{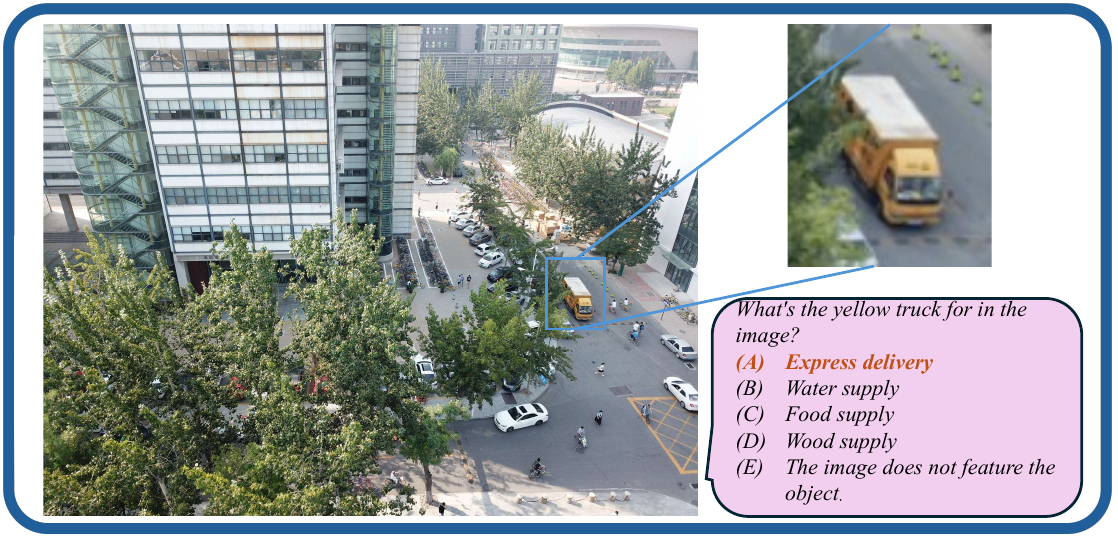}
    \label{fig:reasoning_property}
\end{minipage}%
}%

\centering
\vspace{-0.2cm}
\caption{\textbf{Data Examples for Reasoning Tasks in Monitoring Images}}
\label{fig:monitoring}
\end{figure*}

1. \textbf{Calculate the Sum of Different Objects (Fig.~\ref{fig:reasoning_cal}).} Counting various objects and calculating their total number accurately ($300$ images and $300$ QA pairs).

2. \textbf{Intention Reasoning (Fig.~\ref{fig:reasoning_intention}).} Reasoning the next route and turn of the specific object ($98$ images and $98$ QA pairs).

3. \textbf{Attribute Reasoning (Fig.~\ref{fig:reasoning_property}).} Reasoning the specific materials and functions of the given objects, such as inferring the function of the dustbin via its appearance ($100$ images and $100$ QA pairs).

\begin{figure}
    \centering
    \includegraphics[width=\linewidth]{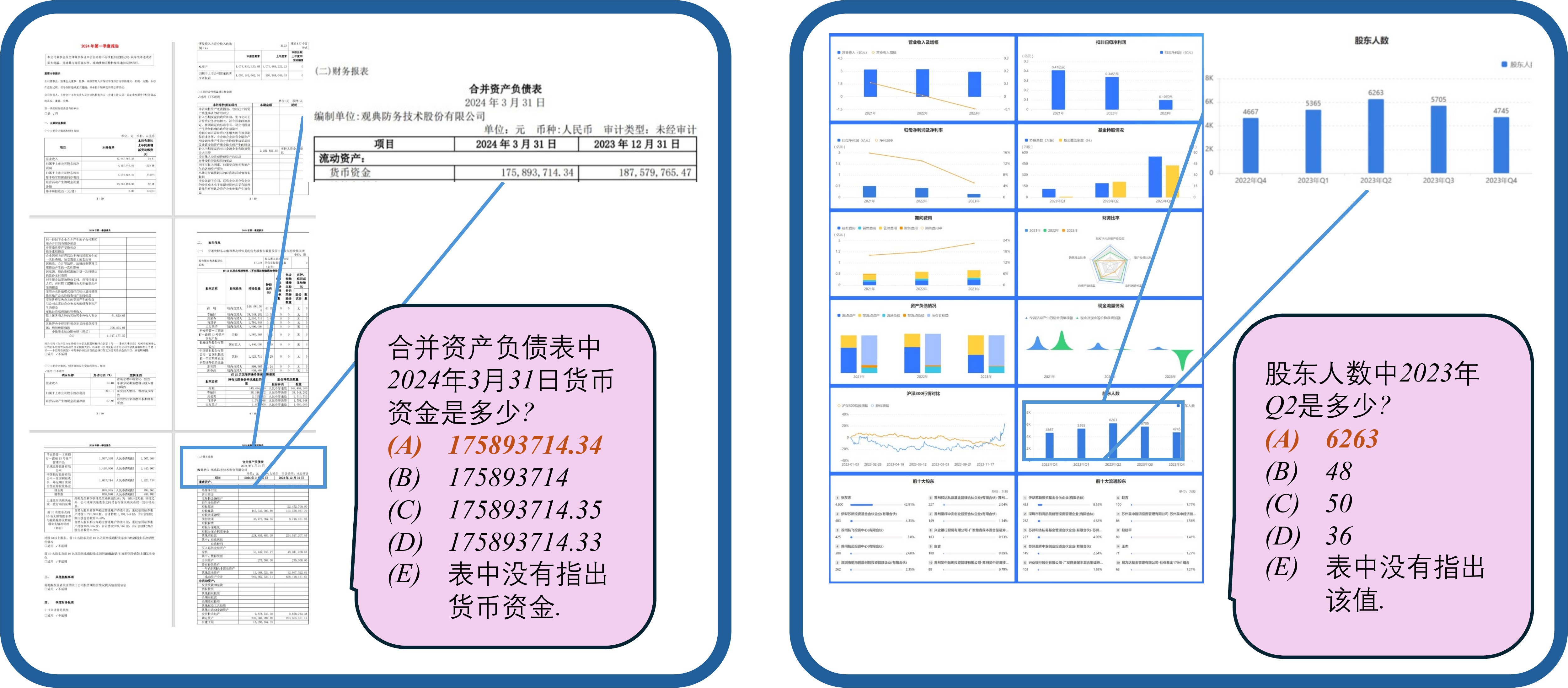}
\caption{\textbf{Data Examples for Perception Tasks in \abbr-CN}}
\label{fig:mme-hd-cn}
\end{figure}

\section{Experimental Results on All Task Splits}\label{app:sec_res}

\textbf{OCR in The Wild Performance.} Tab.~\ref{tab:res_ocr} displays the performance of various models on real-world OCR tasks. Generally speaking, when image resolution is high, the more advanced models still demonstrate commendable OCR capabilities. However, this does not imply that our task is of low difficulty. The average accuracy rates of Qwen-VL and the basic LLaVA model on perception tasks are only slightly better than random guessing. In this task, the gap between open-source and closed-source models is not significant. GPT-4o ranks first in overall performance, while Claude 3.5 Sonnet leads in reasoning tasks.
\begin{table}[]
\caption{\textbf{Experimental results on the \texttt{OCR in the Wild} tasks} are categorized as follows: “Product” represents products and advertisements; “B \& M \& P” represents books, maps, and posters; “Contact” denotes contact information and addresses; “Identity” pertains to identity information; and “Signage” refers to signage and other text. Models are ranked according to their average performance on perception tasks, from highest to lowest. Rows corresponding to proprietary models are highlighted in gray for distinction.}\label{tab:res_ocr}
\resizebox{\textwidth}{!}{%
\begin{tabular}{llccccccccccc}
\toprule
\multicolumn{1}{c}{\multirow{2}{*}{\textbf{Method}}} & \multicolumn{1}{c}{\multirow{2}{*}{\textbf{LLM}}} & \multicolumn{6}{c}{\textbf{Perception}} & \multicolumn{5}{c}{\textbf{Reasoning}} \\ \cmidrule{3-9}\cmidrule{10-13}
\multicolumn{1}{c}{} & \multicolumn{1}{c}{} & \textbf{Product} & \textbf{B \& M \& P} & \textbf{Contact} & \textbf{Identity} & \textbf{Signage} & \textbf{Avg} & \textbf{Avg-C} & \textbf{Scene} & \textbf{Character} & \textbf{Avg} & \textbf{Avg-C} \\ \midrule \Gray
GPT-4o & - & 79.65 & 79.23 & 74.88 & 73.66 & 77.38 & 77.69 & 76.96 & 64.80 & 58.00 & 61.40 & 61.40 \\
InternVL-2 & InternLM2.5-7b-Chat & 72.21 & 80.58 & 72.10 & 73.47 & 68.70 & 73.92 & 73.41 & 56.00 & 58.80 & 57.40 & 57.40 \\ \Gray
Claude 3.5 Sonnet & - & 71.89 & 83.67 & 61.15 & 64.64 & 69.70 & 72.47 & 70.21 & 62.60 & 61.20 & 61.90 & 61.90 \\
InternVL-Chat-V1-5 & InternLM2-Chat-20B & 69.83 & 75.56 & 71.75 & 73.36 & 67.03 & 71.51 & 71.51 & 57.60 & 56.00 & 56.80 & 56.80 \\
CogVLm2-llama3-Chat & LLama3-8B & 70.35 & 66.82 & 74.00 & 76.41 & 67.03 & 69.97 & 70.92 & 58.80 & 49.20 & 54.00 & 54.00 \\
Mini-Gemini-34B-HD & Nous-Hermes-2-Yi-34B & 72.14 & 79.04 & 64.30 & 55.16 & 66.61 & 69.55 & 67.45 & 60.80 & 57.60 & 59.20 & 59.20 \\
InternLM-XComposer2.5 & InternLM2-7B & 65.34 & 71.83 & 66.90 & 78.29 & 65.69 & 69.25 & 69.61 & 50.40 & 56.40 & 53.40 & 53.40 \\ \Gray
Gemini-1.5-pro & - & 65.92 & 66.37 & 74.41 & 70.19 & 66.36 & 67.62 & 68.65 & 56.60 & 48.80 & 52.70 & 52.70 \\
MiniCPM-V 2.5 & LLama3-8B & 64.69 & 69.52 & 71.58 & 68.90 & 62.19 & 66.79 & 67.38 & 48.80 & 39.20 & 44.00 & 44.00 \\
Cambrian-1-34B & Nous-Hermes-2-Yi-34B & 67.65 & 77.30 & 62.22 & 50.47 & 64.19 & 66.45 & 64.37 & 54.00 & 56.00 & 55.00 & 55.00 \\ \Gray
GPT-4o-mini & - & 62.32 & 68.87 & 54.11 & 62.56 & 58.51 & 62.51 & 61.27 & 52.80 & 41.20 & 47.00 & 47.00 \\
Cambrian-1-8B & LLama3-8B-Instruct & 59.18 & 67.27 & 55.98 & 52.93 & 52.25 & 58.68 & 57.52 & 52.80 & 53.60 & 53.20 & 53.20 \\
Monkey & Qwen-7B & 55.58 & 54.47 & 53.38 & 59.98 & 50.42 & 54.63 & 54.77 & 32.40 & 22.00 & 27.20 & 27.20 \\
SliME-8B & LLama3-8B & 55.97 & 57.30 & 41.25 & 55.05 & 49.92 & 53.45 & 51.90 & 55.60 & 50.80 & 53.20 & 53.20 \\
mPLUG-DocOwl 1.5 & LLaMA-7B & 54.62 & 52.22 & 59.10 & 63.03 & 32.99 & 51.15 & 52.39 & 46.80 & 38.40 & 42.60 & 42.60 \\
SliME-13B & Vicuna-13B & 52.25 & 46.50 & 46.97 & 53.76 & 53.17 & 50.58 & 50.53 & 45.60 & 36.40 & 41.00 & 41.00 \\
DeepSeek-VL & DeepSeek-LLM-7b-base & 53.72 & 55.06 & 31.72 & 44.13 & 49.42 & 49.55 & 46.81 & 48.80 & 41.60 & 45.20 & 45.20 \\
LLaVA-Next & LLama3-8B & 50.77 & 49.45 & 38.30 & 38.73 & 53.51 & 47.94 & 46.15 & 58.00 & 52.40 & 55.20 & 55.20 \\
YI-VL-34B & Yi-34B-Chat & 48.33 & 50.55 & 33.97 & 37.91 & 43.57 & 44.95 & 42.87 & 46.80 & 38.00 & 42.40 & 42.40 \\
ShareGPT4V-13B & Vicuna-13B & 46.92 & 40.51 & 40.38 & 46.01 & 47.66 & 44.55 & 44.30 & 31.20 & 20.80 & 26.00 & 26.00 \\
LLaVA1.5-13B & Vicuna-13B & 45.51 & 40.39 & 39.69 & 47.54 & 46.74 & 44.10 & 43.97 & 36.80 & 23.60 & 30.20 & 30.20 \\
Mini-Gemini-7B-HD & Vicuna-7B-v1.5 & 40.05 & 44.44 & 42.98 & 44.37 & 39.32 & 42.02 & 42.23 & 38.00 & 32.80 & 35.40 & 35.40 \\
ShareGPT4V-7B & Vicuna-7B & 40.50 & 35.43 & 37.61 & 42.14 & 41.99 & 39.39 & 39.53 & 25.60 & 22.70 & 24.15 & 24.15 \\
MiniGPT-v2 & Llama 2-7B-Chat & 40.05 & 34.73 & 38.99 & 41.08 & 41.82 & 39.02 & 39.33 & 36.40 & 23.60 & 30.00 & 30.00 \\
LLaVA1.5-7B & Vicuna-7B & 39.67 & 34.92 & 37.26 & 40.26 & 41.90 & 38.69 & 38.80 & 30.80 & 21.20 & 26.00 & 26.00 \\
TextMonkey & Qwen-7B & 38.12 & 31.96 & 44.89 & 45.19 & 33.89 & 37.30 & 38.81 & 36.00 & 24.80 & 30.40 & 30.40 \\
LLaVA-Next & Qwen-72B & 43.58 & 45.72 & 20.28 & 14.91 & 41.24 & 37.07 & 33.15 & 17.60 & 16.80 & 17.20 & 17.20 \\
Qwen-VL-Chat & Qwen-7B & 32.73 & 37.62 & 27.38 & 33.22 & 26.88 & 32.37 & 31.57 & 36.40 & 20.80 & 28.60 & 28.60 \\ \bottomrule
\end{tabular}%
}
\end{table}

\textbf{Diagram and Table.} Tab.~\ref{tab:res_tab_dia} presents the results for the diagram domain, where some of the more advanced models perform relatively well, with three models achieving an average accuracy of over $60\%$. Reasoning tasks, however, have proved to be more challenging. Only Claude 3.5 Sonnet manage to exceed $60\%$ accuracy, standing out significantly, with the second-ranked InternLM-XComposer2.5 trailing by 20\%. Additionally, models like LLaVA-Next, which have performed well on existing benchmarks like chartQA, show noticeably weaker performance on our dataset, underscoring the higher difficulty of the our benchmark.

\begin{table}[]
\caption{\textbf{Experimental results on the \texttt{Diagram and Table} tasks}. Models are ranked according to their average performance. Rows corresponding to proprietary models are highlighted in gray for distinction.}\label{tab:res_tab_dia}
\resizebox{\textwidth}{!}{%
\begin{tabular}{llcccccccc}
\toprule
\multicolumn{1}{c}{\multirow{2}{*}{\textbf{Method}}} & \multicolumn{1}{c}{\multirow{2}{*}{\textbf{LLM}}} & \multicolumn{4}{c}{\textbf{Perception}} & \multicolumn{4}{c}{\textbf{Reasoning}} \\ \cmidrule{3-10}
\multicolumn{1}{c}{} & \multicolumn{1}{c}{} & \textbf{Diagram} & \textbf{Table} & \textbf{Avg} & \textbf{Avg-C} & \textbf{Diagram} & \textbf{Table} & \textbf{Avg} & \textbf{Avg-C} \\ \hline \Gray
Claude 3.5 Sonnet & - & 71.31 & 66.08 & 67.44 & 68.70 & 60.92 & 61.35 & 61.20 & 61.14 \\
InternLM-XComposer2.5 & InternLM2-7B & 69.05 & 62.12 & 63.92 & 65.59 & 43.10 & 39.88 & 41.00 & 41.49 \\
InternVL-2 & InternLM2.5-7B-Chat & 68.83 & 60.68 & 62.80 & 64.76 & 42.53 & 37.12 & 39.00 & 39.83 \\
InternVL-Chat-V1-5 & InternLM2-Chat-20B & 61.55 & 53.81 & 55.83 & 57.68 & 37.36 & 34.36 & 35.40 & 35.86 \\
MiniCPM-V 2.5 & LLama3-8B & 57.81 & 51.05 & 52.81 & 54.43 & 26.44 & 34.66 & 31.80 & 30.55 \\
CogVLm2-llama3-Chat & LLama3-8B & 51.52 & 46.09 & 47.51 & 48.81 & 31.61 & 33.44 & 32.80 & 32.53 \\ \Gray
GPT-4o & - & 47.35 & 46.44 & 46.68 & 46.90 & 44.25 & 45.09 & 44.80 & 44.67 \\
Mini-Gemini-34B-HD & Nous-Hermes-2-Yi-34B & 47.63 & 43.21 & 44.36 & 45.42 & 38.51 & 39.57 & 39.20 & 39.04 \\ \Gray
GPT-4o-mini & - & 46.22 & 43.54 & 44.23 & 44.88 & 38.51 & 40.49 & 39.80 & 39.50 \\
Cambrian-1-34B & Nous-Hermes-2-Yi-34B & 43.67 & 39.30 & 40.44 & 41.49 & 37.93 & 34.97 & 36.00 & 36.45 \\ \Gray
Gemini-1.5-pro & - & 41.41 & 39.37 & 39.90 & 40.39 & 35.63 & 31.90 & 33.20 & 33.77 \\
Cambrian-1-8B & LLama3-8B-Instruct & 36.25 & 31.48 & 32.73 & 33.87 & 27.59 & 27.30 & 27.40 & 27.45 \\
Monkey & Qwen-7B & 34.98 & 31.63 & 32.51 & 33.31 & 18.39 & 22.09 & 20.80 & 20.24 \\
mPLUG-DocOwl 1.5 & LLama-7B & 30.74 & 28.85 & 29.34 & 29.80 & 18.39 & 20.55 & 19.80 & 19.47 \\
SliME-8B & LLama3-8B & 29.75 & 29.19 & 29.34 & 29.47 & 29.89 & 29.14 & 29.40 & 29.52 \\
LLaVA-Next & Qwen-72B & 27.77 & 27.65 & 27.68 & 27.71 & 36.78 & 32.82 & 34.20 & 34.80 \\
LLaVA-Next & LLama3-8B & 26.64 & 26.63 & 26.63 & 26.64 & 22.99 & 23.62 & 23.40 & 23.31 \\
DeepSeek-VL & DeepSeek-LLM-7b-base & 23.67 & 23.27 & 23.38 & 23.47 & 22.99 & 24.23 & 23.80 & 23.61 \\
Mini-Gemini-7B-HD & Vicuna-7B-v1.5 & 21.20 & 22.70 & 22.31 & 21.95 & 27.01 & 23.31 & 24.60 & 25.16 \\
SliME-13B & Vicuna-13B & 19.28 & 21.40 & 20.93 & 20.34 & 38.51 & 39.26 & 39.00 & 38.89 \\
MiniGPT-v2 & Llama 2-7B-Chat & 18.59 & 21.06 & 20.41 & 19.83 & 20.69 & 20.25 & 20.40 & 20.47 \\
LLaVA1.5-13B & Vicuna-13B & 18.30 & 20.83 & 20.17 & 19.57 & 22.41 & 19.94 & 20.80 & 21.18 \\
ShareGPT4V-13B & Vicuna-13B & 18.37 & 20.81 & 20.17 & 19.59 & 21.84 & 20.25 & 20.80 & 21.05 \\
LLaVA1.5-7B & Vicuna-7B & 18.30 & 20.71 & 20.08 & 19.51 & 21.84 & 19.94 & 20.60 & 20.89 \\
ShareGPT4V-7B & Vicuna-7B & 18.30 & 20.71 & 20.08 & 19.51 & 21.84 & 19.94 & 20.60 & 20.89 \\
YI-VL-34B & Yi-34B-Chat & 15.90 & 16.03 & 15.99 & 15.97 & 23.56 & 27.30 & 26.00 & 25.43 \\
Qwen-VL-Chat & Qwen-7B & 18.42 & 14.56 & 15.59 & 16.49 & 14.94 & 12.88 & 13.60 & 13.91 \\
TextMonkey & Qwen-7B & 6.71 & 5.65 & 5.93 & 6.18 & 3.45 & 1.53 & 2.20 & 2.49 \\ \bottomrule
\end{tabular}%
}
\end{table}

\textbf{Remote Sensing.} Tab.~\ref{tab:res_remote} presents the performance of various models on remote sensing tasks. It is evident that models performing well on remote sensing data typically either employ special handling for high-resolution images (e.g., Mini-Gemini-HD, SliME, Cambrian) or have vision encoders designed to support high-resolution inputs (e.g., InternVL). Among these, SliME achieves the highest performance due to its support for the largest resolution. However, even the top-performing model, SliME-8B, shows poor performance on counting tasks with extremely large images, with only $30\%$ accuracy. Some closed-source models perform even worse, with GPT-4o-mini achieving only $2\%$ accuracy. This highlights the high demands of remote sensing data on resolution and detail perception.

\begin{table}[]
\centering
\caption{\textbf{Experimental results on the \texttt{Remote Sensing} tasks.} Models are ranked according to their average performance. Rows corresponding to proprietary models are highlighted in gray for distinction.}\label{tab:res_remote}
\resizebox{0.8\textwidth}{!}{%
\begin{tabular}{llccccc}
\toprule
\multicolumn{1}{c}{\textbf{Method}} & \multicolumn{1}{c}{\textbf{LLM}} & \textbf{Color} & \textbf{Count} & \textbf{Position} & \textbf{Avg} & \textbf{Avg-C} \\ \hline 
SliME-8B & LLama3-8B & 45.66 & 28.63 & 52.19 & 42.27 & 42.16 \\
Mini-Gemini-34B-HD & Nous-Hermes-2-Yi-34B & 41.12 & 21.29 & 58.31 & 40.40 & 40.24 \\
Cambrian-1-8B & LLama3-8B-Instruct & 38.01 & 20.55 & 61.10 & 40.05 & 39.89 \\
InternVL-2 & InternLM2.5-7B-Chat & 47.41 & 25.69 & 44.63 & 39.35 & 39.24 \\
Cambrian-1-34B & Nous-Hermes-2-Yi-34B & 37.05 & 22.76 & 55.69 & 38.63 & 38.50 \\
InternLM-XComposer2.5 & InternLM2-7B & 45.34 & 17.62 & 44.95 & 36.12 & 35.97 \\
InternVL-Chat-V1-5 & InternLM2-Chat-20B & 34.10 & 17.86 & 48.29 & 33.55 & 33.42 \\
YI-VL-34B & Yi-34B-Chat & 34.02 & 19.00 & 41.53 & 31.62 & 31.52 \\
Mini-Gemini-7B-HD & Vicuna-7B-v1.5 & 37.29 & 22.43 & 33.97 & 31.30 & 31.23 \\
LLaVA-Next & Qwen-72B & 33.86 & 23.08 & 30.31 & 29.13 & 29.08 \\ \Gray
GPT-4o & - & 34.18 & 15.17 & 37.07 & 28.92 & 28.81 \\
CogVLm2-llama3-Chat & LLama3-8B & 37.69 & 18.35 & 29.99 & 28.76 & 28.68 \\
MiniCPM-V 2.5 & LLama3-8B & 37.69 & 11.50 & 33.49 & 27.69 & 27.56 \\
SliME-13B & Vicuna-13B & 30.04 & 18.43 & 28.80 & 25.82 & 25.76 \\ \Gray
Claude 3.5 Sonnet & - & 31.87 & 18.11 & 27.95 & 25.74 & 25.98 \\
DeepSeek-VL & DeepSeek-LLM-7b-base & 29.40 & 7.34 & 39.30 & 25.49 & 25.35 \\
LLaVA-Next & LLama3-8B & 30.04 & 20.55 & 25.74 & 25.42 & 25.44 \\
Monkey & Qwen-7B & 22.07 & 16.97 & 35.72 & 24.99 & 24.92 \\
mPLUG-DocOwl 1.5 & LLaMA-7B & 27.81 & 16.39 & 26.81 & 23.71 & 23.67 \\
MiniGPT-v2 & Llama 2-7B-Chat & 23.35 & 20.15 & 26.41 & 23.33 & 23.30 \\
LLaVA1.5-13B & Vicuna-13B & 26.22 & 16.88 & 26.57 & 23.27 & 23.22 \\
ShareGPT4V-13B & Vicuna-13B & 25.58 & 16.97 & 26.49 & 23.06 & 23.01 \\
LLaVA1.5-7B & Vicuna-7B & 23.11 & 16.88 & 26.25 & 22.12 & 22.08 \\
ShareGPT4V-7B & Vicuna-7B & 23.03 & 16.88 & 26.25 & 22.10 & 22.05 \\
Qwen-VL-Chat & Qwen-7B & 16.97 & 11.50 & 16.87 & 15.14 & 15.11 \\ \Gray
Gemini-1.5-pro & - & 13.39 & 8.32 & 20.13 & 13.99 & 13.95 \\
TextMonkey & Qwen-7B & 6.93 & 2.04 & 25.86 & 11.69 & 11.61 \\ \Gray
GPT-4o-mini & - & 5.82 & 2.61 & 11.54 & 6.69 & 6.66 \\ \bottomrule
\end{tabular}%
}
\end{table}

\textbf{Autonomous Driving.} Tab.~\ref{tab:ad_perception} and Tab.~\ref{tab:ad_reasoning} show the perception and reasoning performance of various models in autonomous driving scenarios. As a critical application area, autonomous driving remains a challenge for MLLMs, with no model currently capable of reliably addressing tasks such as intent prediction, traffic light recognition, and object counting solely through text. Only Claude 3.5 Sonnet achieve an average perception accuracy exceeding $40\%$. Reasoning tasks are even more difficult, with even the most advanced models achieving only around $30\%$ accuracy. Autonomous driving is inherently a high-risk domain that demands very high accuracy for practical deployment. This indicates that more powerful multimodal models with 3D spatial prediction and understanding ability, or specialized fine-tuning on domain-specific datasets for driving expertise, are necessary before MLLMs can be effectively applied in this field.

\begin{table}[]
\caption{\textbf{Experimental results on the \texttt{Autonomous Driving} perception tasks.} Models are ranked according to their average performance. Rows corresponding to proprietary models are highlighted in gray for distinction.}\label{tab:ad_perception}

\resizebox{\textwidth}{!}{%
\begin{tabular}{llccccccccc}
\toprule
\multicolumn{1}{c}{} & \multicolumn{1}{c}{} &  & \multicolumn{4}{c}{\textbf{Motion}} &  &  &  &  \\ \cmidrule{4-7}
\multicolumn{1}{c}{\multirow{-2}{*}{\textbf{Method}}} & \multicolumn{1}{c}{\multirow{-2}{*}{\textbf{LLM}}} & \multirow{-2}{*}{\textbf{Identity}} & \textbf{Vehicle} & \textbf{Multi-vehicle} & \textbf{Pedestrain} & \textbf{Multi-pedestrain} & \multirow{-2}{*}{\textbf{Traffic Signal}} & \multirow{-2}{*}{\textbf{Object Count}} & \multirow{-2}{*}{\textbf{Avg}} & \multirow{-2}{*}{\textbf{Avg-C}} \\ \hline
Claude 3.5 Sonnet & - & 58.66 & 18.45 & 35.48 & 32.32 & 31.64 & 37.31 & 33.19 & 40.77 & 49.72 \\
Cambrian-1-8B & LLama3-8B-Instruct & 56.77 & 50.00 & 33.29 & 32.32 & 14.00 & 35.82 & 33.06 & 38.52 & 47.65 \\
InternVL-2 & InternLM2.5-7B-Chat & 46.68 & 39.24 & 30.98 & 34.76 & 17.24 & 37.81 & 34.58 & 35.46 & 41.07 \\
MiniCPM-V 2.5 & LLama3-8B & 44.96 & 41.77 & 30.86 & 31.71 & 19.88 & 37.31 & 29.17 & 34.15 & 39.56 \\
SliME-8B & LLama3-8B & 44.50 & 51.90 & 28.68 & 29.27 & 15.21 & 29.35 & 33.61 & 33.66 & 39.08 \\
InternLM-XComposer2.5 & InternLM2-7B & 46.23 & 48.10 & 26.61 & 32.93 & 11.76 & 40.30 & 32.50 & 33.63 & 39.93 \\
Cambrian-1-34B & Nous-Hermes-2-Yi-34B & 43.96 & 38.61 & 31.96 & 32.93 & 12.37 & 27.36 & 33.89 & 33.39 & 38.68 \\
DeepSeek-VL & DeepSeek-LLM-7b-base & 44.05 & 63.29 & 25.88 & 36.59 & 18.05 & 39.30 & 27.22 & 33.39 & 38.72 \\
Mini-Gemini-34B-HD & Nous-Hermes-2-Yi-34B & 39.78 & 42.41 & 36.09 & 31.10 & 16.63 & 17.41 & 31.53 & 32.70 & 36.24 \\
InternVL-Chat-V1-5 & InternLM2-Chat-20B & 40.42 & 39.87 & 26.97 & 27.44 & 18.66 & 32.34 & 30.14 & 31.42 & 35.92 \\
CogVLm2-llama3-Chat & LLama3-8B & 33.15 & 36.71 & 29.77 & 31.71 & 19.27 & 43.78 & 28.19 & 30.22 & 31.69 \\
Monkey & Qwen-7B & 35.97 & 60.76 & 24.30 & 37.20 & 18.26 & 32.34 & 24.72 & 29.67 & 32.82 \\
YI-VL-34B & Yi-34B-Chat & 36.24 & 41.77 & 29.60 & 31.71 & 20.89 & 16.92 & 19.17 & 28.31 & 32.28 \\
mPLUG-DocOwl 1.5 & LLama-7B & 26.74 & 60.76 & 24.79 & 31.10 & 22.72 & 43.28 & 26.53 & 28.28 & 27.51 \\
SliME-13B & Vicuna-13B & 26.61 & 46.84 & 24.54 & 32.32 & 17.65 & 43.28 & 27.50 & 27.16 & 26.89 \\
Gemini-1.5-pro & - & 32.61 & 10.13 & 30.23 & 8.54 & 16.02 & 10.45 & 31.31 & 26.64 & 29.63 \\
LLaVA1.5-13B & Vicuna-13B & 23.25 & 31.65 & 24.91 & 31.10 & 25.96 & 36.32 & 26.80 & 26.12 & 24.69 \\
ShareGPT4V-13B & Vicuna-13B & 23.25 & 31.01 & 24.91 & 31.10 & 25.96 & 36.82 & 26.81 & 26.12 & 24.69 \\
LLaVA1.5-7B & Vicuna-7B & 23.25 & 31.01 & 24.91 & 31.10 & 25.96 & 35.32 & 26.81 & 26.04 & 24.65 \\
ShareGPT4V-7B & Vicuna-7B & 23.25 & 31.01 & 24.91 & 31.10 & 25.96 & 35.32 & 26.81 & 26.04 & 24.65 \\
MiniGPT-v2 & Llama 2-7B-Chat & 23.71 & 53.16 & 22.36 & 28.66 & 20.49 & 35.32 & 28.06 & 25.96 & 24.84 \\
Mini-Gemini-7B-HD & Vicuna-7B-v1.5 & 27.25 & 60.76 & 23.57 & 26.83 & 14.81 & 36.32 & 17.78 & 24.81 & 26.03 \\
GPT-4o-mini & - & 19.07 & 45.57 & 24.67 & 23.78 & 11.36 & 40.30 & 31.11 & 24.18 & 21.63 \\
GPT-4o & - & 15.26 & 23.42 & 25.39 & 26.22 & 9.94 & 41.29 & 32.22 & 22.43 & 18.85 \\
LLaVA-Next & LLama3-8B & 21.44 & 41.77 & 22.36 & 29.88 & 9.23 & 22.39 & 8.06 & 18.66 & 20.05 \\
LLaVA-Next & Qwen-72B & 19.26 & 26.58 & 26.37 & 29.88 & 12.58 & 16.42 & 5.97 & 17.98 & 18.62 \\
Qwen-VL-Chat & Qwen & 9.26 & 35.44 & 15.43 & 23.17 & 8.32 & 34.83 & 16.39 & 15.08 & 12.17 \\
TextMonkey & Qwen-7B & 8.54 & 37.34 & 22.72 & 15.85 & 16.23 & 14.93 & 6.39 & 14.26 & 11.40 \\ \bottomrule
\end{tabular}%
}
\end{table}

\begin{table}[]
\caption{\textbf{Experimental results on the \texttt{Autonomous Driving} reasoning tasks.} Models are ranked according to their average performance on perception tasks, from highest to lowest. Rows corresponding to proprietary models are highlighted in gray for distinction.}\label{tab:ad_reasoning}
\resizebox{\textwidth}{!}{%
\begin{tabular}{llcccccccccc}
\toprule
\multicolumn{1}{c}{\multirow{2}{*}{\textbf{Method}}} & \multicolumn{1}{c}{\multirow{2}{*}{\textbf{LLM}}} & \multicolumn{3}{c}{\textbf{Intention}} & \multicolumn{4}{c}{\textbf{Relation}} & \textbf{Attention} & \multirow{2}{*}{\textbf{Avg}} & \multirow{2}{*}{\textbf{Avg-C}} \\ \cmidrule{6-9}
\multicolumn{1}{c}{} & \multicolumn{1}{c}{} & \textbf{Ego} & \textbf{Pedestrian} & \textbf{Verhicle} & \textbf{Ego2P} & \textbf{Ego2T} & \textbf{Ego2V} & \textbf{O2O} & \textbf{Signal} &  &  \\ \hline
Monkey & Qwen-7B & 28.62 & 56.31 & 30.43 & 27.36 & 22.86 & 32.67 & 11.94 & 58.06 & 33.04 & 33.48 \\ \Gray
Claude 3.5 Sonnet & - & 26.32 & 32.04 & 24.64 & 23.58 & 25.71 & 20.79 & 24.38 & 65.90 & 31.92 & 30.59 \\
YI-VL-34B & Yi-34B-Chat & 28.26 & 46.60 & 33.33 & 21.70 & 24.76 & 31.68 & 15.42 & 49.77 & 31.55 & 31.45 \\
SliME-8B & LLama3-8B & 28.29 & 39.81 & 33.33 & 24.53 & 19.05 & 22.77 & 10.45 & 63.59 & 31.55 & 30.37 \\
CogVLm2-llama3-Chat & LLama3-8B & 30.26 & 25.24 & 25.60 & 35.85 & 20.95 & 28.71 & 18.41 & 56.22 & 31.18 & 30.27 \\
MiniCPM-V 2.5 & LLama3-8B & 24.01 & 37.86 & 31.88 & 20.75 & 30.48 & 15.84 & 26.87 & 53.00 & 31.03 & 30.19 \\
SliME-13B & Vicuna-13B & 25.00 & 41.75 & 28.99 & 28.30 & 21.90 & 24.75 & 25.87 & 48.39 & 30.80 & 30.64 \\
LLaVA-Next & LLama3-8B & 32.89 & 49.51 & 33.82 & 28.30 & 25.71 & 24.75 & 7.96 & 43.32 & 30.73 & 30.78 \\
Cambrian-1-8B & LLama3-8B-Instruct & 25.00 & 41.75 & 35.27 & 23.58 & 23.81 & 16.83 & 11.44 & 60.37 & 30.73 & 29.86 \\
InternVL-2 & InternLM2.5-7B-Chat & 24.01 & 43.69 & 32.85 & 22.64 & 28.57 & 21.78 & 21.89 & 43.78 & 29.84 & 29.89 \\
LLaVA-Next & Qwen-72B & 30.59 & 52.43 & 35.27 & 23.58 & 27.62 & 29.70 & 8.96 & 35.48 & 29.69 & 30.37 \\
InternVL-Chat-V1-5 & InternLM2-Chat-20B & 25.99 & 32.04 & 31.88 & 16.98 & 22.86 & 25.74 & 9.45 & 57.14 & 28.94 & 27.89 \\
DeepSeek-VL & DeepSeek-LLM-7b-base & 30.26 & 17.48 & 27.05 & 22.64 & 25.71 & 24.75 & 6.97 & 51.15 & 27.31 & 25.92 \\ \Gray
GPT-4o-mini & - & 11.51 & 19.42 & 24.64 & 22.64 & 28.57 & 17.82 & 31.34 & 54.84 & 26.79 & 26.40 \\ \Gray
GPT-4o & - & 17.11 & 19.42 & 27.54 & 15.09 & 20.00 & 22.77 & 16.92 & 60.83 & 26.41 & 25.12 \\
mPLUG-DocOwl 1.5 & LLama-7B & 20.72 & 26.21 & 30.43 & 19.81 & 31.43 & 25.74 & 12.94 & 41.94 & 26.04 & 26.14 \\
LLaVA1.5-13B & Vicuna-13B & 23.36 & 18.45 & 24.15 & 26.42 & 23.81 & 22.77 & 25.37 & 30.41 & 24.78 & 24.39 \\
Qwen-VL-Chat & Qwen & 20.39 & 21.36 & 20.77 & 16.04 & 23.81 & 17.82 & 16.92 & 50.69 & 24.63 & 23.60 \\
ShareGPT4V-13B & Vicuna-13B & 23.36 & 17.48 & 26.09 & 25.47 & 27.63 & 22.77 & 25.37 & 26.27 & 24.55 & 24.33 \\
Cambrian-1-34B & Nous-Hermes-2-Yi-34B & 14.14 & 24.27 & 22.71 & 24.53 & 30.48 & 31.68 & 10.95 & 46.54 & 24.40 & 25.52 \\
LLaVA1.5-7B & Vicuna-7B & 23.36 & 17.48 & 26.09 & 25.47 & 27.62 & 22.77 & 25.37 & 23.96 & 24.18 & 24.03 \\
ShareGPT4V-7B & Vicuna-7B & 23.36 & 17.48 & 26.09 & 25.47 & 27.62 & 22.77 & 25.37 & 23.96 & 24.18 & 24.03 \\
InternLM-XComposer2.5 & InternLM2-7B & 25.33 & 36.89 & 34.30 & 17.92 & 26.67 & 25.74 & 23.88 & 44.24 & 24.03 & 28.78 \\
MiniGPT-v2 & Llama 2-7B-Chat & 23.68 & 25.24 & 28.02 & 28.30 & 22.86 & 21.78 & 2.49 & 37.33 & 23.66 & 23.71 \\
Mini-Gemini-7B-HD & Vicuna-7B-v1.5 & 21.05 & 6.80 & 14.49 & 19.81 & 15.24 & 25.74 & 15.42 & 54.38 & 23.29 & 21.80 \\
Mini-Gemini-34B-HD & Nous-Hermes-2-Yi-34B & 14.14 & 22.33 & 21.74 & 23.58 & 31.43 & 30.69 & 10.45 & 47.00 & 22.84 & 24.91 \\
TextMonkey & Qwen-7B & 9.54 & 24.27 & 23.67 & 24.53 & 17.14 & 20.79 & 11.44 & 35.94 & 20.01 & 20.81 \\ \Gray
Gemini-1.5-pro & - & 13.49 & 18.45 & 28.02 & 6.60 & 6.67 & 6.93 & 23.88 & 32.72 & 19.20 & 17.33 \\ \bottomrule
\end{tabular}%
}
\end{table}

\textbf{Monitoring Performance}. Tab.~\ref{tab:res_monitoring} presents the performance of various models under monitoring scenarios. As can be observed, the monitoring task poses a high degree of difficulty. Traditional models like Qwen-VL and LLaVA have an accuracy rate of around $20\%$, which is nearly equivalent to random guessing. Open-source models significantly outperform closed-source models. For instance, InternVL-2 has an average accuracy rate of $53.19$ on perception tasks, greatly surpassing GPT-4o's $33.93$. We notice that closed-source models such as GPT-4o have a high frequency of answering “E”, with over $35\%$ of responses choosing “E”. This suggests that closed-source models may be more inclined to refrain from responding when the answer is uncertain. Furthermore, we find that while most models perform reasonably well on counting tasks, they struggle with tasks related to spatial relationship judgment and attribute recognition. Moreover, related reasoning tasks also pose a high level of difficulty, with no model achieving an accuracy rate over $40\%$ to date. In combination with the results from autonomous driving tasks, we observe that MLLMs exhibit significant deficiencies in understanding, predicting, and reasoning about the dynamic information of objects in 2D or 3D space. Although the input to these models is a single frame image rather than a temporal sequence of video frames, there remains a considerable gap between their performance and that of humans. For humans, who possess rich experiential knowledge of dynamics, it is not difficult to infer the future states of objects from a single image in unambiguous situations. Therefore, modern MLLMs are still far from having the capability to function as world models.

\begin{table}[]
\caption{\textbf{Experimental results on the \texttt{Monitoring} tasks}. Models are ranked according to their average performance. Rows corresponding to proprietary models are highlighted in gray for distinction.}\label{tab:res_monitoring}
\resizebox{\textwidth}{!}{%
\begin{tabular}{llcccccccccccc}
\toprule
\multicolumn{1}{c}{\multirow{3}{*}{\textbf{Method}}} & \multicolumn{1}{c}{\multirow{3}{*}{\textbf{LLM}}} & \multicolumn{7}{c}{\textbf{Perception}} & \multicolumn{5}{c}{\textbf{Reasoning}} \\ \cmidrule{3-14}
\multicolumn{1}{c}{} & \multicolumn{1}{c}{} & \multicolumn{3}{c}{\textbf{Vehicle}} & \multicolumn{2}{c}{\textbf{Pedestrain}} & \multirow{2}{*}{\textbf{Avg}} & \multirow{2}{*}{\textbf{Avg-C}} & \multirow{2}{*}{\textbf{Calculate}} & \multirow{2}{*}{\textbf{Intention}} & \multirow{2}{*}{\textbf{Property}} & \multirow{2}{*}{\textbf{Avg}} & \multicolumn{1}{l}{\multirow{2}{*}{\textbf{Avg-C}}} \\ \cmidrule{3-7}
\multicolumn{1}{c}{} & \multicolumn{1}{c}{} & \multicolumn{1}{l}{\textbf{Counting}} & \multicolumn{1}{l}{\textbf{Location}} & \multicolumn{1}{l}{\textbf{Attribute}} & \multicolumn{1}{l}{\textbf{Counting}} & \multicolumn{1}{l}{\textbf{Attribute}} &  &  &  &  &  &  & \multicolumn{1}{l}{} \\\midrule
InternVL-2 & InternLM2.5-7B-Chat & 70.07 & 25.74 & 28.98 & 59.68 & 12.04 & 53.19 & 41.62 & 51.67 & 21.43 & 41.00 & 43.57 & 38.03 \\
InternVL-Chat-V1-5 & InternLM2-Chat-20B & 72.53 & 23.53 & 27.27 & 55.24 & 7.41 & 51.16 & 39.52 & 39.33 & 26.53 & 42.00 & 37.35 & 35.95 \\
Cambrian-1-8B & LLama3-8B-Instruct & 62.01 & 29.41 & 20.45 & 55.44 & 7.41 & 47.68 & 37.07 & 46.00 & 29.59 & 44.00 & 42.37 & 39.86 \\
Cambrian-1-34B & Nous-Hermes-2-Yi-34B & 51.32 & 33.09 & 26.14 & 55.14 & 12.96 & 45.98 & 37.44 & 11.33 & 18.37 & 45.00 & 19.48 & 24.90 \\
SliME-8B & LLama3-8B & 60.53 & 33.82 & 28.98 & 34.48 & 31.48 & 40.62 & 38.32 & 32.33 & 40.82 & 43.00 & 36.14 & 38.72 \\
Mini-Gemini-34B-HD & Nous-Hermes-2-Yi-34B & 53.95 & 17.65 & 22.73 & 43.45 & 7.41 & 39.61 & 30.80 & 11.67 & 17.35 & 50.00 & 20.48 & 26.34 \\
InternLM-XComposer2.5 & InternLM2-7B & 52.63 & 13.24 & 17.61 & 46.98 & 0.93 & 39.48 & 28.48 & 13.67 & 13.27 & 34.00 & 17.67 & 20.31 \\
MiniCPM-V 2.5 & LLama3-8B & 62.66 & 16.91 & 22.73 & 36.49 & 4.63 & 38.70 & 30.35 & 36.00 & 35.71 & 41.00 & 36.95 & 37.57 \\
YI-VL-34B & Yi-34B-Chat & 56.25 & 8.09 & 23.86 & 32.47 & 5.56 & 34.85 & 26.85 & 28.00 & 28.57 & 44.00 & 31.33 & 33.52 \\
Mini-Gemini-7B-HD & Vicuna-7B-v1.5 & 47.86 & 13.97 & 25.00 & 34.07 & 13.89 & 34.15 & 28.16 & 21.00 & 24.49 & 42.00 & 25.90 & 29.16 \\ \Gray
GPT-4o & - & 50.66 & 15.44 & 19.89 & 34.17 & 6.48 & 33.93 & 26.76 & 4.00 & 13.27 & 41.00 & 19.42 & 19.42 \\
CogVLm2-llama3-Chat & LLama3-8B & 48.19 & 26.47 & 22.16 & 32.36 & 12.04 & 33.74 & 29.16 & 40.00 & 40.82 & 45.00 & 41.16 & 41.94 \\ \Gray
Claude 3.5 Sonnet & - & 50.99 & 33.77 & 11.93 & 33.37 & 8.33 & 32.19 & 28.43 & 34.37 & 18.37 & 44.00 & 32.25 & 32.25 \\ \Gray
Gemini-1.5-pro & - & 52.63 & 9.56 & 10.80 & 31.05 & 10.08 & 31.11 & 24.21 & 11.67 & 13.27 & 27.00 & 17.31 & 17.31 \\
LLaVA-Next & Qwen-72B & 57.89 & 10.39 & 27.27 & 15.32 & 28.70 & 29.37 & 28.16 & 23.33 & 38.78 & 28.00 & 27.31 & 30.04 \\
Monkey & Qwen-7B & 42.76 & 40.44 & 21.02 & 21.77 & 9.26 & 28.01 & 27.21 & 25.33 & 21.43 & 39.00 & 27.31 & 28.59 \\
DeepSeek-VL & DeepSeek-LLM-7b-base & 43.91 & 7.35 & 17.33 & 24.70 & 9.26 & 26.97 & 21.59 & 5.33 & 19.39 & 48.00 & 16.67 & 24.24 \\ \Gray
GPT-4o-mini & - & 44.57 & 8.82 & 8.52 & 26.71 & 3.70 & 26.50 & 19.80 & 7.33 & 8.16 & 19.00 & 11.50 & 11.50 \\
mPLUG-DocOwl 1.5 & LLaMA-7B & 34.87 & 19.12 & 26.42 & 21.27 & 6.48 & 24.97 & 22.19 & 10.00 & 33.67 & 39.00 & 20.48 & 27.56 \\
SliME-13B & Vicuna-13B & 20.56 & 22.79 & 23.30 & 29.33 & 12.96 & 24.73 & 22.28 & 33.00 & 26.53 & 40.00 & 33.13 & 33.18 \\
Qwen-VL-Chat & Qwen-7B & 37.66 & 16.18 & 21.88 & 14.62 & 12.04 & 22.13 & 20.75 & 14.67 & 17.35 & 21.00 & 16.47 & 17.67 \\
LLaVA1.5-13B & Vicuna-13B & 14.47 & 22.06 & 21.59 & 24.29 & 12.96 & 20.45 & 19.30 & 27.67 & 23.47 & 31.00 & 27.51 & 27.38 \\
LLaVA-Next & LLama3-8B & 46.71 & 0.00 & 22.16 & 4.13 & 23.15 & 19.46 & 19.27 & 13.67 & 46.94 & 18.00 & 21.08 & 26.20 \\
ShareGPT4V-13B & Vicuna-13B & 14.31 & 22.06 & 15.62 & 23.99 & 12.04 & 19.26 & 17.88 & 27.33 & 24.49 & 30.00 & 27.31 & 27.27 \\
MiniGPT-v2 & Llama 2-7B-Chat & 13.98 & 22.06 & 15.34 & 24.40 & 11.11 & 19.26 & 17.69 & 13.67 & 19.39 & 24.00 & 16.87 & 19.02 \\
LLaVA1.5-7B & Vicuna-7B & 14.31 & 22.06 & 15.62 & 23.79 & 11.11 & 19.13 & 17.67 & 27.33 & 23.47 & 24.00 & 25.90 & 24.93 \\
ShareGPT4V-7B & Vicuna-7B & 14.31 & 22.06 & 15.62 & 23.79 & 11.11 & 19.13 & 17.67 & 27.33 & 23.47 & 25.00 & 26.10 & 25.27 \\
TextMonkey & Qwen-7B & 39.47 & 6.62 & 7.10 & 8.17 & 0.00 & 16.14 & 12.92 & 0.67 & 4.08 & 16.00 & 4.42 & 6.92 \\ \bottomrule
\end{tabular}%
}
\end{table}

\end{document}